\documentclass[11pt,openany]{book}
\usepackage[T1]{fontenc}
\usepackage[utf8]{inputenc}
\usepackage{lmodern}
\usepackage{arxiv}
\usepackage{amsmath,amssymb}
\usepackage{microtype}
\usepackage{enumitem}
\usepackage{xcolor}
\usepackage{listings}
\usepackage[hidelinks]{hyperref}
\hypersetup{pdftitle={Natural Language Processing: An Applied Research Practicum},pdfauthor={Mullosharaf K. Arabov}}

\renewcommand{\shorttitle}{NLP: An Applied Research Practicum}
\renewcommand{\headeright}{Research-oriented textbook}
\lstdefinestyle{pythonex}{language=Python,basicstyle=\ttfamily\footnotesize,columns=fullflexible,keepspaces=true,breaklines=true,showstringspaces=false,frame=single,framerule=0.3pt,rulecolor=\color{black!30},aboveskip=7pt,belowskip=7pt}
\title{Natural Language Processing: An Applied Research Practicum\\From Classical Text Processing to Multimodal and Agentic Systems}
\author{Mullosharaf K. Arabov}
\date{November 2026}
\begin{document}
\frontmatter
\begin{titlepage}
\centering
{\large Kazan (Volga Region) Federal University\par}
{\large Institute of Computational Mathematics and Information Technologies\par}
\vspace*{2.3cm}
{\Large Mullosharaf K. Arabov\par}
\vspace{2cm}
{\Huge\bfseries Natural Language Processing\par}
\vspace{0.6cm}
{\LARGE An Applied Research Practicum\par}
\vspace{0.6cm}
{\large From Classical Text Processing to Multimodal and Agentic Systems\par}
\vfill
{\large Kazan --- 2026\par}
\end{titlepage}
\chapter*{Abstract}
\addcontentsline{toc}{chapter}{Abstract}
This research-oriented practicum introduces natural language processing through eighteen practical works, moving from text preparation, vector representations and classical machine learning to transformer models, large language models, prompt engineering, retrieval-augmented generation, multimodal systems, preference-based alignment, agents, safety, model compression and LLMOps. Each work has three connected core tasks, a verifiable example, optional extended engineering projects and independent research directions. Shared requirements for reporting, evaluation, data provenance and reproducibility appear in one place. The material is intended for senior undergraduate students, master's students, doctoral researchers and developers, with explicit attention to low-resource and morphologically rich languages. No unexecuted experiment is represented as an empirical finding.
\medskip\noindent\textbf{Keywords:} natural language processing; NLP; transformers; large language models; RAG; prompt engineering; multimodal learning; AI safety; reproducible research; LLMOps.
\setcounter{tocdepth}{0}
\tableofcontents
\chapter*{Preface}
\addcontentsline{toc}{chapter}{Preface}
Natural language processing encompasses statistical text analysis, representation learning, information retrieval, text generation and systems built from pretrained language models. A useful practical curriculum must distinguish robust evidence from a successful software demonstration. This practicum therefore connects technical implementation with data provenance, model evaluation, experimental design and critical error analysis.

The eighteen works progress from corpus preparation through increasingly advanced methods, culminating in secure and observable applications. They are designed for different educational levels: the same three-task core can be completed with compact models and modest resources, while optional projects permit a more ambitious engineering or doctoral-level investigation. Language and dataset choices may be adapted to Russian, Tatar, Tajik and other languages, subject to suitable resources, evaluation protocols and licensing terms.

Theoretical chapters provide context rather than a substitute for empirical testing. Reproducibility here means an accurately documented procedure whose outcomes can be checked under stated conditions; it is not a promise of identical numbers across all machines. Each experimental claim must be supported by a completed run, and known limitations must be stated. The bibliography includes foundational and relevant methodological work, including selected author research on under-resourced languages.

This book may be used in instructor-led classes, seminars and independent research projects. Practical examples are distributed as accompanying Python code. Readers should check versions, model permissions and data access conditions at execution time.

The author's work on lexical resources for Tajik--Persian, subword tokenisation for Tajik, and language representations for Tatar motivates the attention given here to under-resourced settings . These examples illustrate possible research directions; they do not substitute for language-specific evaluation.

Generative AI assistance was used for English-language editing and LaTeX preparation. The human author remains responsible for approving the final text, references and claims about experimental reproducibility.

\chapter*{General Instructions, Reporting and Assessment}
\addcontentsline{toc}{chapter}{General Instructions, Reporting and Assessment}
\section*{Audience and levels of study}
The practicum serves senior undergraduates, master's students and doctoral researchers. The core of every practical work consists of exactly three connected tasks. The core is required for completing that work. Extended tasks and research directions are optional and allow further investigation without imposing large-model training, publication, cloud hosting or extensive web application development on every student. An instructor may agree on a narrower language or dataset when licences, hardware or availability are restrictive, provided that the analytical objective is preserved.

\section*{Required research workflow}
\begin{enumerate}[leftmargin=*]
\item State a testable question, define the input data and provide a meaningful baseline before selecting an elaborate model.
\item Document the data source, any licence conditions, language, number of documents, label definitions, date of access and known coverage limitations. Do not publish text or weights unless redistribution is permitted.
\item Separate training, validation and final test data. Fit learned preprocessing parameters on training material only, and avoid repeated test-set tuning. For duplicate or related documents, consider group- or time-aware splits where appropriate. Explicitly describe any transductive setting.
\item Specify the model identifier and revision, library and Python versions, hardware, random seeds, decoding options and important optimisation parameters. Seeds limit some stochasticity but do not guarantee exact reproducibility across devices or services.
\item Calculate appropriate metrics together with sample sizes, class supports or denominators, and evaluate uncertainty where feasible. Compare configurations using the same test material and budget whenever a fair comparison is intended.
\item Conduct an error analysis: include correctly solved cases, failure examples, and plausible limitations supported by inspected evidence. Do not infer a broad safety or reliability guarantee from a small demonstration.
\end{enumerate}

\section*{One report structure for all eighteen works}
A report may be submitted as a runnable notebook, a repository accompanied by a structured Markdown/HTML/PDF report, or a documented application with accessible code. In every format, include: (i) aim and task definition; (ii) data and legal provenance; (iii) methods and baseline; (iv) experimental setup and versions; (v) results in tables or directly generated figures; (vi) analysis of errors, limitations and possible biases; (vii) conclusions and references; and (viii) precise instructions for rerunning the available parts. An openly accessible URL is optional where publication would breach copyright, privacy, licence conditions or institutional rules. A private instructor-accessible copy is acceptable.

A concise dataset record (datasheet) and model record (model card) should be supplied where relevant; these may be brief in core experiments and expanded in optional research. Do not report guessed scores or claim training that did not take place. Synthetic smoke tests establish only that components execute, not that a model achieves a stated quality on real data. Any significant use of automated text-generation tools in preparing a scientific submission should be disclosed in accordance with the relevant venue's policy.

\section*{Assessment principles}
Assessment rewards methodological correctness, functioning core code, appropriate baselines, interpretation of observed results and transparent limitations. An \emph{excellent} core submission can be earned without optional tasks if all three tasks are carried out rigorously, results are reproducible under specified conditions, and the error analysis is substantive. A \emph{good} submission performs the core correctly but has limited robustness analysis; a \emph{satisfactory} submission reaches the main learning objectives with minor documented omissions; an \emph{unsatisfactory} submission lacks a verifiable experiment, essential evidence or an interpretable report. Optional engineering and research tasks can extend the grade or support thesis work, but quantity of code alone is not an assessment criterion.

\section*{Computing, ethics and security}
Use a smaller open model, fixed embeddings or a restricted subset when computational resources are limited; document the consequence for external validity. Respect data protection, consent, applicable licences and platform terms. Perform adversarial tests only in an authorised isolated environment. Retrieved text and tool output are data, not sources of permission to alter files or disclose secrets.

\section*{Accompanying examples}
The companion package contains \texttt{examples/practikum\_examples.py} and three task-specific Python modules. It uses the 20 Newsgroups dataset for many text-based examples, CIFAR-10 for the CLIP example, and a public preference-pair dataset for the data-quality demonstration. The first real run may download data or weights. The command-line option \texttt{-{}-smoke} uses synthetic inputs and is \emph{not} an empirical evaluation. See the companion README for dependencies, verification status and model/data limitations. The example programs are starting points, not completed studies of every advanced algorithm mentioned in the book.

\mainmatter

\chapter{Tokenisation and Text Normalisation}
\label{work:1}

\section{Aim and Learning Outcomes}
The objective is to build a traceable preprocessing workflow for natural-language corpora and to understand where different tokenisation and normalisation decisions alter downstream evidence. Learners compare rule-based and learned segmentation, measure coverage and lexical fragmentation, and inspect linguistically challenging cases in both well-resourced and under-resourced languages. Corpus size and publication are research design choices rather than universal prerequisites.

The three tasks below constitute the required experimental core. Larger engineering studies and independent research questions are available as optional extensions.

\section{Theoretical Background}

\noindent\textit{Principal theoretical and methodological sources for this work:} \cite{ru_reference_01,ru_reference_02,ru_reference_03,ru_reference_04,ru_reference_05,ru_reference_06,bojanowski2017enriching,arabov2026analysis,bird2009natural,ru_reference_10,arabov2026tajiknlp}.\par

New corpus-oriented tools for Tajik allow controlled testing of the same preprocessing and tokenisation decisions in a low-resource Cyrillic setting \cite{arabov2026analysis,arabov2026tajiknlp}.

Tokenisation represents a fundamental stage of the natural language processing pipeline, the quality of which determines the effectiveness of all subsequent stages of text analysis. The accuracy of computing frequency characteristics, the effectiveness of information retrieval and filtering, the quality of statistical language modelling, and the robustness of subword representations to unknown vocabulary depend on how correctly the text is segmented into linguistically or statistically meaningful units. Errors made at the tokenisation stage have a tendency to accumulate and amplify at subsequent stages of processing, making this stage critically important for any NLP project. \cite{bird2009natural,bojanowski2017enriching}

The degree of complexity of the tokenisation task varies substantially depending on the typological characteristics of a specific language. For languages with relatively clear orthographic word boundaries and limited morphological variability, such as English, superficial heuristics based on whitespace characters and punctuation prove sufficient in many cases. However, for morphologically rich languages, including Russian, Tatar, Kazakh, Ukrainian, Armenian, Georgian, and many others, a single lexeme is capable of generating dozens of surface forms through the productive processes of inflection and word formation. Under these conditions, the choice of tokenisation strategy has a direct and measurable impact on the performance of downstream tasks --- classification, clustering, information retrieval, and machine translation. Moreover, the problem of out-of-vocabulary (OOV) words, which inevitably arises in any real application, cannot be satisfactorily solved solely by dictionary methods and requires the involvement of fundamentally different approaches to segmenting the text stream.

In the modern theory and practice of natural language processing, it is customary to distinguish three main classes of tokenisation methods and the associated normalisation procedures, which differ in their linguistic foundations, computational complexity, and characteristic limitations.

The first class --- superficial methods --- performs text segmentation based on whitespace characters and punctuation marks using fixed heuristic rules. This class includes naive whitespace tokenisation, regular expression-based tokenisation, as well as library implementations provided by tools such as NLTK, spaCy, and Razdel. Superficial methods are characterised by high processing speed, linear computational complexity with respect to text length, and minimal requirements for external resources, making them convenient for prototyping and processing large data arrays in streaming mode. At the same time, these methods do not account for the morphological complexity of inflectional and agglutinative languages: word forms belonging to a single lexeme are treated as independent tokens, leading to unjustified inflation of the vocabulary and fragmentation of frequency information.

The second class --- morphologically oriented methods --- includes stemming procedures aimed at truncating a word to its stem according to fixed rules (in particular, the Porter algorithm and its variants, as well as SnowballStemmer), and lemmatization procedures that restore the normal, i.e., dictionary, form of a word using morphological analysers and dictionaries. For the Russian language, the main lemmatization tools are pymorphy3 and spaCy; for other languages, their analogues are employed, where available. These approaches provide a substantially higher degree of semantic consistency, as different word forms of a single lexeme are reduced to a unified representation. However, the price for this is dependence on external linguistic resources, potential incompleteness of dictionary coverage, computational costs of morphological analysis, and the risk of accumulating errors in cases of lexical homonymy or non-standard word usage.

The third class --- subword tokenisation methods --- consists of statistically trained algorithms that build a vocabulary of tokens directly from data, without relying on predefined linguistic rules. The key representatives of this family are Byte-Pair Encoding (BPE), WordPiece, and the Unigram Language Model. These algorithms iteratively form a vocabulary from subword units --- characters, character pairs, and longer fragments --- optimising a certain statistical criterion on the training corpus. The principal advantage of the subword approach is its ability to effectively resolve the OOV problem: any unknown word can be represented as a sequence of known subword units, down to individual characters. Furthermore, subword methods allow flexible management of vocabulary size and finding a balance between representation compactness and its expressive power. Among the limitations of these methods are the lower transparency of the segmentation process compared to linguistically motivated approaches, the sensitivity of segmentation quality to the volume and domain diversity of the training corpus, and the need for thorough empirical validation of the resulting vocabularies on representative samples.

The three classes of methods considered form a spectrum of trade-offs, the key axes of which are computational efficiency, linguistic adequacy, and the degree of dependence on external resources. Superficial methods are fast, transparent, and undemanding of data, yet linguistically limited. Morphologically oriented methods introduce explicit linguistic knowledge into the processing and provide higher normalisation quality, but depend on the completeness and relevance of the dictionaries and rules used. Subword methods extract the optimal segmentation strategy directly from data and offer a fundamental solution to the problem of unknown words; however, their behaviour is less interpretable and requires systematic empirical verification on the material of various languages and domains. Forming the learner's ability to quantitatively evaluate and meaningfully interpret these trade-offs constitutes one of the central educational outcomes of this work.

\subsection{Experimental Interpretation}
Subword tokenisation must not be conflated with linguistic morphological analysis. BPE, WordPiece and Unigram tokenisers learn different statistical objectives; exact round-trip reconstruction depends on normalisation, pre-tokenisation, alphabet coverage, decoder configuration and treatment of whitespace. A corpus size such as five million tokens can be a useful extended-project target but is not a universal theoretical minimum for training a subword model.

\section{Execution Procedure}

\subsection{Core: Three Required Tasks}
\begin{enumerate}[leftmargin=*,label=\textbf{\arabic*.}]

\item Choose a licensed corpus and document its source, language, domain and unit of splitting. Use a fixed training/held-out separation; retain raw texts and configuration files without publishing material lacking redistribution rights.

Start with a named dataset or a corpus obtained under an explicitly identified access and reuse policy. Define the document unit before any split, record the approximate number of documents and languages represented, and save a manifest of document identifiers. Separate the data at document level, checking for duplicates or near-duplicates where appropriate; fit any learned tokenizer only on the training portion. Explain whether the chosen material represents one domain or several, since this affects how far the observations can be generalised.

\item Implement two distinct tokenisation or normalisation configurations on the same documents. Compare token counts, vocabulary sizes, segmentation of difficult examples, and processing time using a fixed evaluation sample.

State the exact order of cleaning, Unicode handling, case conversion, punctuation processing and tokenisation for each configuration. Where a statistical subword tokenizer is included, document its training subset, algorithm, vocabulary size, normaliser and decoder; where it is not available, compare two genuine, clearly different rule-based approaches. Apply both configurations to the same frozen evaluation sample and produce a comparison table with counts, throughput or elapsed time, and concrete segmentations. Do not compare timings without reporting the execution environment and measurement procedure.

\item Describe at least ten observed errors or disagreements, including non-standard orthography or inflection. Verify that learned vocabularies are fitted on training texts only; distinguish character coverage from an unqualified promise of exact text recovery.

Build a small, inspection-ready error table with the source string, both outputs, the expected linguistic interpretation if one can be justified, and a reasoned error category. Include inflection, compounds, punctuation, numbers or other phenomena actually present in the sample; do not manufacture examples and present them as corpus observations. If round-trip reconstruction is tested, compare decoded text byte-for-byte or under a stated normalisation policy and identify every lossy operation that was applied before tokenisation. Conclude which method suits the observed material and what evidence is still missing.

\end{enumerate}

\subsection{Reproducible Example and Its Limits}

The companion project contains executable code for this work. Run the real-data path with
\begin{center}\texttt{python practikum\_examples.py 1}\end{center}
Real 20 Newsgroups messages; token counts and boundary inspection only. For a technical-only check, add \texttt{-{}-smoke}; its synthetic inputs must not be used as scientific evaluation results. The program does not contain fabricated model scores. Record actual outputs, sample sizes, versions, timings and failures after running it.

\subsection{Extended Engineering Tasks (Optional)}
The following tasks provide a fuller engineering and research programme. They are optional rather than prerequisites for passing the core. The numerical scales and model choices below describe ambitious projects; they are not universal requirements. Documentation and reporting follow the shared protocol above.

\paragraph{Extended Task 1.} The learner forms a text corpus of at least five million tokens in Russian or another native language, the choice of which is made independently. Permitted languages include Russian, Tatar, Kazakh, Ukrainian, Belarusian, Armenian, Georgian, Kyrgyz, Azerbaijani, Uzbek, Tajik, Bashkir, Chuvash, and other languages, as well as English for international learners or as a contrasting example of an analytic language. The choice of a native language is methodologically encouraged, as it gives the learner the opportunity to rely on their own linguistic intuition when qualitatively assessing normalisation results and simultaneously contributes to the creation of corpus resources for low-resource languages.

Five million tokens is an optional extended-project scale rather than a necessary threshold for statistical validity. In cases where the collection of a corpus of five million tokens is objectively difficult --- for example, for low-resource languages with limited corpus material --- the learner has the right, in agreement with the instructor, to use a smaller volume corpus, explicitly stipulating this circumstance in the datasheet and discussing its influence on the interpretability of subword tokenisation metrics.

The corpus must include texts originating from at least two different domains, by which is meant genre or functional-stylistic categories: news texts, e-books, popular science publications, legislative acts and official documents, social media texts, and encyclopaedic articles. Sources for text data may include any publicly accessible news sites, electronic libraries, archives of popular science publications, Wikipedia dumps in the corresponding language, open datasets of a scientific and educational nature, as well as other resources whose use is not prohibited by current legislation and does not violate the principle of fair use concerning educational and research purposes.

The learner is entitled either to carry out independent collection of texts from web resources using tools such as \texttt{requests} and \texttt{BeautifulSoup4} for static pages or \texttt{selenium} for dynamic content, or to utilise one or more ready-made open corpora --- such as the Taiga Corpus, Lenta.ru Dataset, OpenCorpora, Tatar Corpus, UberText, Wikipedia dumps, and others. When using a ready-made corpus, the learner is obliged to indicate its origin and licence. Methodologically, both tracks are considered equivalent; the choice between them is determined by data availability, the technical preparation of the learner, and their research interests and does not affect the maximum possible grade.

The collected or prepared corpus is saved in JSONL format with UTF-8 encoding, where each line represents a JSON object containing the following fields: \texttt{id} --- unique document identifier; \texttt{text} --- full text after cleaning; \texttt{title} --- heading (optional); \texttt{source} --- source; \texttt{category} --- domain or category (optional); \texttt{language} --- language code according to the ISO 639-1 or ISO 639-3 standard; \texttt{date} --- publication date in ISO 8601 format (optional); \texttt{url} --- source URL (optional); \texttt{tokens\_approx} --- approximate number of tokens, to be filled in after tokenisation. All texts undergo a procedure for removing HTML tags, scripts, CSS insertions, navigation and advertising blocks, as well as normalising whitespace characters.

Compliance with the \texttt{robots.txt} rules for each source website is mandatory. Bypassing technical protection measures, such as CAPTCHA and rate-limiting systems, without the explicit permission of the resource owner is not permitted. Respect data protection law and relevant licences, including for publicly accessible documents. 

\paragraph{Extended Task 2.} A software module \texttt{text\_cleaner.py} is implemented, intended for the preliminary cleaning of the collected text data and their unification before being fed to the input of tokenisation and normalisation modules. The module must ensure the removal of all residual HTML tags, scripts, and other non-textual components that may have remained after the primary processing at the data collection stage; the normalisation of whitespace characters, including collapsing multiple spaces, tab characters, and redundant line breaks into a single space; optional lowercasing of the text while preserving linguistically significant elements, including abbreviations, hyphens in compound words, and unit-of-measurement designations; as well as the filtering of stop words using standard dictionaries. For the Russian language, the \texttt{stop-words-ru} library is used; for other languages, corresponding lists from \texttt{nltk.corpus.stopwords} are used or are provided by the learner independently. Particular attention is paid to the correct handling of various encodings and alphabets, including Cyrillic scripts, Armenian script, Georgian Mkhedruli, and Arabic script, if these are present in the selected language material.

Methodologically important is the choice of the stage at which stop word filtering is applied. Removing stop words before morphological normalisation (stemming or lemmatization) may lead to incomplete filtering: word forms relating to stop words, but morphologically altered, will not be recognised from the dictionary and will remain in the text. On the other hand, removing stop words prior to training subword models deprives the model of contextual information that may be significant for segmentation quality. In connection with this, the module \texttt{text\_cleaner.py} must support the ability to specify the order of operations via a configuration parameter, namely: filtering before normalisation, after normalisation, or disabling filtering altogether. The learner justifies the choice of a specific strategy in the methodological section of the report, based on the nature of the task being solved and the features of the tokenisation methods used. A detailed investigation of the influence of this choice on corpus characteristics is relegated to the first additional task (Section 4).

The module code is formatted as a separate file with exhaustive documentation of all public functions and classes, which guarantees its reusability in subsequent tasks and its testability by automated means.

\paragraph{Extended Task 3.} A universal preprocessing and text standardisation module is developed, which serves as a connecting link between the cleaning stage and the tokenisation stage. The module must perform unified processing of punctuation marks and hyphens; the replacement of URLs, email addresses, numerical values, and other types of special tokens with standardised markers, such as \texttt{<URL>}, \texttt{<EMAIL>}, \texttt{<NUMBER>}, and analogous ones; and also optionally support the expansion of common contractions and abbreviations specific to the chosen language. All processing logic is implemented on the basis of regular expressions, which ensures robustness to the variability of input data. The key architectural requirement is the support for loading processing rules from an external configuration file in YAML or JSON format: this approach allows the module's behaviour to be modified promptly without changing its source code, which is critically important for ensuring the reproducibility of experiments and the adaptation of the pipeline to various languages and domains. The module must be designed in such a way that it can be used as part of a unified preprocessor combining cleaning, standardisation, and tokenisation.

\paragraph{Extended Task 4.} On the material of the prepared and cleaned corpus, a descriptive statistical analysis and visualisation of its key characteristics are performed. Within the framework of this task, the following basic statistics are computed: the total number of tokens in the corpus; the number of unique tokens; the mean text length and mean heading length in tokens; the proportion of stop words before and after applying filtering. Histograms of the distributions of text and heading lengths are constructed. If metadata on categories is present in the corpus, box plots or summary tables are constructed, characterising the distribution of text length by category. An analysis of the structure of missing values is carried out using the \texttt{missingno} library or analogous tools, which makes it possible to identify potential problems with the completeness of metadata --- the absence of categories, publication dates, or other attributes --- which may affect the interpretability of the results of the subsequent analysis. The set of obtained statistics and visualisations forms the factual basis for comparing tokenisation and normalisation methods, making it possible to assess their influence on the key characteristics of the corpus not speculatively, but on the basis of empirical data.

\paragraph{Extended Task 5.} A comparative analysis of classical tokenisation and normalisation methods is carried out based on a system of formalised metrics. The set of methods under investigation includes: naive whitespace tokenisation; regular expression-based tokenisation; tokenisation using library implementations provided by NLTK, spaCy, and Razdel tools; stemming using the PorterStemmer and SnowballStemmer algorithms for supported languages; lemmatization by means of pymorphy3 and spaCy for Russian or their counterparts for other languages, where available.

Each method is evaluated according to the following set of criteria. Firstly, the size of the resulting vocabulary, i.e., the number of unique tokens generated by the method. Secondly, the proportion of out-of-vocabulary words on a held-out sample not used during the method's configuration. Thirdly, the time complexity of processing, measured as the mean time spent processing a fixed volume of text.

Fourthly --- and this is the central metric of the entire comparison --- a direct quantitative assessment of semantic consistency, by which is meant the ability of the method to reduce different word forms of a single lexeme to a unified representation. To measure it, the learner forms a control list containing from twenty to thirty lexemes, each of which is represented in the corpus by several different word forms (it is recommended to select lexemes having at least five different surface forms in the corpus texts). For each of the compared tokenisation and normalisation methods, the word-form nest compression coefficient is computed: the ratio of the number of unique tokens generated by the method for all word forms of the given lexeme to the number of word forms themselves. A coefficient value close to one divided by the number of word forms corresponds to ideal normalisation, whereby all word forms are reduced to a single representation; a value equal to one means that normalisation has produced no effect. The final metric of semantic consistency is computed as the arithmetic mean of the compression coefficients over all lexemes in the control list. This procedure formalises the concept of normalisation quality and makes the comparison of methods objective and reproducible. The results for each method and each lexeme are summarised in a table and accompanied by a discussion of cases in which normalisation proved ineffective or, conversely, excessive.

\paragraph{Extended Task 6.} The training of three subword tokenisation models --- Byte-Pair Encoding (BPE), WordPiece, and Unigram Language Model --- and their subsequent comparison under controlled experimental conditions is carried out. Training is performed exclusively on the training portion of the corpus; all evaluation metrics are computed on a held-out sample that was not used in the process of building the vocabularies, which excludes the risk of an overly optimistic estimate of the models' generalisation ability. The \texttt{tokenizers} library from Hugging Face and/or \texttt{sentencepiece} are used as the toolset. The experimental parameters fixed for all models are: vocabulary size, for which values of 8,000, 16,000, and 32,000 tokens are investigated, and a minimum frequency threshold set at two occurrences of a token in the training corpus. For each model and each vocabulary size, the following metrics are computed: the mean token length in characters; the word fragmentation coefficient, defined as the mean number of tokens per source word; the compression ratio of the text, measured in characters and in the number of tokens compared with the original representation; as well as the possibility of complete and exact reconstruction of the source text from the token sequence, confirming the correctness and invertibility of the transformation. Additionally, the stability of segmentation quality is analysed when models trained on one domain are transferred to texts of another domain present in the corpus. The training and comparison results are formatted as summary tables and graphs, accompanied by analytical commentary.

\paragraph{Extended Task 7.} An in-depth linguo-statistical analysis of the corpus is performed with a focus on its lexical structure and on how the choice of tokenisation and normalisation method influences the observed statistical regularities. Within this task, the empirical Zipf's law curve is constructed, reflecting the dependence of the logarithm of word frequency on its logarithmic rank; a comparison of the obtained curve with the theoretical distribution is performed, and a quantitative estimate of the deviation is made. Word clouds are visualised before and after applying normalisation procedures, which provides a clear representation of how stemming and lemmatization transform the frequency landscape of the corpus. For subword models, heat maps of the frequency of the most heavily fragmented words are constructed, making it possible to identify lexical groups that are most subject to fragmentation during subword segmentation. Interactive diagrams of token distribution by frequency and length are created for each of the compared methods, providing an opportunity for intuitive comparison of their properties. All visualisations are generated directly in the course of code execution and are included in the analytical report.

\paragraph{Extended Task 8.} The learner performs a formal verification of the correctness of the corpus splitting and the absence of data leakage between the training and test sets. Within this task, the splitting procedure is documented: the learner records, in the form of a separate script or notebook cell, the algorithm by which the corpus was divided into training and test portions, with the mandatory indication of the random seed value used to ensure reproducibility. Next, a check for the absence of intersections is performed: the sets of unique document identifiers that ended up in the training and test sets are programmatically compared; any non-empty intersection is classified as a critical error and must be immediately corrected. For the final verification protocol, the following indicators are computed and output: the absolute volume of the training and test sets in tokens; the proportion of the test set relative to the total corpus volume; the number and proportion of unique tokens in the test set that are absent from the training set, which gives an empirical estimate of the expected OOV proportion; as well as the degree of overlap of the vocabularies built separately on the training and on the full data collection, as an additional indicator of the representativeness of the split. This verification protocol is included in the analytical report and serves as documentary confirmation of the methodological rigour of the experiment.

\paragraph{Extended Task 9.} A unified text preprocessor is developed in the form of a Python class with the methods \texttt{.fit()} and \texttt{.transform()}, providing a uniform interface for all tokenisation and normalisation methods implemented in the work. The preprocessor must support the preservation of internal state --- built vocabularies, tokenizer parameters, stop word lists, and other training artefacts --- in its object and provide means for the serialisation and deserialisation of this state, for example, using the \texttt{pickle} or \texttt{joblib} libraries. An instance of the preprocessor, having been once trained on the training set by calling \texttt{.fit()}, must provide identical results upon multiple calls to \texttt{.transform()} on the same data, which facilitates repeated analysis under documented conditions. The class architecture is designed with the prospect of its subsequent embedding into machine learning pipelines, including jointly with scikit-learn tools, which forms in the learner the skills of developing NLP system components suitable for industrial operation.

\paragraph{Extended Task 10.} A suite of automated tests is created to ensure the verification of the correctness and reproducibility of all key project components. Using the pytest framework, unit tests are developed that verify the following properties of the software implementation: the idempotency of the cleaning module, i.e., the identity of the result upon repeated application to already cleaned text; the determinism of the preprocessor with a fixed random seed value; the correctness and invertibility of transformations --- the ability of the decoder of a trained BPE model to fully reconstruct the source text from the token sequence without loss or distortion; the absence of data leakage between the training and test sets, verifiable through the intersection of sets of unique document identifiers; as well as the conformity of the types and dimensionalities of the output data to those declared in the module interfaces. The versions of all software libraries used are recorded in a \texttt{requirements.txt} or \texttt{environment.yml} file. A script is developed for the automatic execution of the entire test suite and the generation of a pass/fail report, which ensures the long-term reliability and maintainability of the codebase. The presence of a working test suite is regarded as an integral part of the software artefact, not as an optional supplement.

\paragraph{Extended Task 11.} An interactive web tool is developed, intended for the visual and functional analysis of text data and the comparison of the processing methods implemented in the work. The application must provide the user with the following capabilities: uploading a text file in CSV or JSONL format or selecting from built-in examples that include corpus fragments; choosing a tokenisation method and its parameters, including the vocabulary size for trained subword models; automatic construction of token length distributions, frequency spectra, and OOV token proportion indicators for the selected method; visualisation of subword decomposition for an arbitrary entered word, making it possible to see how a given word is segmented into subword units by each of the trained models; and generation of an analytical report in HTML or PDF format, summarising the key metrics and visualisations. The recommended technologies for implementation are Streamlit in combination with Plotly or Bokeh to ensure interactivity of the visualisations. The web tool must be designed in such a way that it can be used by a user without programming skills --- for example, a linguist-researcher interested in the express analysis of textual material.

\paragraph{Extended Task 12.} To ensure the possibility of fully reproducing all the obtained results, as well as for the potential citation and use of the created artefact by other researchers, the learner is recommended, but not strictly obligated, to publish the corpus, source code, configuration files, trained models, and analytical report in a public repository on GitHub or GitLab platforms. Additionally, it is recommended to deploy the web application on an open platform --- Hugging Face Spaces or Streamlit Cloud --- which makes the tool accessible for use without the need for local installation. In the event of a decision to publish, the learner must compile a detailed meta-description of the project, including the corpus characteristics (volume in documents and tokens, sources, language composition, distribution by categories), the training parameters of the subword models (vocabulary sizes, frequency threshold, number of epochs), key metrics (OOV proportion, fragmentation coefficient, processing time), a description of identified limitations and systematic biases, licensing conditions, and a step-by-step instruction for reproducing all stages of the work in a \texttt{reproduce.md} file. For code, the use of open licences MIT or Apache 2.0 is recommended; for data, Creative Commons Attribution 4.0 (CC BY 4.0) or a compatible open licence. Learners performing the work within the framework of closed research projects or using restricted-distribution data are exempted from the obligation to publish without any negative consequences for the grade; openness is welcomed but is not an end in itself.

\subsection{Further Research Directions (Optional)}
\textbf{First Additional Task.} This presupposes a systematic analysis of the influence of stop words on corpus characteristics and tokenisation quality. The learner is asked to quantitatively investigate how the removal of stop words at various stages of the pipeline --- before normalisation, after normalisation, and in both variants simultaneously --- affects the final vocabulary size, the frequency distribution of tokens, and the proportion of OOV tokens when using subword models. The results must be presented in the form of comparative tables and accompanied by a discussion of the application scenarios in which stop word filtering is justified, and those in which it leads to the loss of significant syntactic information.

\textbf{Second Additional Task.} This is aimed at comparing the subword segmentation obtained by statistical methods with the results of the work of linguistic tokenizers. The learner is asked to train a simple BPE model without the merge operation, i.e., effectively a model for extracting character-frequency patterns, on a corpus from which all spaces have been preliminarily removed, and to compare the token boundaries found by such a model with the boundaries determined by library tokenizers such as spaCy and Razdel. This task allows a practical assessment of the extent to which the statistical regularities extracted from raw text are capable of approximating linguistically motivated segmentation of the speech stream.

\textbf{Third Additional Task.} This provides for the adaptation of the developed preprocessing pipeline to social media texts --- messages from VK, Twitter, Telegram channels. The specificity of such texts, including the presence of hashtags, mentions, emojis, non-standard punctuation, spelling errors, and slang expressions, requires modification of the cleaning and standardisation modules, as well as a critical reconsideration of the applicability of standard stop word dictionaries and morphological analysers. The learner must modify the preprocessor's configuration file and, if necessary, supplement the \texttt{text\_cleaner.py} module with new rules, after which a comparative analysis of tokenisation quality is performed on the material of social media and news texts.

\textbf{Fourth Additional Task.} This is aimed at an in-depth investigation of the invertibility of subword tokenisation. The learner is invited to conduct an experiment measuring the proportion of information lost during the `text $\rightarrow$ tokens $\rightarrow$ text' transformation for each of the three trained algorithms, with an emphasis on borderline cases: hyphenated words, multi-component terms, numerical expressions with units of measurement, emojis, and symbols of non-Latin alphabets. The result should be a classification of the types of reconstruction errors and a quantitative estimate of their frequency.

\textbf{Fifth Additional Task.} This consists of a comparative analysis of the computational efficiency of the implemented methods on the material of corpora of various sizes. The learner constructs plots of the dependence of processing time on the volume of input data in tokens for each of the tokenisation and normalisation methods investigated in the work, determines the asymptotic complexity, empirically verifies it, and formulates practical recommendations for choosing a method under conditions of limited computational resources.

\section{Conclusion}
This work is not merely an exercise in segmenting text into words or reducing word forms to their dictionary form. It represents the foundation upon which all procedures of automatic natural language analysis are subsequently built, without exception. It is at this stage that the learner develops an understanding of the fact that the quality of any NLP solution --- from the simplest search index to complex generative models --- is laid down even before the first vector representation, the first classifier, or the first predicted tag appears.

In the process of performing the work, the learner learns to pose the key questions that determine the reliability of the text pipeline. Whether the text is correctly segmented into meaningful units --- this is a question about the linguistic adequacy of tokenisation. Whether the word forms of a single lexeme are reduced to a unified representation without mixing with the forms of other lexemes --- this is a question about the semantic accuracy of normalisation. Whether the reproducibility of results is guaranteed throughout the pipeline --- from data collection to the final token --- this is a question about the transparency, testability, and openness of the software code. The set of answers to these questions constitutes the boundary separating raw text from a corpus ready for analysis.

It is precisely this kind of systematic, critically considered, and technically disciplined approach to working with text data that lies at the foundation of all modern research and applied developments in the field of natural language processing. Upon completion of the work, the learner acquires not only a functional toolset for tokenisation and normalisation but also a methodological foundation applicable to all subsequent projects --- from academic research to the creation of industrial language systems. The acquired skills in documenting data, constructing reproducible pipelines, performing comparative analyses of algorithms, and formally evaluating their quality form the core of professional competencies that are in demand in both scientific and industrial environments.

\chapter{Methods of Text Vectorisation}
\label{work:2}

\section{Aim and Learning Outcomes}
This work examines how lexical counts, TF--IDF and learned vector spaces represent documents and where each representation is useful or misleading. Learners acquire the ability to fit transformations without leakage, quantify dimensionality and sparsity, test representations against an unchanged task baseline, and explain errors arising from vocabulary and domain coverage.

The three tasks below constitute the required experimental core. Larger engineering studies and independent research questions are available as optional extensions.

\section{Theoretical Background}

\noindent\textit{Principal theoretical and methodological sources for this work:} \cite{manning1999foundations,ru_reference_13,ru_reference_14,ru_reference_15,mikolov2013efficient,reimers2019sentence,arabov2026tatar2vec,ru_reference_19,arabov2026tatarclassification}.\par

A dedicated Tatar word-embedding resource and a Tatar text-classification benchmark offer examples of language-specific representation evaluation \cite{arabov2026tatar2vec,arabov2026tatarclassification}.

Text vectorisation --- the transformation of unstructured text information into numerical vector representations --- is the central stage of the NLP pipeline, largely determining the quality of solving tasks such as classification, clustering, semantic search, machine translation, and text generation. A vector representation is not simply a convenient format for computation; it represents a hypothesis about exactly which properties of the text carry meaning. Two documents represented by close vectors will be regarded by any subsequent algorithm as similar, regardless of whether this similarity is due to semantics, stylistics, or is an artefact of the chosen vectorisation method. Therefore, the choice of a vectorisation strategy is just as significant as the choice of model architecture and requires a systematic analysis of the trade-offs between sparsity and density, dimensionality and expressiveness, and interpretability and semantic depth. \cite{mikolov2013efficient,bojanowski2017enriching}

In the modern theory and practice of NLP, three main classes of vectorisation methods are distinguished, differing in the method of forming the feature space, computational complexity, and the nature of the obtained representations.

The first class --- statistical methods --- is based on the frequency characteristics of words and documents. It includes One-Hot Encoding, representing each word as a binary vector with a single one; Bag of Words (BoW), describing a document as a frequency vector over the corpus vocabulary; and TF-IDF, which weights the term frequency taking into account its rarity in the corpus, reducing the weight of frequent but uninformative words. Statistical methods are transparent, easily interpretable at the level of individual features, and allow efficient computation; however, they suffer from the ``curse of dimensionality'': as the vocabulary grows, the dimensionality of the vectors increases linearly, and the representations become extremely sparse, which complicates the modelling of semantic relations. Moreover, such vectors capture only the superficial co-occurrence of words and ignore synonymy and polysemy: the words ``automobile'' and ``car'' will be far apart in the BoW/TF-IDF space, despite their semantic closeness.

The second class --- dimensionality reduction methods --- is aimed at revealing latent semantic structures in frequency matrices. The best-known representative is Latent Semantic Analysis (LSA), which performs truncated singular value decomposition (TruncatedSVD) of the ``term-document'' matrix and yields low-dimensional dense vectors in which words or documents close in meaning are grouped together. Non-linear methods such as UMAP and $t$-SNE are used primarily for visualising high-dimensional data on a plane, preserving the local neighbourhood structure. Methods of this class mitigate the sparsity problem and are capable of revealing thematic groupings; however, the resulting dimensions are often difficult to interpret, and projection is inevitably accompanied by information loss. Furthermore, LSA is sensitive to frequency distortions and does not model non-linear relationships.

The third class --- distributed representations (embeddings) --- forms the dominant paradigm of modern vector semantics. The key idea is to represent a word or document as a dense vector of a fixed dimensionality (usually from 50 to 300), trained in such a way that geometric closeness in the vector space reflects semantic or syntactic closeness. Within this class, word-level models are distinguished: Word2Vec (with CBOW and Skip-gram architectures), predicting a word from its context or context from a word; GloVe, combining global co-occurrence statistics with a local context window; and FastText, extending Word2Vec by representing words as the sum of subword $n$-gram vectors, which is especially useful for morphologically rich languages and rare words. At the document level, the key model is Doc2Vec (Paragraph Vector), which, in the PV-DM and PV-DBOW variants, jointly trains a document vector together with word vectors. Embeddings are compact, possess a rich semantic structure, and allow the performance of operations such as vector arithmetic (e.g., ``king $-$ man $+$ woman $\approx$ queen''); however, their training requires significant computational resources and careful hyperparameter tuning, and the individual dimensions of dense vectors generally do not have a transparent linguistic interpretation.

The distinction between sparse (statistical) and dense (embedding) representations is of fundamental importance for practical application. Sparse vectors are interpretable at the feature level: one can directly indicate which words contributed to a classification decision. Dense embeddings, on the contrary, distribute meaning across all coordinates, making individual features uninterpretable, but thereby capturing much higher-level semantic relations, including analogies, thematic clusters, and even socio-cultural biases. Understanding these trade-offs and forming the ability to quantitatively evaluate them constitutes the methodological core of this work.

\subsection{Experimental Interpretation}
When computing test-set features, fit the vocabulary and any TF--IDF statistics exclusively on the training partition. Embeddings learned from a document collection may require separate analysis of transductive versus inductive assumptions. Nearest-neighbour similarity does not by itself establish semantic equivalence.

\section{Execution Procedure}

\subsection{Core: Three Required Tasks}
\begin{enumerate}[leftmargin=*,label=\textbf{\arabic*.}]

\item Fit count-based and TF--IDF representations using training documents only. Report vocabulary size, matrix shape, sparsity and the handling of unseen terms.

Choose a documented labelled corpus or a clearly defined retrieval/similarity collection and keep a reproducible split. Use the same tokenisation and document boundaries for a count vectoriser and a TF--IDF vectoriser; record options such as n-gram range, minimum document frequency and case handling. Inspect the learned feature names and verify that inverse document frequencies and vocabulary construction use training documents only. Present the number of rows and columns, nonzero entries, and any out-of-vocabulary behaviour on held-out documents.

\item Compare the two representations under a fixed classifier or similarity task and an identical held-out split. Report macro-F1 (or another task-appropriate measure) and the number of test examples.

Select one downstream model, for example logistic regression or a linear support-vector classifier, and hold its training protocol constant while changing only the document representation. Fit model selection or hyperparameter tuning within the training portion or a separate validation portion; reserve the final test set for a single comparison. Report a class-aware metric, individual class support and at least one simple reference such as majority-class prediction. If the chosen task is similarity rather than classification, justify its relevance criterion and use a suitable labelled evaluation set instead of reporting a classification metric.

\item Interpret important features and at least ten contrasting document pairs; discuss how preprocessing changes the representation and avoid selecting settings on the held-out test set.

Inspect representative positive and negative feature coefficients or similar feature-attribution evidence only for models where their interpretation is meaningful. Select contrasting examples by an explicit rule (for example, high-confidence errors and low-confidence correct predictions) rather than reporting only favourable cases. Explain vocabulary sparsity, domain vocabulary and the effect of case-folding or stop-word decisions. Acknowledge that TF--IDF reflects corpus statistics and cannot by itself establish semantic equivalence between texts.

\end{enumerate}

\subsection{Reproducible Example and Its Limits}

The companion project contains executable code for this work. Run the real-data path with
\begin{center}\texttt{python practikum\_examples.py 2}\end{center}
20 Newsgroups; lexical feature extraction is fitted on training data. For a technical-only check, add \texttt{-{}-smoke}; its synthetic inputs must not be used as scientific evaluation results. The program does not contain fabricated model scores. Record actual outputs, sample sizes, versions, timings and failures after running it.

\subsection{Extended Engineering Tasks (Optional)}
The following tasks provide a fuller engineering and research programme. They are optional rather than prerequisites for passing the core. The numerical scales and model choices below describe ambitious projects; they are not universal requirements. Documentation and reporting follow the shared protocol above.

\paragraph{Extended Task 1.} The text corpus formed during Practical Work~\ref{work:1} is used as the experimental base. The learner loads the corpus from a JSONL file, checks the integrity and completeness of the metadata (the presence of the fields \texttt{id}, \texttt{text}, \texttt{source}, \texttt{category}, \texttt{language}, \texttt{date} where available), and also ensures that the corpus volume is at least five million tokens. If, in Practical Work~\ref{work:1}, by agreement with the instructor, a smaller volume corpus was used (e.g., for a low-resource language), this circumstance is recorded in the datasheet, and its possible influence on the stability of embedding metrics is discussed in the analytical report.

If various tokenisation and normalisation strategies were applied within the framework of the previous work, the learner makes a substantiated choice of one of them as the base for vectorisation, guided by the previously obtained quality metrics (vocabulary size, OOV proportion, word-form nest compression coefficient). The application of several strategies with a subsequent comparison of their influence on the quality of vector representations is permitted within the framework of an additional task. A reference to the specific artefacts of Practical Work~\ref{work:1} (the path to the corpus file, the datasheet, the chosen tokenisation and normalisation scheme) is recorded in the methodological section of the report to ensure full traceability of the experiment.

If the learner performs the present work independently of Practical Work~\ref{work:1}, they must first create or select a corpus satisfying the requirements set out in Task 1 of the previous work and explicitly indicate its origin and characteristics in the datasheet. In all cases, special attention is paid to the fact that all further model training operations are performed exclusively on the training portion of the corpus, and the split must be fixed before the start of the experiments.

\paragraph{Extended Task 2.} A software module \texttt{classical\_vectorizers.py} is developed, implementing classical statistical methods for transforming text into a vector space. The module must provide for the construction of representations based on One-Hot Encoding, Bag of Words, and TF-IDF, as well as support working with $n$-grams --- unigrams, bigrams, and trigrams --- both separately and in combination. The implementation is based on the scikit-learn library: \texttt{CountVectorizer} is used for BoW, and \texttt{TfidfVectorizer} for TF-IDF. The learner may provide for dimensionality control of the resulting matrices through the parameters \texttt{max\_features}, \texttt{min\_df}, and \texttt{max\_df}, as well as ensure efficient memory usage when working with large corpora, in particular, by storing matrices in sparse format (scipy sparse CSR/CSC) and monitoring RAM consumption. The module code is furnished with exhaustive documentation and supports the saving of the trained vectorizer for subsequent application to new data.

\paragraph{Extended Task 3.} A module for revealing latent topics and semantic relations in the corpus by dimensionality reduction methods is implemented. Within this module, truncated singular value decomposition (TruncatedSVD) is applied as part of Latent Semantic Analysis (LSA) with control of the proportion of explained variance, the recommended range of which is from eighty to ninety-five per cent. Additionally, non-linear visualisation methods --- UMAP and $t$-SNE --- are employed for displaying high-dimensional vector representations on a two-dimensional plane while preserving the local neighbourhood structure. The aim of the stage is not only the compression of the representation but also the meaningful interpretation of the principal components: the learner analyses which words or documents make the greatest contribution to each of the extracted components and attempts to identify the corresponding thematic clusters. The dependence of the quality of thematic separation on the number of components and the chosen visualisation method is assessed; the results are visualised in the form of two-dimensional scatter plots with colour coding by category, if such are available in the metadata.

\paragraph{Extended Task 4.} An empirical comparison of statistical approaches --- One-Hot Encoding, BoW, TF-IDF, and $n$-gram models --- is carried out based on a system of formalised metrics. For each method, the following are recorded: the dimensionality of the vector space (number of features); the degree of sparsity, computed as the proportion of non-zero elements in the ``document-feature'' matrix; semantic consistency, operationalised as the mean cosine similarity between documents belonging to the same category (if categorical annotation is available), or as similarity to expertly selected paraphrases; the time spent on constructing the matrix and computing pairwise similarities; and the amount of RAM consumed. The results are summarised in a single table and saved in CSV format (\texttt{vectorization\_metrics.csv}). The obtained data are accompanied by a qualitative interpretation revealing the trade-offs between the expressiveness of the representations and computational efficiency: in particular, the learner must indicate under what conditions the use of TF-IDF with unigrams is preferable, and under what conditions a transition to bigrams or to dimensionality reduction is preferable.

\paragraph{Extended Task 5.} The training and comparison of modern embedding models on the prepared corpus are performed. Both word-level models are investigated --- Word2Vec in the CBOW and Skip-gram variants, FastText, and GloVe --- as well as the document-level model Doc2Vec in the PV-DM and PV-DBOW modes. The experiment is conducted under controlled conditions: the vector dimensionality is fixed at values of 100, 200, and 300; the context window size is set to 5 and 10; the minimum word frequency (\texttt{min\_count}) is taken as 5 or 10 depending on the corpus volume and the proportion of rare vocabulary. The gensim library (for Word2Vec, FastText, Doc2Vec), fasttext (from Facebook Research), and glove-python are used as the toolset. The evaluation of the quality of the trained models includes: vocabulary coverage, i.e., the proportion of corpus tokens for which an embedding has been obtained; the quality of semantic analogies, measured as the accuracy of solving tasks of the form ``A is to B as C is to ?'' on standard test sets or on lists compiled by the learner; and clustering efficiency, assessed by the Adjusted Rand Index (ARI) when comparing the obtained clusters with known document categories. The results are formatted as summary tables and graphs.

\paragraph{Extended Task 6.} The deep semantic structure of the trained embeddings is investigated. This stage includes four complementary directions of analysis. Firstly, the cosine similarity is computed for pairs of synonyms, antonyms, and thematically related terms for the purpose of checking to what extent the geometry of the vector space corresponds to lexicographic notions of closeness of meaning. Secondly, the verification of vector arithmetic is performed on examples of semantic analogies (such as ``Moscow $-$ Russia $+$ France $\approx$ Paris''), morphological transformations (``worked\textsubscript{MASC} $+$ feminine $\approx$ worked\textsubscript{FEM}''), and geographical comparisons; the accuracy of arithmetic operations is quantitatively assessed by the proportion of cases in which the nearest neighbour to the computed vector is the expected word. Thirdly, semantic axes (e.g., gender, sentiment, social status) are identified and quantitatively characterised, and possible systematic biases are assessed by measuring the projections of vectors onto these axes. Fourthly, for a series of target words, an analysis of the ten nearest neighbours in the vector space is performed with an expert assessment of their linguistic and thematic consistency. All results are recorded in the form of tables, screenshots, and extensive analytical commentary for subsequent inclusion in the final report.

\paragraph{Extended Task 7.} An interactive web application is created, allowing the user to explore and compare vectorisation methods in real time. The application must support: the input of arbitrary text by the user or selection from built-in examples (including documents from the corpus); an interactive calculator of cosine similarity and vector arithmetic, with the ability to specify vectors by entering words or mathematical expressions; the visualisation of the projection of words or documents onto user-selected semantic axes using two-dimensional diagrams; and the automatic generation of an HTML report containing graphs, tables, and textual conclusions. The recommended technologies are Streamlit or Gradio in combination with Plotly or Bokeh to ensure interactivity. The application is designed in such a way that it can be used by a researcher without programming skills and must be deployed locally with the possibility of subsequent deployment to a cloud platform at the learner's discretion.

\paragraph{Extended Task 8.} A formal verification of the correctness of the corpus splitting into training and test sets and the absence of data leakage is performed. The learner documents the splitting procedure, recording the script or notebook cell with an indication of the random seed value used for reproducibility. The absence of intersection of the sets of unique document identifiers in the training and test portions is programmatically verified; any non-empty intersection is classified as a critical error. For the final verification protocol, the following are calculated: the absolute volume of both sets in tokens, the proportion of the test set, the number and proportion of test set tokens absent from the training set (an empirical estimate of the expected OOV proportion), as well as the degree of lexical overlap of the vocabularies built on the training and full data collection. The verification protocol is included in the analytical report and serves as documentary confirmation of the methodological rigour of the experiment.

\paragraph{Extended Task 9.} A unified interface for all vectorisation methods is developed in the form of a class with \texttt{.fit()} and \texttt{.transform()} methods. The class must provide uniform processing of input data regardless of the chosen method, preservation of internal state (trained vectorizer, model parameters, document vectors), and support for serialisation --- for example, using \texttt{pickle} or \texttt{joblib}, --- so that the model can be saved to disk and reloaded without retraining. Such an approach allows any vectorisation method to be integrated into a unified machine learning pipeline, including as part of a \texttt{Pipeline} from scikit-learn, and conforms to the principles of developing code suitable for industrial operation.

\paragraph{Extended Task 10.} A suite of unit tests is developed based on the pytest framework, which verifies the correctness and reproducibility of the key project components. The tests must check: the determinism of the results with a fixed random seed value; the correctness of the dimensionality and type of the output vectors for each method; the robustness of the models to small changes in the input data (e.g., replacing one word with a synonym); and, critically, the absence of information leakage between the training and test sets. All libraries used are recorded in \texttt{requirements.txt} or \texttt{environment.yml} files with exact versions. A script is developed for the automatic execution of the full test suite and the generation of a pass/fail report. The presence of a successfully passing test suite is regarded as an integral component of the software artefact, guaranteeing its reliability and maintainability.

\paragraph{Extended Task 11.} The influence of the choice of vectorisation method on the quality of solving a real applied task is assessed. The learner conducts an experiment on the classification of news texts by category using logistic regression or a support vector machine (SVM) as the classifier, applied on top of the various vector representations obtained in the previous tasks. The comparison is performed by the metrics of accuracy, $F_1$ score (macro- and micro-averaged), and the training time of the classifier. This stage links the abstract properties of vectors --- sparsity, dimensionality, semantic richness --- with their practical applicability and makes it possible to formulate empirically substantiated recommendations for the choice of vectorisation method for text classification tasks.

\subsection{Further Research Directions (Optional)}
\textbf{First Additional Task.} This consists of constructing and comparing thematic profiles of documents obtained by different vectorisation methods. The learner selects several documents from different categories and visualises their representations in the spaces of BoW, TF-IDF, LSA, and one of the embeddings (e.g., the mean word vector of the document), analysing how well each method separates the thematic clusters.

\textbf{Second Additional Task.} This is aimed at analysing the influence of the morphological complexity of the language on the effectiveness of FastText in comparison with Word2Vec. The learner forms a list of rare and morphologically derived words, computes vectors for them using both models, and evaluates the extent to which the subword $n$-grams used by FastText improve the quality of representation of rare forms compared to Word2Vec, which relies only on full-word tokens.

\textbf{Third Additional Task.} This is devoted to the investigation of semantic drift. The learner splits the corpus into two time periods (e.g., 2020--2022 and 2023--2025) and trains embeddings separately for each period. Then, for a series of key terms (such as ``crisis'', ``technology'', ``education''), the shift in meaning is quantitatively measured by computing the cosine distance between the vectors of the same word from different time slices. The results are visualised and accompanied by a discussion of the possible sociolinguistic causes of the observed shifts.

\section{Conclusion}
This work is not reducible to the simple translation of words into numbers. It represents a dive into the semantics of language, encoded in the geometry of vector spaces. It is here that the understanding is formed that the quality of any NLP solution --- from a recommendation system to a fake news detector --- is determined by how accurately the vector reflects meaning.

In the course of performing the work, the learner learns to pose the key questions that determine the effectiveness of vector representation: whether the method preserves semantic relations between words --- this is a question about the structural adequacy of embeddings; how compactly and efficiently the information is encoded --- this is a question about the balance between dimensionality, sparsity, and expressiveness; whether the results can be trusted today and in a year's time --- this is a question about reproducibility, testing, and openness. The answers to these questions constitute the difference between simply a ``vector'' and a meaningful representation.

It is precisely this kind of systematic, critically considered, and technically rigorous approach to vectorisation that lies at the foundation of all modern NLP systems, from search engines to generative models. Upon completion of the work, the learner receives not only a set of tools for transforming text into vectors but also a methodological compass for choosing, evaluating, and adapting vectorisation methods in any future projects --- from scientific research to industrial solutions.

\chapter{Text Clustering and Semantic Similarity Analysis}
\label{work:3}

\section{Aim and Learning Outcomes}
The purpose is to explore semantic structure in collections of documents without assuming that a clustering algorithm discovers a single correct taxonomy. Students evaluate cluster stability, size balance and interpretability, distinguish internal scores from external topic labels, and learn to use qualitative inspection as a complement to numerical indices.

The three tasks below constitute the required experimental core. Larger engineering studies and independent research questions are available as optional extensions.

\section{Theoretical Background}

\noindent\textit{Principal theoretical and methodological sources for this work:} \cite{ru_reference_20}.\par

Text clustering represents one of the fundamental tasks of unsupervised learning in NLP and serves as the basis for a wide spectrum of applications: automatic thematic segmentation, exploratory data analysis, the construction of recommendation systems, and anomaly detection. Unlike classification, clustering does not use pre-assigned labels and makes it possible to discover natural groupings in semantic space that reflect the internal structure of the data. However, the absence of labels also means that any cluster solution is, to a certain extent, a construct imposed by the analyst. One and the same collection of documents, when viewed through the prism of different feature spaces or clustering algorithms, can yield completely different groupings --- each of which is mathematically substantiated, but not all are equally useful. This circumstance imposes special requirements on the evaluation strategy: it is necessary to demonstrate not simply the fact of the existence of clusters, but their stability, interpretability, and correspondence to some external quality criterion. \cite{reimers2019sentence,manning1999foundations}

Depending on the problem statement and data structure, the following main types of clustering algorithms are distinguished, differing in their mathematical foundations, computational complexity, and the nature of the obtained partitions.

The first class --- centroid-based methods --- is based on minimising the total distance from points to cluster centres. The $k$-means algorithm minimises the within-cluster sum of squared distances; it requires the \textit{a priori} specification of the number of clusters $k$, is sensitive to outliers, and assumes a spherical shape of clusters, which is far from always true for text data in sparse spaces. Its variant, $k$-medoids (PAM), is more robust to outliers, since it chooses real data points as centres, but is more computationally expensive and is less frequently applied to large text corpora.

The second class --- hierarchical methods --- builds a dendrogram by sequentially merging the closest clusters (agglomerative clustering). Such an approach allows the number of clusters to be chosen post hoc by analysing the dendrogram and supports various linkage criteria: Ward's method (minimising the increment of variance), average, complete, and single linkage. Hierarchical clustering is particularly useful for analysing the hierarchical thematic structure of news streams, where topics may have subtopics.

The third class --- density-based methods --- groups points that are in dense regions of space and automatically identifies noise. The DBSCAN algorithm does not require specifying the number of clusters, but is sensitive to the parameters $\varepsilon$ (neighbourhood radius) and \texttt{min\_samples} (minimum number of points in a neighbourhood). Its generalisation, HDBSCAN, works better with clusters of variable density and automatically determines their number, which makes it especially suitable for heterogeneous text corpora, where some topics are represented by compact groups of documents and others by sparse clouds.

The fourth class --- probabilistic methods --- assumes that the data are generated by a mixture of distributions. Gaussian Mixture Models (GMM) are appropriate for overlapping clusters and allow the estimation of the probability of a document's membership in each of the clusters, which is a valuable property for tasks of fuzzy thematic segmentation, where a single document may touch upon several topics simultaneously.

The fifth class --- modern scalable and spectral methods --- includes spectral clustering, which uses the eigenvectors of the similarity matrix and is effective for non-linearly separable clusters, but requires $O(n^2)$ memory, which limits its applicability on large corpora; and the BIRCH algorithm, optimised for large datasets by constructing a CF-tree and single-pass processing, which makes it suitable for near-real-time scenarios.

The quality of clustering depends directly on three key components. Firstly, on the correctness and volume of the corpus --- clusters reflect only what is encoded in the data. Secondly, on the method of transforming text into a feature space --- it is here that the ``language'' of the analysis is formed. Thirdly, on the strategy of evaluation and interpretation --- without a rigorous metric and linguistic assessment, clustering risks turning into ``digital astrology''.

Critically important for the choice of vectorisation method in clustering is the concept of distance. In sparse high-dimensional spaces created by TF-IDF, Euclidean distance is often uninformative: all documents turn out to be approximately equidistant from one another --- this is a manifestation of the ``curse of dimensionality''. In such spaces, cosine similarity is, as a rule, preferable, since it measures the angle between vectors rather than their magnitude, which better corresponds to the intuition that documents devoted to a single topic should be close regardless of their length. In dense embedding spaces, Euclidean distance may recover meaningful geometry; however, this depends on the method of training the embedding model. The absence of a universally optimal distance metric means that its choice must be substantiated empirically for each combination of tokenizer, vectorizer, and clustering algorithm.

The main tools for such empirical substantiation are internal and external clustering quality metrics. Internal metrics --- including Silhouette Score, Calinski--Harabasz Index, and Davies--Bouldin Index --- assess the cohesion and separation of clusters, using only the data and the computed labels, without reference to any external ground truth. They are necessary when a labelled corpus is absent; however, they are sensitive to the choice of distance metric and may favour algorithms making similar geometric assumptions. External metrics --- such as Adjusted Rand Index (ARI) and Normalized Mutual Information (NMI) --- require the presence of an approximate reference labelling (e.g., the thematic category of a news article) and measure the degree to which the discovered clusters recover this labelling. A rigorous evaluation protocol combines both types of metrics: internal metrics are used to guide hyperparameter tuning, and external metrics for assessing correspondence to a meaningful external criterion.

A special role is played by the text transformation pipeline, which includes two successive stages. At the first stage, tokenisation is performed: from simple whitespace splitting to subword models (BPE, WordPiece, Unigram), trained on the learner's own corpus during Practical Work~\ref{work:1}. These methods are especially important for morphologically rich languages, such as Russian, where a single lexeme can generate dozens of word forms. At the second stage, vectorisation is performed: classical statistical methods --- Bag-of-Words and TF-IDF with unigrams, bigrams, and trigrams --- create sparse but interpretable spaces; static embeddings trained on the learner's own corpus during Practical Work~\ref{work:2} (Word2Vec, FastText, GloVe) form dense vectors that capture semantic relations; and external pre-trained embeddings, such as fastText trained on Common Crawl or GloVe from Stanford, provide universal but potentially biased representations. Each combination of ``tokenizer + vectorizer'' forms a unique feature space, and the clustering quality may depend substantially on this choice. It is for this reason that the systematic comparative analysis of all components of the pipeline is not an optional exercise but a methodological necessity, forming the core of the present work.

\subsection{Experimental Interpretation}
There is no universal clustering quality metric that is equally valid across algorithms and applications. Silhouette and related internal indices depend on the distance measure and geometry of the representation; density-based methods and non-convex clusters require particular care. Report internal scores alongside cluster sizes, robustness to random seeds or resampling, and independently assessed semantic coherence when feasible. Ground-truth topic labels, if available, should be used for external validation rather than for unsupervised model fitting.

\section{Execution Procedure}

\subsection{Core: Three Required Tasks}
\begin{enumerate}[leftmargin=*,label=\textbf{\arabic*.}]

\item Create document representations from one training split and select two clustering configurations such as K-means with alternative normalisation or initialisation. Record the data representation and all seeds.

State what counts as a document and define the feature pipeline, including vector normalisation and dimensionality reduction if used. Use identical documents for both clustering configurations and document whether the representation was fitted solely on the intended training collection. For algorithms requiring a cluster count, explain how the candidate values were chosen without using test labels to tune an unsupervised result. Preserve the random states and a compact configuration log so that observed changes can be attributed to the clustering method rather than to altered input data.

\item For each configuration report internal indices only when their mathematical assumptions apply, along with cluster size distribution and stability across at least three random initialisations. Use labels, if available, for independent external evaluation only.

Report the number and size of clusters, the number of noise points if the algorithm supports noise, and at least three outcomes under controlled changes of random initialisation or resampling. Explain whether silhouette, Davies--Bouldin or another internal index is well-defined for the data geometry, dissimilarity function and cluster configuration being analysed. If independent class labels exist, compute a clearly labelled external statistic without allowing it to determine the unsupervised training outcome. Do not rank methods based on a single index when its assumptions are violated.

\item Inspect cluster examples and error cases, including very small or empty clusters and changes after resampling. Explain why no single metric establishes semantic usefulness of all clustering methods.

Inspect several texts from multiple clusters, especially small, heterogeneous or unstable clusters, and identify recurring terms or semantic themes only when the examined content supports them. For selected texts, compare assigned clusters across initialisations or perturbations to illustrate instability. Distinguish an empty cluster, a noise point, an outlier and a small legitimate topic. The interpretation should discuss both what the automatic indices show and what they fail to establish about topic usefulness.

\end{enumerate}

\subsection{Reproducible Example and Its Limits}

The companion project contains executable code for this work. Run the real-data path with
\begin{center}\texttt{python practikum\_examples.py 3}\end{center}
20 Newsgroups; class labels are used for external inspection, never to train clusters. For a technical-only check, add \texttt{-{}-smoke}; its synthetic inputs must not be used as scientific evaluation results. The program does not contain fabricated model scores. Record actual outputs, sample sizes, versions, timings and failures after running it.

\subsection{Extended Engineering Tasks (Optional)}
The following tasks provide a fuller engineering and research programme. They are optional rather than prerequisites for passing the core. The numerical scales and model choices below describe ambitious projects; they are not universal requirements. Documentation and reporting follow the shared protocol above.

\paragraph{Extended Task 1.} The text corpus formed during Practical Work~\ref{work:1} is used as the experimental base. The learner loads the corpus from a JSONL file, checks the integrity and completeness of the metadata (the presence of the fields \texttt{id}, \texttt{text}, \texttt{title}, \texttt{source}, \texttt{category}, \texttt{language}, \texttt{date} where available), and ensures that the corpus contains at least five thousand documents and at least five million tokens. The \texttt{category} field, if present in the metadata, is isolated from the training process: clustering is a fully unsupervised task, and the category labels are not used in constructing the partition; however, they may be employed at the stage of external quality evaluation through the ARI and NMI metrics. All texts must be pre-cleaned and normalised in accordance with the requirements of Practical Work~\ref{work:1}. References to the specific artefacts of Practical Work~\ref{work:1} and Work~\ref{work:2} (the path to the corpus file, the datasheet, the chosen tokenisation and normalisation schemes, the files of trained embedding models) are recorded in the methodological section of the report to ensure full traceability of the experiment.

\paragraph{Extended Task 2.} The generation of all relevant combinations of tokenizers and vectorizers is performed, forming a complete text preprocessing pipeline before clustering. The set of tokenizers includes: na\"ive whitespace tokenisation, lemmatization using pymorphy3 (or an analogue for the chosen language), as well as the subword models BPE, WordPiece, and Unigram, trained on the corpus during Practical Work~\ref{work:1}. The set of vectorizers includes: classical methods --- Bag-of-Words and TF-IDF with unigrams, bigrams, and trigrams and $L_2$-normalisation; own embeddings trained on the corpus during Practical Work~\ref{work:2} (Word2Vec in the CBOW and Skip-gram variants, FastText, GloVe); external pre-trained embeddings --- fastText (\texttt{cc.ru.300.vec} for the Russian language or analogous ones for other languages) and GloVe from Stanford. For embedding representations, the document vector is formed by averaging the vectors of all tokens occurring in the document, followed by $L_2$-normalisation. For each ``tokenizer + vectorizer'' combination, a meta-description is formed in JSON format, recording all transformation parameters, which ensures reproducibility and allows the automation of subsequent comparison.

\paragraph{Extended Task 3.} A configurable software module \texttt{clustering\_algorithms.py} is developed, providing a unified interface for applying and comparing eight clustering algorithms: KMeans, KMedoids, AgglomerativeClustering, DBSCAN, HDBSCAN, GaussianMixture, SpectralClustering, and Birch. The module must support working with both sparse matrices in scipy sparse CSR/CSC format (necessary for TF-IDF with large vocabularies) and dense embedding matrices. A key requirement is support for automatic hyperparameter tuning: for $k$-means and hierarchical clustering, the choice of the number of clusters $k$ is implemented based on the maximisation of the Silhouette Score, with visualisation of the curve of the dependence of the metric on $k$; for DBSCAN and HDBSCAN, a grid search over the parameters $\varepsilon$ and \texttt{min\_samples} is performed. The module code is furnished with exhaustive documentation and is designed with the prospect of extension by new algorithms.

\paragraph{Extended Task 4.} A rigorous evaluation of the quality of cluster solutions is performed for each generated ``tokenizer + vectorizer'' combination and each clustering algorithm. A system of internal metrics is applied as the main measurement tool: Silhouette Score (the main metric, measuring the compactness and separation of clusters), Calinski--Harabasz Index, and Davies--Bouldin Index. If metadata on categories (\texttt{category}) are present in the corpus, external metrics are additionally applied: Adjusted Rand Index (ARI) and Normalized Mutual Information (NMI), which quantitatively assess how well the obtained partition recovers the \textit{a priori} categorical division. For each pipeline, the optimal hyperparameters are selected, diagnostic visualisations are constructed --- the elbow method plot (dependence of inertia on $k$), silhouette diagrams, dendrograms for hierarchical clustering, --- and the results are saved in a structured format for subsequent analysis.

\paragraph{Extended Task 5.} An empirical assessment of the influence of the choice of tokenizer and vectorizer on the final clustering quality is performed. The results are presented in the form of a final summary table, the rows of which correspond to the ``tokenizer $\times$ vectorizer $\times$ clustering algorithm'' combinations, and the columns to the computed metrics. Additionally, quality heat maps are constructed, visualising the values of the key metrics (Silhouette Score, ARI) along the ``tokenizer --- vectorizer'' axes for each algorithm, and box plots are constructed, reflecting the stability of the metrics under bootstrap analysis (multiple clustering on random subsamples). On the basis of the obtained data, conclusions are formulated about the optimal pipeline configurations for various types of corpora: news texts, social media texts, and multilingual collections.

\paragraph{Extended Task 6.} An understanding of the content of each obtained cluster is ensured. For each cluster, the ten keywords with the highest TF-IDF value relative to the subcorpus formed by the documents of that cluster are computed; the most frequent unigrams, bigrams, and trigrams are analysed, making it possible to capture stable word combinations and thematic markers. On the basis of the lexical composition, each cluster is assigned a semantic label (e.g., ``sport'', ``politics'', ``economy'', ``healthcare''), which transforms abstract numerical identifiers into meaningful thematic groups. The interpretation results are formatted as a table, where for each cluster the following are indicated: the assigned label, a list of keywords, and three examples of the documents closest to the cluster centroid.

\paragraph{Extended Task 7.} An assessment of the reliability and performance of the implemented methods is carried out. A bootstrap analysis of cluster stability is performed: clustering is repeatedly carried out on random subsamples of documents (e.g., eighty per cent of the original volume), and for each pair of repetitions, the degree of label agreement is computed (e.g., via ARI between two partitions). A high mean ARI value of bootstrap repetitions indicates the stability of the solution; a low value indicates sensitivity to the sample composition and requires caution in interpretation. Additionally, the execution time and RAM consumption are measured for various data volumes (one thousand, five thousand, ten thousand documents). On the basis of the obtained results, recommendations are formulated for the choice of pipeline depending on the available computational resources and the goals of the analysis: a research scenario requiring maximum interpretability, versus an industrial scenario oriented towards speed and scalability.

\paragraph{Extended Task 8.} An interactive web application is created using Streamlit or Gradio, providing the user with the following capabilities: selection of a preprocessing pipeline (tokenizer + vectorizer) and a clustering algorithm from among those implemented; visualisation of a two-dimensional projection of documents using UMAP or $t$-SNE with interactive highlighting of clusters and display of categories (if available); viewing of keywords and examples of documents for each cluster; side-by-side parallel comparison of two clustering configurations; as well as a nearest-neighbour search for an arbitrary text entered by the user, with the determination of which of the existing clusters it belongs to. The recommended technologies are Streamlit in combination with Plotly or Bokeh to ensure interactivity of the visualisations. The application is designed in such a way that it can be used by a researcher without programming skills.

\paragraph{Extended Task 9.} A formal verification of the correctness of the data splitting and the absence of information leakage is performed. The learner documents the splitting procedure (if such was applied, e.g., for allocating a test set for external evaluation), recording the script or notebook cell with an indication of the random seed value. Special attention is paid to the fact that the category labels (\texttt{category}) are not used at any stage in constructing the clusters and are employed exclusively for computing external metrics at the evaluation stage. The verification protocol is included in the analytical report and serves as documentary confirmation of the methodological rigour of the experiment.

\paragraph{Extended Task 10.} A unified interface for all clustering methods is developed in the form of a class with \texttt{.fit()} and \texttt{.predict()} methods, supporting serialisation (via \texttt{pickle} or \texttt{joblib}) and compatibility with the scikit-learn API. The class must provide uniform processing of input data regardless of the chosen algorithm, preservation of internal state (trained model, cluster labels, parameters), and the ability to load a saved model for application to new data without retraining. Such an approach facilitates the integration of clustering methods into end-to-end NLP pipelines and supports reproducibility under a documented and stable environment.

\paragraph{Extended Task 11.} A suite of unit tests is developed based on the pytest framework, verifying the correctness and reproducibility of the key project components. The tests must check: the determinism of the results with a fixed random seed value (for algorithms supporting this property); the correctness of saving and loading cluster labels --- the coincidence of the labels assigned to documents before model serialisation and after its deserialisation; the stability of the metrics upon repeated execution; and the absence of information leakage between folds during bootstrap analysis. All libraries used are recorded in \texttt{requirements.txt} or \texttt{environment.yml} files with an indication of the exact versions. A script is developed for the automatic execution of the full test suite and the generation of a pass/fail report. The presence of a successfully passing test suite is regarded as an integral component of the software artefact, guaranteeing its reliability and maintainability.

\paragraph{Extended Task 12.} Cards for each pipeline configuration (Pipeline Card) and an updated datasheet are formed. The Pipeline Card must contain: the name and version; a list of stages with the specification of the method at each stage (tokenizer with parameters, vectorizer with dimensionality, clustering algorithm with hyperparameters); key metrics (Silhouette Score, ARI, NMI, execution time); identified limitations (e.g., the sensitivity of DBSCAN to density, the inability of $k$-means to identify clusters of non-convex shape). The updated datasheet is supplemented with information about exactly which subsets of the corpus were used for training and evaluation, and about the distribution of documents across the final clusters. All cards are included in the final report and, in the case of publication, are placed in the repository together with the artefacts.

\subsection{Further Research Directions (Optional)}
\textbf{First Additional Task.} This consists of a factorial analysis of the influence of tokenisation on clustering quality. The learner systematically compares the extent to which subword models (BPE, WordPiece, Unigram) improve clustering quality compared to lemmatization and na\"ive tokenisation, with a fixed vectorizer, and assesses the statistical significance of the observed differences.

\textbf{Second Additional Task.} This is aimed at comparing own and external embeddings. The learner ascertains which embeddings --- those trained on the learner's own corpus (Word2Vec, FastText, GloVe from Work~\ref{work:2}) or universal pre-trained ones (fastText Common Crawl) --- yield more compact and interpretable clusters, and formulates recommendations for cases where own training is impossible.

\textbf{Third Additional Task.} This is devoted to clustering on the material of low-resource languages. The learner applies the developed pipeline to a corpus in Tatar, Bashkir, or another language with limited digital resources and determines which combinations of tokenizers and vectorizers prove most effective under conditions of a small data volume.

\textbf{Fourth Additional Task.} This presupposes a comparison of methods for automatically determining the number of clusters: HDBSCAN (determining the number of clusters automatically), $k$-means with the choice of $k$ by the maximum Silhouette Score, and the simple elbow method. The stability of the determination of the ``natural'' number of clusters by each of the methods is assessed.

\textbf{Fifth Additional Task.} This consists of investigating the influence of $n$-grams in TF-IDF on the quality of thematic separation. The learner ascertains how the expansion of the feature space by bigrams and trigrams affects the separability of thematic groups in news texts.

\textbf{Sixth Additional Task.} This presupposes an assessment of the ``speed--quality'' trade-off. The learner compares fast vectorizers (FastText) and heavier ones (GloVe, Doc2Vec) from the point of view of the ``inference time -- Silhouette Score'' trade-off and formulates practical recommendations for industrial scenarios.

\textbf{Seventh Additional Task.} This is devoted to the analysis of the robustness of clustering to noise. The learner investigates how typos, spelling errors, and random distortions affect the stability of clusters under various preprocessing pipelines.

\textbf{Eighth Additional Task.} This consists of an in-depth analysis of cluster boundaries. The learner visualises two-dimensional projections of the same documents obtained using TF-IDF and various embeddings and compares how the shape and density of clusters differ depending on the nature of the feature space.

\section{Conclusion}
This work is not reducible to the simple grouping of documents by similarity. It represents an investigation of the internal semantic topography of language, encoded in vector spaces. It is here that the understanding is formed that clustering is not an algorithm, but a dialogue between the data, the method, and the interpreter.

In the course of performing the work, the learner learns to pose the key questions that determine the scientific and practical value of cluster analysis. Whether the clusters reflect real thematic or semantic groups --- this is a question about the validity of the solution. How stable the given solution is upon repeated runs or changes in the data --- this is a question about robustness. Whether it is possible to explain why document A ended up in cluster X and not in Y --- this is a question about interpretability. Whether the reproducibility of the results is guaranteed --- from the first token to the final cluster label --- this is a question about transparency and scientific integrity. The answers to these questions constitute the difference between ``clustering'' and meaningful thematic analysis.

It is precisely this kind of systematic, critically considered, and technically rigorous approach to working with texts that lies at the foundation of modern systems for media monitoring, public opinion analysis, automatic moderation, and thematic archiving. Upon completion of the work, the learner receives not only a functional tool for grouping texts but also a methodological framework for designing, evaluating, and justifiably trusting any solutions in the field of unsupervised NLP --- from academic research to industrial systems.

\chapter{Text Classification with Classical Machine Learning}
\label{work:4}

\section{Aim and Learning Outcomes}
This work develops a scientifically controlled text-classification workflow. Learners set up labelled data, majority and linear baselines, appropriate preprocessing and a protected test partition, then interpret confusion matrices and performance differences. The methodological emphasis is on generalisation and error analysis, not only on achieving a high score.

The three tasks below constitute the required experimental core. Larger engineering studies and independent research questions are available as optional extensions.

\section{Theoretical Background}

\noindent\textit{Principal theoretical and methodological sources for this work:} \cite{ru_reference_21,peters2018deep,ru_reference_24,ru_reference_25,guo2017calibration}.\par

Text classification is one of the fundamental tasks of machine learning in the field of NLP and serves as the basis for a wide spectrum of applications: sentiment analysis, thematic categorisation, spam filtering, genre identification, and many others. Unlike clustering, where the grouping is performed without pre-assigned labels, classification relies on labelled data and aims at constructing a model capable of generalising to unseen examples. Depending on the problem statement, three main types of classification are distinguished, differing in the structure of the target variable and, accordingly, in the mathematical formulation and evaluation methods. \cite{manning1999foundations,guo2017calibration}

The first type --- binary classification --- presupposes a choice between two mutually exclusive labels, for example, a ``positive'' or ``negative'' review. Mathematically, such a problem is formulated as the estimation of the probability of belonging to the positive class, and the key metrics are $F_1$, ROC-AUC, and PR-AUC, the latter being especially important under strong class imbalance.

The second type --- multi-class classification --- requires the choice of one label from a finite set, for example, determining the topic of a news item: ``politics'', ``economy'', ``sport'', ``culture''. Here, the model must not only separate one class from all the rest but also distinguish the classes among themselves, which makes the analysis of the confusion matrix critically important for identifying systematically confused pairs.

The third type --- multi-label classification --- presupposes the simultaneous assignment of several labels to a document, which is particularly relevant for news texts covering several topics at once. In this case, the standard metrics require adaptation: $F_1$ is computed in the micro, macro, and samples variants, and Hamming Loss is also applied, measuring the proportion of incorrectly predicted labels.

In modern practice, classical machine learning methods are widely applied to solve the enumerated tasks; these work with a fixed feature space and do not require the training of neural networks with millions of parameters. These methods are divided into two large families.

The first family --- linear models --- includes logistic regression, which is a probabilistic model with the possibility of $L_1$ and $L_2$ regularisation, providing high interpretability and resistance to overfitting in sparse spaces of large vocabularies; and the support vector machine (SVM), which is a powerful tool for linear classification in sparse spaces, especially effective when using TF-IDF. Linear models are transparent: the feature weights directly indicate which words make the greatest contribution to one or another prediction, which makes them preferable in scenarios where interpretability is more important than ultimate quality.

The second family --- ensemble methods based on decision trees --- includes random forest, which is an ensemble robust to noise and outliers and not requiring feature scaling, which makes it a good baseline choice for tasks with heterogeneous data; and gradient boosting (CatBoost, XGBoost) --- high-performance algorithms that achieve competitive quality even under complex non-linear dependencies. Modern implementations of gradient boosting include built-in handling of categorical features and efficiently utilise parallel computations, which makes them suitable for industrial application.

The quality of the enumerated models depends directly on three key components, each of which can become a source of systematic errors. The first component is the correctness of labelling and class balance: imbalance can lead to inflated accuracy and reduced sensitivity to minority classes, when the model simply ``memorises'' the majority. The second component is the method of transforming text into a feature space: from the choice of tokenizer to the type of vectorisation, wherein different combinations may prove optimal for different algorithms. The third component is the strategy for tuning hyperparameters and assessing generalisation ability: without stratified cross-validation, the estimate risks being unjustifiably optimistic, and without bias/variance analysis, it is impossible to diagnose underfitting or overfitting.

These three components are not independent. A carefully tuned model, evaluated on incorrectly split data, will yield misleading conclusions; an ideally balanced dataset, passed through an unsuitable vectorisation pipeline, will show mediocre results. It is precisely the simultaneous attention to all three components --- labelling, representation, and evaluation --- that distinguishes a methodologically correct classification experiment from a superficial one.

Unlike regression tasks, classification does not use the concept of residuals in the traditional sense; however, error analysis is critically important. A single aggregated metric, such as accuracy or $F_1$, conceals as much as it reveals, especially under conditions of class imbalance, when a model can achieve high accuracy simply by predicting the majority class. Rigorous evaluation therefore investigates not only overall performance but also per-class metrics, confusion matrices that reveal systematically confused pairs of categories, and the calibration of predicted probabilities. The identification of systematic biases, the assessment of the quality of probabilities through reliability diagrams and Expected Calibration Error, as well as the interpretation of the causes of incorrect predictions with the aid of tools such as SHAP, LIME, and feature weight analysis, are necessary components of conscientious analysis. Only such a comprehensive approach makes it possible to transition from a ``working model'' to a reliable and trustworthy system.

\subsection{Experimental Interpretation}
To prevent data leakage, fit cleaning dictionaries, vectorisers, feature selectors, scalers and models within each training fold, not on the full dataset before splitting. Never repeatedly tune decisions against the same purportedly final test set. Macro-F1 is informative for imbalanced multi-class tasks but should be presented with class supports and a confusion matrix.

\section{Execution Procedure}

\subsection{Core: Three Required Tasks}
\begin{enumerate}[leftmargin=*,label=\textbf{\arabic*.}]

\item Build a majority-class baseline and two classical text classifiers with a train-only preprocessing pipeline. Specify the dataset split, label mapping and class balance.

Identify the target labels, class distribution and criteria for excluding missing or duplicated documents. Preserve one final test partition and create a validation procedure within the remaining data. Build the majority-class baseline and two named methods, for example multinomial naive Bayes and a linear support-vector classifier, using the same text preprocessing decisions. Fit vectorisation and any feature selection inside a pipeline rather than on the complete dataset, and record class weighting and random-state settings where applicable.

\item Compare macro-F1 and confusion matrices on the same untouched test set; select hyperparameters only with training/validation data.

Use the same final examples for all models. Report macro-F1 together with per-class precision, recall and support, and show a confusion matrix with explicit label ordering. If numerical performance is compared, report the exact dataset size and whether any validation-based model choice occurred before final testing. Include the computational cost of training when measured under comparable conditions. A modest performance difference on a small test set should not be described as a universal advantage without discussing variability.

\item Inspect at least ten misclassifications, report class-wise performance and discuss leakage risks, distribution shift and uncertainty from sample size.

Select misclassified documents from several classes rather than a single easy-to-interpret category, and record the predicted and reference labels. Distinguish label ambiguity, distribution shift, preprocessing losses and genuine algorithmic failure when evidence permits. Examine whether duplicates, source metadata or corpus artefacts could have leaked information across partitions. The final comparison should identify practical trade-offs, not merely present the model with the highest score.

\end{enumerate}

\subsection{Reproducible Example and Its Limits}

The companion project contains executable code for this work. Run the real-data path with
\begin{center}\texttt{python practikum\_examples.py 4}\end{center}
20 Newsgroups; train/test classification with a documented baseline. For a technical-only check, add \texttt{-{}-smoke}; its synthetic inputs must not be used as scientific evaluation results. The program does not contain fabricated model scores. Record actual outputs, sample sizes, versions, timings and failures after running it.

\subsection{Extended Engineering Tasks (Optional)}
The following tasks provide a fuller engineering and research programme. They are optional rather than prerequisites for passing the core. The numerical scales and model choices below describe ambitious projects; they are not universal requirements. Documentation and reporting follow the shared protocol above.

\paragraph{Extended Task 1.} On the basis of the corpus formed during Practical Work~\ref{work:1}, a labelled dataset is prepared for three types of classification tasks. For binary classification, the sentiment of the document is determined: the labels ``positive'' and ``negative'' may be obtained automatically using sentiment lexicons (e.g., RuSentiLex for the Russian language), semi-automatically by analysing headings containing explicit evaluative markers, or expertly on the basis of manual annotation. For multi-class classification, the \texttt{category} field of the corpus is used, which contains the thematic category of the document (``politics'', ``economy'', ``sport'', ``culture'', and so on). For multi-label classification, each document may be assigned to several topics simultaneously; such labelling may be performed on the basis of keywords characterising each topic, using automatic annotation (e.g., with the aid of a pre-trained classifier or a support vector machine in one-vs-rest mode) or expert analysis.

The minimum total volume of labelled data is ten thousand documents. If the labelling of all three types of tasks in full is objectively difficult --- for example, due to the high cost of expert annotation or the limited volume of the source corpus, --- the learner has the right, in agreement with the instructor, to reduce the number of labelled documents or to confine themselves to one or two types of tasks, explicitly stipulating this circumstance in the datasheet and discussing its influence on the statistical significance and generalisation ability of the obtained models in the analytical report.

Each document is saved as a JSON object with the following fields: \texttt{id} --- unique document identifier; \texttt{title} --- heading; \texttt{text} --- full text after cleaning; \texttt{sentiment} --- sentiment label (for the binary task, values ``positive'' / ``negative''); \texttt{category} --- thematic category (for the multi-class task); \texttt{categories} --- list of thematic labels (for the multi-label task). The data are saved in JSONL format with UTF-8 encoding.

The split into training, validation, and test sets is performed in a $70/15/15$ ratio. To ensure the representativeness of all classes in each of the subsets, a stratified split is applied with respect to the target variable: for the binary task --- with respect to the \texttt{sentiment} field; for the multi-class task --- with respect to the \texttt{category} field; for the multi-label task --- using iterative stratification (e.g., the \texttt{iterative\_train\_test\_split} algorithm from the \texttt{skmultilearn} library), which takes into account the joint distribution of labels and aims to preserve the proportions of all category combinations. The random seed value used in the split is fixed and documented to ensure full reproducibility.

All texts are pre-cleaned and normalised in accordance with the requirements of Practical Work~\ref{work:1}. References to the specific artefacts of the previous works (the path to the corpus file, the datasheet, the chosen tokenisation and normalisation schemes) are recorded in the methodological section of the report to ensure full traceability of the experiment. If the learner performs the present work independently of the previous ones, they must first create or select a corpus satisfying the requirements of Works~\ref{work:1} and Work~\ref{work:2} and explicitly indicate its origin and characteristics in the datasheet.

\paragraph{Extended Task 2.} Several independent feature spaces are formed by combining the tokenisation and vectorisation methods investigated in the previous works. The set of tokenizers includes: na\"ive whitespace tokenisation, lemmatization using pymorphy3 (or an analogue for the chosen language), as well as the subword models BPE, WordPiece, and Unigram, trained on the corpus during Practical Work~\ref{work:1}. The set of vectorizers includes: TF-IDF with unigrams, bigrams, and trigrams and $L_2$-normalisation; own embeddings trained on the corpus during Practical Work~\ref{work:2} (Word2Vec in the CBOW and Skip-gram variants, FastText, GloVe); and external pre-trained embeddings, such as fastText Common Crawl (\texttt{cc.ru.300.vec} for the Russian language) and GloVe from Stanford. For embedding representations, the document vector is formed by averaging the vectors of all tokens occurring in the document, followed by $L_2$-normalisation. The metadata for each configuration are saved in JSON format, which ensures reproducibility and the possibility of automated search during subsequent comparison.

\paragraph{Extended Task 3.} A configurable software module \texttt{classical\_classifiers.py} is developed, providing a unified interface for training and applying the following models: LogisticRegression, LinearSVC (linear SVM), RandomForest, XGBoost, and CatBoost. The module must support working with both sparse matrices in scipy sparse CSR/CSC format (necessary for TF-IDF) and dense embedding matrices. For SVM, when working with sparse TF-IDF representations, only the linear kernel (LinearSVC) is used; when working with dense embeddings, the investigation of linear, RBF, and polynomial kernels is permitted. The module code is furnished with exhaustive documentation, including a description of all parameters and methods, and is designed with the prospect of extension by new algorithms.

\paragraph{Extended Task 4.} Rigorous tuning of the models is performed using stratified five-fold cross-validation. The target metrics are: $F_1$ (macro) --- for binary and multi-class classification, $F_1$ (samples) --- for multi-label classification. The search for hyperparameters is carried out using Grid Search for linear models and Random Search or Optuna for boosting models, which makes it possible to efficiently explore the parameter space without an exhaustive search. For each model and each ``tokenizer + vectorizer'' combination, the optimal hyperparameters are selected, and the results are saved in a structured manner. The mandatory visualisations are heat maps of hyperparameters, learning curves showing the dynamics of the metrics as the volume of the training set increases, and convergence plots for boosting models. All configurations and metrics are saved in JSON format for subsequent reproduction and analysis.

\paragraph{Extended Task 5.} An empirical assessment of the effectiveness of the models and preprocessing pipelines is performed. The following metrics are used: Accuracy, Precision, Recall, $F_1$ (in the macro, micro, and weighted variants for multi-class), ROC-AUC (in the one-vs-rest variant for multi-class), PR-AUC, as well as Hamming Loss for multi-label classification. The results are presented in the form of a final summary table \texttt{classification\_metrics.csv}, the rows of which correspond to the ``tokenizer $\times$ vectorizer $\times$ model'' combinations, and the columns to the computed metrics. Additionally, ROC and PR curves are constructed for each model, as well as box plots of the metrics across the cross-validation folds, making it possible to assess not only the average quality but also its stability.

\paragraph{Extended Task 6.} The robustness of the models under conditions of an uneven distribution of labels is ensured. For this, the mechanisms for accounting for class imbalance built into many models are applied (e.g., the parameter \texttt{class\_weight='balanced'} in LogisticRegression, SVM, and RandomForest). Where necessary, the method of synthetic generation of examples SMOTE is additionally used, which is applied exclusively to the training folds and does not affect the validation and test data, in order to avoid information leakage. The quality assessment under conditions of imbalance is performed with reliance on the metrics PR-AUC and $F_1$, which more adequately reflect the classification quality of minority classes than accuracy. The learner may explicitly record which imbalance correction strategies were applied and compare the metrics before and after their use.

\paragraph{Extended Task 7.} An in-depth analysis of the causes of incorrect predictions and the quality of the probabilistic outputs of the models is performed. Reliability diagrams are constructed, visualising how well the probability predicted by the model corresponds to the actual frequency of correct answers, and the Expected Calibration Error (ECE) is computed. The confusion matrix is analysed to identify systematically confused pairs of classes --- for example, a model may regularly mix up ``economy'' and ``politics'' due to the overlap of thematic vocabulary. Documents with high model confidence but an erroneous prediction are selected, and their manual analysis is performed to identify possible causes of the errors. The interpretation of predictions is carried out using three complementary approaches: SHAP (SHapley Additive Explanations), which makes it possible to assess the contribution of each feature to a specific prediction; LIME (Local Interpretable Model-agnostic Explanations), which builds a locally approximating model around an individual example; and the analysis of feature weights for linear models, directly indicating the most influential words and $n$-grams. The visualisation of token contributions is performed using waterfall and force diagrams of SHAP, as well as textual explanations of LIME.

\paragraph{Extended Task 8.} A formal verification of the correctness of the data splitting and the absence of information leakage between the sets is performed. The learner documents the splitting procedure, recording the script or notebook cell with an indication of the random seed value used for reproducibility. The absence of intersection of the sets of unique document identifiers in the training, validation, and test sets is programmatically verified; any non-empty intersection is classified as a critical error. It is separately verified that the synthetic SMOTE examples (if applied) did not end up in the validation and test sets. The verification protocol is included in the analytical report and serves as documentary confirmation of the methodological rigour of the experiment.

\paragraph{Extended Task 9.} A unified interface for all implemented classifiers is developed in the form of a class with \texttt{.fit()} and \texttt{.predict\_proba()} methods, compatible with the scikit-learn API. The class must provide uniform processing of input data regardless of the chosen model, preservation of internal state (trained model, parameters, class labels), and support for serialisation via \texttt{joblib} for the possibility of saving to disk and reloading without retraining. Such an approach facilitates the integration of classifiers into end-to-end NLP pipelines and supports reproducibility under a documented and stable environment.

\paragraph{Extended Task 10.} A suite of unit tests is developed based on the pytest framework, verifying the correctness and reproducibility of the key project components. The tests must check: the determinism of the results with a fixed random seed value (for algorithms guaranteeing this property); the correctness of the stratified split --- the preservation of class proportions in the training and test sets; the absence of data leakage between the training and test sets, verifiable through the intersection of sets of identifiers; and the stability of the metrics upon repeated execution. All libraries used are recorded in \texttt{requirements.txt} or \texttt{environment.yml} files with an indication of the exact versions. A script is developed for the automatic execution of the full test suite and the generation of a pass/fail report. The presence of a successfully passing test suite is regarded as an integral component of the software artefact.

\paragraph{Extended Task 11.} An interactive web application is created using Streamlit or Gradio, providing the user with the following capabilities: entry of arbitrary text; selection of the task type (binary, multi-class, multi-label), the preprocessing pipeline, and the trained model; display of the prediction with an indication of the probabilities for all classes and with the visualisation of the interpretation (word contributions via SHAP or LIME); side-by-side parallel comparison of two models; and automatic generation of an HTML report with the results of the analysis. The recommended technologies are Streamlit in combination with Plotly or Bokeh to ensure interactivity of the visualisations. The application is designed in such a way that it can be used by a specialist without programming skills and must be deployed locally with the possibility of subsequent deployment to a cloud platform at the learner's discretion.

\subsection{Further Research Directions (Optional)}
\textbf{First Additional Task.} This is devoted to comparing ``own'' and ``external'' embeddings. The learner ascertains how the origin of the embeddings (trained on the learner's own corpus in Work~\ref{work:2} versus universal pre-trained ones) affects the quality and generalisation ability of the classifiers.

\textbf{Second Additional Task.} This is aimed at analysing the influence of subword tokenisation on classification quality. The learner determines whether BPE, WordPiece, and Unigram offer advantages in tasks with high morphological diversity, especially for rare and derived words.

\textbf{Third Additional Task.} This consists of using the weights of logistic regression to identify lexical markers of sentiment. The learner determines which words are most strongly associated with positive and negative sentiment and visualises them in the form of horizontal bar charts.

\textbf{Fourth Additional Task.} This presupposes an analysis of the robustness of classifiers to noise. The learner investigates how typos, spelling errors, and slang expressions affect the accuracy of predictions under various preprocessing pipelines.

\textbf{Fifth Additional Task.} This consists of the visualisation of the hyperparameter space: the construction of two-dimensional and three-dimensional projections of the dependence of $F_1$ on the model parameters, which makes it possible to intuitively assess the landscape of the optimisation problem.

\textbf{Sixth Additional Task.} This is devoted to comparing hyperparameter search strategies: Grid Search, Random Search, and Optuna --- from the point of view of efficiency according to the ``quality / tuning time'' criterion.

\textbf{Seventh Additional Task.} This presupposes an assessment of the influence of probability calibration on decision making. The learner ascertains how the application of Platt scaling and isotonic regression affects the Expected Calibration Error and the model's confidence under conditions where the cost of an error is high.

\textbf{Eighth Additional Task.} This consists of an in-depth analysis of SVM kernels. The learner conducts a systematic comparison of linear, radial (RBF), and polynomial kernels in binary and multi-class classification tasks, investigating how the type of vector representation (sparse TF-IDF versus dense averaged embeddings) affects the quality, computational complexity, resistance to overfitting, and interpretability of the separating surface. For an intuitive understanding of the behaviour of the models, it is recommended to visualise the separating surfaces on reduced projections obtained using PCA or $t$-SNE.

\section{Conclusion}
This work is not reducible to the simple training of a model to predict labels. It represents the construction of a trustworthy decision-making system based on text. It is here that the understanding is formed that the quality of classification is not only the $F_1$-score but also transparency, reliability, and conscientiousness.

In the course of performing the work, the learner learns to pose the key questions that determine the applied and ethical value of the model. Whether the model predicts correctly or is simply guessing on balanced data --- this is a question about the rigour of evaluation. Why precisely this answer --- this is a question about interpretability and user trust. To what extent one can rely on its confidence --- this is a question about the calibration of probabilities. Whether the reproducibility of the results is guaranteed today and in a year's time --- this is a question about transparency and openness. The answers to these questions constitute the difference between a ``model'' and a responsible intelligent system.

It is precisely this kind of systematic, critically considered, and technically rigorous approach to classification that lies at the foundation of modern systems for moderation, user experience analysis, automatic diagnostics, and other socially significant applications. Upon completion of the work, the learner receives not only a set of trained classifiers but also a methodological foundation for designing, evaluating, and deploying any reliable NLP systems --- from research prototypes to industrial solutions.

\chapter{Automated Model Selection for Text Classification}
\label{work:5}

\section{Aim and Learning Outcomes}
Students investigate systematic model selection under a finite search budget and learn to compare automated search with an explicit manual baseline. The work distinguishes training-fold validation from final evaluation, addresses the computational and interpretability costs of search, and identifies which capabilities belong to full AutoML platforms as opposed to limited hyperparameter optimisation.

The three tasks below constitute the required experimental core. Larger engineering studies and independent research questions are available as optional extensions.

\section{Theoretical Background}

\noindent\textit{Principal theoretical and methodological sources for this work:} \cite{ru_reference_26,erickson2020autogluon,wang2023flaml}.\par

Automated Machine Learning (AutoML) aims to minimise manual intervention in the process of building ML models, including the choice of algorithms, data preprocessing, hyperparameter tuning, ensembling, and feature selection. In text classification tasks, modern AutoML systems demonstrate a fundamentally different level of support for unstructured text, which directly affects their applicability to morphologically rich languages, such as Russian. \cite{erickson2020autogluon,wang2023flaml}

Three main approaches to applying AutoML in text tasks are distinguished, differing in the degree of automation and the required level of preprocessing effort.

The first approach --- end-to-end AutoML with native text support --- is represented by systems that accept raw text without manual vectorisation and automatically apply transformers, convolutional networks, or static embeddings. A typical example is AutoGluon (TextPredictor), which uses pre-trained language models (e.g., RuBERT for the Russian language) and automatically adapts them to the data by fine-tuning or feature extraction, achieving high quality without user intervention. Such an approach relieves the analyst of the burden of choosing a feature space but is demanding of computational resources and the quality of the pre-trained model.

The second approach --- AutoML on top of feature representations --- unites frameworks that require the preliminary transformation of text into numerical features (e.g., TF-IDF or embeddings) but fully automate model selection and hyperparameter tuning. Examples are FLAML and H2O AutoML. These systems are effective under conditions of limited resources and work well with sparse representations; however, they are sensitive to the quality of the input representation: if the tokenisation or vectorisation is performed inadequately, the automatic selection of models will not compensate for the loss of information.

The third approach --- hybrid --- presupposes a combination of automatic optimisation and manual control of key stages: for example, the choice of a tokenizer from among those trained in Practical Work~\ref{work:1}, the substitution of own embeddings from Practical Work~\ref{work:2}, or the targeted construction of features specific to the subject domain. The hybrid approach makes it possible to achieve a balance between quality, speed, and interpretability, using the strengths of both automation and expert knowledge.

The quality of AutoML solutions depends on four key factors. Firstly, on the volume and quality of labelling: AutoML does not substitute for data, and no automation is capable of extracting signal from noise. Secondly, on the language support of the framework: many systems are optimised for the English language and may ignore Russian morphology if they are not equipped with corresponding pre-trained models or tokenizers. Thirdly, on the text preprocessing strategy: even in end-to-end systems, preliminary cleaning can substantially affect the result. Fourthly, on the allocated computational budget: quality, as a rule, grows with increasing search time, but with saturation, which requires a conscious choice of the ``quality--speed'' trade-off.

Alongside their advantages, AutoML systems possess noticeable limitations. They tend to generate complex ensembles whose decision-making logic is difficult to survey, which reduces interpretability compared to individually manually tuned models. The automatic search procedure, if not constrained, may over-optimise to the peculiarities of the validation set, generating models incapable of generalisation. Furthermore, the computational costs of the search itself can be significant, and the carbon footprint of a lengthy AutoML run is capable of exceeding the benefit from a negligible increase in accuracy relative to a well-tuned baseline solution. For these reasons, AutoML should be regarded not as a replacement for expert judgement but as an accelerator: it quickly establishes a competitive baseline, freeing the practitioner's resources for solving substantive tasks --- interpretability, fairness, and consideration of deployment constraints --- which no automated system is capable of solving. Awareness of these trade-offs constitutes the methodological core of the present work.

\subsection{Experimental Interpretation}
An automated hyperparameter search is not automatically equivalent to an end-to-end AutoML system. Compare configurations under a fixed search budget, record stochastic seeds, and distinguish cross-validation scores used for model selection from the final held-out evaluation.

\section{Execution Procedure}

\subsection{Core: Three Required Tasks}
\begin{enumerate}[leftmargin=*,label=\textbf{\arabic*.}]

\item Specify an automated selection/search space with a fixed computational budget and compare it with one non-optimised baseline.

Specify the supervised task, permitted feature extraction and algorithm families, a baseline pipeline, and a bounded search budget expressed in trials, time or compute. Define the parameter ranges and their justification before inspecting final-test results. Explain whether the procedure searches hyperparameters for a fixed estimator or also selects preprocessing and estimator families; these are different scopes of automation. Use a consistent evaluation metric and freeze the test set until the end.

\item Run cross-validation or a held-out validation search entirely within training data, recording all candidates and selection criteria; evaluate the selected pipeline once on test.

Run stratified cross-validation when appropriate, or a clearly identified validation split, with each trial including all transformations that learn from data. Prevent leakage from duplicate or grouped texts by selecting the corresponding grouped split strategy when needed. Record failed trials and search costs, not just the winning configuration. Keep the baseline under a fair evaluation protocol and never use the final test score as feedback for another round of search.

\item Report quality, training time and the best configuration. Clarify whether the implementation is full AutoML or a limited automated hyperparameter search.

Show the chosen configuration, its validation score distribution, the number of evaluated candidates and resource use. Then evaluate both the winning model and the baseline on the untouched test set, using sample counts and class-aware metrics. Discuss the risk of overfitting to cross-validation and any discrepancy between validation and test results. Explicitly name the software mechanism and its search scope rather than claiming to have reproduced a full AutoML framework if only grid or random search was performed.

\end{enumerate}

\subsection{Reproducible Example and Its Limits}

The companion project contains executable code for this work. Run the real-data path with
\begin{center}\texttt{python practikum\_examples.py 5}\end{center}
20 Newsgroups; constrained grid search is an automated-search baseline, not a full AutoML platform. For a technical-only check, add \texttt{-{}-smoke}; its synthetic inputs must not be used as scientific evaluation results. The program does not contain fabricated model scores. Record actual outputs, sample sizes, versions, timings and failures after running it.

\subsection{Extended Engineering Tasks (Optional)}
The following tasks provide a fuller engineering and research programme. They are optional rather than prerequisites for passing the core. The numerical scales and model choices below describe ambitious projects; they are not universal requirements. Documentation and reporting follow the shared protocol above.

\paragraph{Extended Task 1.} On the basis of the corpus formed during Practical Work~\ref{work:1}, a labelled dataset is prepared for two types of classification tasks: binary (sentiment) and multi-class (topic). For binary classification, the sentiment of the document is determined: the labels ``positive'' and ``negative'' may be obtained automatically using sentiment lexicons (e.g., RuSentiLex for the Russian language), semi-automatically by analysing headings containing explicit evaluative markers, or expertly on the basis of manual annotation. For multi-class classification, the \texttt{category} field of the corpus is used, which contains the thematic category of the document (``politics'', ``economy'', ``sport'', ``culture'', and so on).

The minimum total volume of labelled data is five thousand documents. If the labelling of both types of tasks in full is objectively difficult --- for example, due to the high cost of expert annotation or the limited volume of the source corpus, --- the learner has the right, in agreement with the instructor, to reduce the number of labelled documents or to confine themselves to one type of task, explicitly stipulating this circumstance in the datasheet and discussing its influence on the statistical significance and generalisation ability of the obtained models in the analytical report.

Each document is saved as a JSON object with the following fields: \texttt{id} --- unique document identifier; \texttt{text} --- full text after cleaning; \texttt{title} --- heading (optional); \texttt{sentiment} --- sentiment label (for the binary task, values ``positive'' / ``negative''); \texttt{category} --- thematic category (for the multi-class task). The data are saved in JSONL format with UTF-8 encoding.

The split into training, validation, and test sets is performed in a $70/15/15$ ratio. To ensure the representativeness of all classes in each of the subsets, a stratified split is applied with respect to the target variable: for the binary task --- with respect to the \texttt{sentiment} field; for the multi-class task --- with respect to the \texttt{category} field. The random seed value used in the split is fixed and documented to ensure full reproducibility.

All texts are pre-cleaned and normalised in accordance with the requirements of Practical Work~\ref{work:1}. References to the specific artefacts of the previous works (the path to the corpus file, the datasheet, the chosen tokenisation and normalisation schemes) are recorded in the methodological section of the report to ensure full traceability of the experiment. If the learner performs the present work independently of the previous ones, they must first create or select a corpus satisfying the requirements of Works~\ref{work:1} and Work~\ref{work:2} and explicitly indicate its origin and characteristics in the datasheet.

\paragraph{Extended Task 2.} The preparation of input data is performed in formats compatible with each of the three AutoML frameworks under investigation. For AutoGluon (TextPredictor), raw text is prepared without any vectorisation --- the framework independently performs tokenisation and feature extraction using pre-trained language models. For FLAML and H2O AutoML, the text is preliminarily transformed into numerical representations: the main variant is TF-IDF with unigrams and bigrams, providing a sparse but interpretable feature space. Additionally, the influence of various preprocessing strategies mastered in Practical Work~\ref{work:1} is investigated: classification quality is compared when using raw text, text that has undergone lemmatization using pymorphy3 (or an analogue for the chosen language), as well as text tokenised by the subword models BPE, WordPiece, and Unigram, trained on the corpus during Practical Work~\ref{work:1}. All prepared data are saved in CSV format with the columns \texttt{text} and \texttt{label} to unify feeding into all frameworks.

\paragraph{Extended Task 3.} A configurable software module \texttt{automl\_classifiers.py} is developed, implementing a unified pipeline for training models using three AutoML systems: AutoGluon (TextPredictor), FLAML, and H2O AutoML. Training is conducted under constraints uniform for all frameworks: the computational budget is ten minutes per task (binary and multi-class), and the RAM limit is no more than eight gigabytes. For each framework and each task, the training procedure is logged, and the obtained models are saved in serialised form (using \texttt{pickle}, \texttt{joblib}, or the native means of the framework) with the mandatory indication of library versions. The module code is furnished with exhaustive documentation describing the interface of all public functions and configuration parameters, which guarantees its reusability and testability.

\paragraph{Extended Task 4.} A rigorous assessment of the generalisation ability of the trained AutoML models is performed. For each framework and each preprocessing combination, stratified five-fold cross-validation is applied on the training portion of the corpus. The target metrics are $F_1$ (macro), Accuracy, and ROC-AUC (for the binary task --- standard; for the multi-class task --- in the one-vs-rest variant). For AutoGluon, the built-in evaluation method is additionally used with the same metrics. The results across all folds are averaged, and the standard deviation is computed as an indicator of the stability of quality. The final values are saved in a summary table \texttt{automl\_text\_metrics.csv}, where the rows correspond to the ``framework $\times$ task $\times$ preprocessing type'' combinations, and the columns to the metrics and their standard deviations.

\paragraph{Extended Task 5.} An empirical assessment of the effectiveness of the AutoML systems and the influence of preprocessing on classification quality is performed. The comparison is carried out according to the following criteria: classification quality ($F_1$, Accuracy); training and inference time; requirements for text preprocessing (ability to work with raw text or necessity of manual vectorisation); support for the Russian language and accounting for morphology; robustness to noise and class imbalance. The results are presented in the form of a final summary table, as well as in the form of graphs of the dependence of $F_1$ on training time for each framework and box plots of the metrics across the cross-validation folds. On the basis of the obtained data, conclusions are formulated about the optimal choice of framework depending on the nature of the task and the available resources.

\paragraph{Extended Task 6.} The causes of incorrect predictions are investigated, and the interpretability of the models obtained by different frameworks is compared. The confusion matrix is constructed and analysed to identify systematically confused pairs of classes. Documents for which the model produced high confidence with an erroneous prediction are selected, and their qualitative analysis is performed. For models based on TF-IDF (FLAML, H2O), interpretation is applied through the analysis of feature weights --- the words and $n$-grams that made the greatest contribution to assigning the document to one or another class are determined. For AutoGluon, where technically possible, built-in methods for visualising the influence of tokens are used (e.g., via attention weights or gradient methods). The results of the interpretation are recorded in the form of tables and screenshots with extensive commentary.

\paragraph{Extended Task 7.} A formal verification of the correctness of the data splitting and the absence of information leakage between the training, validation, and test sets is performed. The learner documents the splitting procedure, recording the script or notebook cell with an indication of the random seed value. The absence of intersections of the sets of unique document identifiers in the three sets is programmatically verified; any non-empty intersection is classified as a critical error. It is additionally verified that synthetic examples, if any were applied for imbalance correction, did not end up in the validation and test portions. The verification protocol is included in the analytical report and serves as documentary confirmation of the methodological rigour of the experiment.

\paragraph{Extended Task 8.} A unified interface for all trained AutoML models is developed in the form of a class with \texttt{.fit()} and \texttt{.predict\_proba()} methods, encapsulating the internal logic of each framework and providing a uniform API for loading, training, and inference. The class must ensure format compatibility, support for serialisation (via \texttt{joblib} or the native means of the framework), and the possibility of repeated reproduction of results with fixed seeds and library versions. Such an approach simplifies the comparison of models, their integration into industrial pipelines, and deployment.

\paragraph{Extended Task 9.} An interactive web application is created using Streamlit or Gradio, providing the user with the following capabilities: entry of arbitrary text; selection of the task type (binary or multi-class); display of the prediction and probabilities for all three AutoML systems simultaneously; side-by-side parallel comparison of predictions; and automatic generation of a brief HTML report containing the labels and probabilities. The recommended technologies are Streamlit in combination with Plotly or Bokeh to ensure interactivity. The application is designed in such a way that it can be used by a specialist without programming skills.

\paragraph{Extended Task 10.} A suite of unit tests is developed based on the pytest framework, verifying the correctness and reproducibility of the key project components. The tests must check: the fixation of the seed and library versions; the correctness of the split into training, validation, and test sets; the stability of the metrics upon repeated execution of training and evaluation; and the absence of data leakage between the sets. All libraries used are recorded in \texttt{requirements.txt} or \texttt{environment.yml} files with an indication of the exact versions. A script is developed for the automatic execution of the full test suite and the generation of a pass/fail report. The presence of a successfully passing test suite is regarded as an integral component of the software artefact, guaranteeing its reliability and maintainability.

\paragraph{Extended Task 11.} The integration of the obtained results into the ecosystem of the previous works (Work~\ref{work:1}--4) is carried out. The learner traces and documents the end-to-end methodology: from tokenisation (Work~\ref{work:1}) and vectorisation (Work~\ref{work:2}) --- to clustering (Work~\ref{work:3}) and classification by classical methods (Work~\ref{work:4}), and, finally, to automated classification by AutoML means. The report provides a comparative table including the metrics of the best classical models from Work~\ref{work:4} and the best AutoML models obtained in the present work, with a discussion of the gain in quality, development time, and interpretability. Such an analysis transforms disjoint works into a holistic research programme.

\subsection{Further Research Directions (Optional)}
\textbf{First Additional Task.} This is devoted to comparing preprocessing strategies. The learner quantitatively assesses the extent to which the use of own tokenizers from Practical Work~\ref{work:1} (BPE, WordPiece, Unigram) improves the classification quality of the FLAML and H2O frameworks compared to feeding raw text, and formulates recommendations for cases where end-to-end AutoML systems are unavailable.

\textbf{Second Additional Task.} This presupposes an analysis of the dependence of quality on the computational budget. The learner constructs plots of $F_1$ as a function of training time (1, 5, 10, 20 minutes) for each of the three frameworks and determines the saturation point, after which an increase in the budget does not lead to a significant gain in quality.

\textbf{Third Additional Task.} This is aimed at assessing support for low-resource languages. The learner applies the AutoML frameworks to a corpus in Tatar, Bashkir, or another language with limited digital resources, in order to ascertain which of the frameworks shows the best results under conditions of the absence of pre-trained language models.

\textbf{Fourth Additional Task.} This consists of a direct comparison of AutoML models with the classical classifiers from Practical Work~\ref{work:4}. The learner ascertains the extent to which AutoML surpasses carefully tuned XGBoost and logistic regression under the same computational budget and the same input data.

\textbf{Fifth Additional Task.} This presupposes an assessment of resource consumption: the learner measures the peak RAM usage and, where present, GPU usage for each framework and constructs comparative diagrams.

\textbf{Sixth Additional Task.} This consists of an analysis of the composition of the final ensembles. The learner investigates exactly which models (transformers, gradient boosting, linear models) entered the final ensemble of AutoGluon and FLAML, and draws conclusions about the frameworks' preferences regarding types of algorithms.

\textbf{Seventh Additional Task.} This is devoted to robustness to class imbalance. The learner ascertains how AutoML systems cope with strong class imbalance without manual correction and compares their behaviour with classical models equipped with class weights and SMOTE.

\textbf{Eighth Additional Task.} This consists of an assessment of the transferability of models. The learner loads a model trained on a news corpus and applies it to texts from another domain (e.g., social media), measuring the drop in quality and analysing its causes.

\section{Conclusion}
This work is not reducible to the simple launching of a ``black box'' with automatic tuning. It represents a conscious choice of a tool under conditions of multiple constraints: linguistic, computational, temporal, and methodological. It is here that the understanding is formed that AutoML is not a replacement for the expert but an amplifier of their competencies, making it possible to reach a level of baseline performance more quickly and to concentrate on interpretation, ethics, and applied aspects.

In the course of performing the work, the learner learns to pose the key questions that determine the value of automation. What the system does automatically --- and what must be controlled manually (a question about the boundaries of automation). To what extent the quality depends on the quality of the data, and not on the ``magic'' of the framework (a question about responsibility for the data). Whether it is possible to explain why the model chose precisely this class (a question about interpretability and trust). Whether the result will be reproducible tomorrow and by another researcher (a question about scientific integrity and openness). The answers to these questions constitute the difference between an ``automaton'' and a responsible automated system.

It is precisely this kind of critically considered, technically rigorous, and methodologically disciplined approach to AutoML that lies at the foundation of modern practices of MLOps, rapid prototyping, and applied NLP under conditions of resource constraints. Upon completion of the work, the learner receives not simply a set of models but a practical compass for choosing, evaluating, and applying AutoML solutions in any future projects --- from startups to large-scale research initiatives.

\chapter{Text Classification with Deep Learning and Transformers}
\label{work:6}

\section{Aim and Learning Outcomes}
This work introduces deep neural text classifiers in the context of more powerful pretrained transformer architectures. The goal is to compare a genuine implemented neural baseline with classical methods while understanding architecture-specific claims, computational cost, regularisation and the distinction between an MLP demonstration and transformer fine-tuning.

The three tasks below constitute the required experimental core. Larger engineering studies and independent research questions are available as optional extensions.

\section{Theoretical Background}

\noindent\textit{Principal theoretical and methodological sources for this work:} \cite{ru_reference_27,devlin2018bert,ru_reference_29,ru_reference_30,ru_reference_31,vaswani2017attention,kim2014convolutional}.\par

Text classification using deep learning has passed through several evolutionary stages, each of which solved the fundamental limitations of the previous generation. Depending on the problem statement, three main types of text classification are distinguished. Binary classification presupposes a choice between two mutually exclusive labels, for example, a positive or negative tone of a review. Multi-class classification requires the assignment of a document to one of several categories --- ``politics'', ``economy'', ``sport'', ``culture'', and so on. Multi-label classification, particularly relevant for news texts, permits the simultaneous assignment of several thematic labels to a document, which makes the task more complex and requires specific loss functions and metrics. \cite{vaswani2017attention,devlin2018bert,kim2014convolutional}

To solve the enumerated tasks within the framework of deep learning, architectures are applied that differ in their ability to model sequences, context, and semantic dependencies. Three key directions are distinguished.

The first direction --- basic neural network models --- includes the multilayer perceptron (MLP), which operates on fixed vector representations of text, e.g., averaged word embeddings, and serves as a simple baseline for comparison; and convolutional neural networks (CNN), which apply one-dimensional convolutions to reveal local semantic patterns, such as key phrases. CNNs are effective under limited computational resources and for short texts; however, they do not capture long-range dependencies between words.

The second direction --- recurrent architectures and the attention mechanism --- models the sequential nature of text. Recurrent networks RNN, their improved gated variants LSTM and GRU, bidirectional (BiLSTM, BiGRU), and multi-layer (Stacked) modifications extract context in both directions and at several levels of abstraction, which is especially important for languages with free word order. The addition of an attention mechanism (in the encoder-decoder with attention format) allows the model to dynamically focus on relevant parts of the text when computing the final representation, substantially increasing both quality and interpretability --- attention weights show which words the model relied on.

The third direction --- transformers --- has revolutionised NLP. The rejection of recurrence in favour of fully connected self-attention ensures parallel processing and the modelling of global dependencies, which makes transformers the dominant architecture. For the Russian language, the most effective are pre-trained models: RuBERT (DeepPavlov), ruRoBERTa (Sber AI), DistilRuBERT, rubert-tiny, mDeBERTa-v3-base, and others. All of them were trained on large Russian-language corpora and support the fine-tuning procedure, whereby the pre-trained weights are adapted to a specific classification task by adding a classification head configured for the corresponding task type (binary, multi-class, or multi-label).

Each successive architectural generation brought not only higher accuracy but also a different type of relationship between the researcher and the model. An MLP working with averaged embeddings is simple to train, and its separating surface can be inspected through feature weights, but it completely ignores word order. An LSTM with an attention mechanism preserves order and provides the opportunity to look, through attention weights, at which parts of the input text influenced the prediction. A transformer gives the richest representations and the most interpretable attention patterns; however, it also requires the greatest computational resources and the utmost caution to avoid overfitting. The choice of architecture is thus a decision about which trade-off the researcher is prepared to accept: between simplicity and expressiveness, transparency and performance, computational costs, and the marginal gain in quality.

The quality of deep models depends directly on several key factors. The correctness of labelling and class balance are especially critical for neural networks, which are prone to overfitting on majority examples. The preprocessing strategy for transformers differs from classical NLP: lemmatization and stemming are not recommended, since the model is optimised for the original word forms and subword tokens. The choice of loss function, regularisation strategy, and evaluation metrics also has decisive importance: simple accuracy is often misleading under class imbalance, and it is necessary to apply $F_1$, PR-AUC, and other more informative indicators.

Unlike classical methods, deep models require a visual analysis of internal mechanisms: neuron activations, attention maps, and embedding projections. Contextual error analysis is also critically important, since the errors of deep models are often systematic but semantically subtle in nature --- for example, confusion between ``sanctions'' and ``support'' in political texts. It is for this reason that interpretability is not an optional feature but a mandatory part of modern NLP research.

\subsection{Experimental Interpretation}
A PyTorch multilayer perceptron can serve as a useful neural baseline but is not a transformer. Transformer-specific claims should be supported by an experiment that actually loads or trains a transformer architecture. Fixing random seeds can reduce variation but does not guarantee bit-for-bit reproducibility across hardware, drivers or nondeterministic operations.

\section{Execution Procedure}

\subsection{Core: Three Required Tasks}
\begin{enumerate}[leftmargin=*,label=\textbf{\arabic*.}]

\item Implement a compact neural classifier and a classical reference on the same training/test corpus. Record tokenisation, input dimensions and parameter count.

Use one fixed corpus and label mapping, and describe the neural input representation (for example averaged word vectors or a bounded vocabulary of token IDs). Report input length handling, architecture layers, parameter count and the loss function for the compact neural model. Train a classical reference on the same training documents and explain any representation differences. Reserve separate validation and test data before model development so that stopping decisions do not inspect the final test labels.

\item Train the neural baseline with explicit training/validation separation; compare macro-F1, training cost and inference latency.

Define batch size, optimiser, learning rate, stopping criterion, number of passes and random seed in the training configuration. Track training and validation losses to distinguish convergence from overfitting. On a fixed test set, report class-aware quality and transparent timing measures; indicate whether preprocessing and data transfer were included in latency. If hardware is limited, use a smaller stated subset rather than implying that a pretrained transformer was trained when no such run occurred.

\item Analyse errors and explain which properties would require an actual pretrained transformer experiment. Do not describe a small MLP as a transformer implementation.

Use predictions and a confusion matrix to select examples of systematic failure, then relate those failures to limited context, vocabulary coverage or class imbalance where justified. Explain what changes when a pretrained transformer is introduced: subword segmentation, contextual states, attention masks, truncation, pretrained weights and the options of freezing or updating those weights. Treat a transformer comparison as a measured extension only if its model identifier, revision and executed settings are available.

\end{enumerate}

\subsection{Reproducible Example and Its Limits}

The companion project contains executable code for this work. Run the real-data path with
\begin{center}\texttt{python practikum\_examples.py 6}\end{center}
20 Newsgroups; this executable demonstration uses a PyTorch MLP, not a transformer. For a technical-only check, add \texttt{-{}-smoke}; its synthetic inputs must not be used as scientific evaluation results. The program does not contain fabricated model scores. Record actual outputs, sample sizes, versions, timings and failures after running it.

\subsection{Extended Engineering Tasks (Optional)}
The following tasks provide a fuller engineering and research programme. They are optional rather than prerequisites for passing the core. The numerical scales and model choices below describe ambitious projects; they are not universal requirements. Documentation and reporting follow the shared protocol above.

\paragraph{Extended Task 1.} On the basis of the text corpus formed during Practical Work~\ref{work:1} and the labelling prepared in Practical Work~\ref{work:4} and Work~\ref{work:5}, a final dataset is formed for three types of classification tasks: binary (sentiment), multi-class (topic), and multi-label (list of topics). For binary classification, the \texttt{sentiment} field is used, containing the labels ``positive'' and ``negative'', obtained in the previous works automatically using sentiment lexicons (e.g., RuSentiLex for the Russian language), semi-automatically by analysing headings, or expertly. For multi-class classification, the \texttt{category} field is used, containing the thematic category of the document (``politics'', ``economy'', ``sport'', ``culture'', and so on). For multi-label classification, the \texttt{categories} field is used, containing the list of thematic labels simultaneously assigned to the document.

The minimum total volume of labelled data is ten thousand documents. If the labelling of all three types of tasks in full is objectively difficult --- for example, due to the high cost of expert annotation, the limited volume of the source corpus, or the unavailability of multi-label annotation, --- the learner has the right, in agreement with the instructor, to reduce the number of labelled documents or to confine themselves to one or two types of tasks, explicitly stipulating this circumstance in the datasheet and discussing its influence on the statistical significance and generalisation ability of the obtained models in the analytical report.

Each document is saved as a JSON object with the following fields: \texttt{id} --- unique document identifier; \texttt{text} --- full text after cleaning; \texttt{title} --- heading (optional); \texttt{sentiment} --- sentiment label (for the binary task, values ``positive'' / ``negative''); \texttt{category} --- thematic category (for the multi-class task); \texttt{categories} --- list of thematic labels (for the multi-label task). The data are saved in JSONL format with UTF-8 encoding.

The split into training, validation, and test sets is performed in a $70/15/15$ ratio. To ensure the representativeness of all classes in each of the subsets, a stratified split is applied with respect to the target variable: for the binary task --- with respect to the \texttt{sentiment} field; for the multi-class task --- with respect to the \texttt{category} field; for the multi-label task --- using iterative stratification (e.g., the \texttt{iterative\_train\_test\_split} algorithm from the \texttt{skmultilearn} library), which takes into account the joint distribution of labels and aims to preserve the proportions of all category combinations. The random seed value used in the split is fixed and documented to support repeatability, subject to library, driver and hardware constraints.

For architectures based on transformers (BERT, RuBERT, ruRoBERTa, DistilRuBERT, mDeBERTa, and others), the text is preserved in its original form, without the application of lemmatization or stemming, since the pre-trained tokenizers of these models are optimised for the original word forms and subword units. For the remaining architectures --- multilayer perceptron, convolutional and recurrent networks --- the application of the tokenisation and normalisation methods mastered in Practical Work~\ref{work:1} (na\"ive, lemmatization, BPE, WordPiece, Unigram) is permitted, with the explicit recording of the chosen strategy in the metadata of the experiment.

In view of the significant computational demands imposed by the training of deep neural network architectures, the learner is recommended to use the Google Colab environment with a GPU accelerator or other cloud platforms providing access to graphics processors. The absence of a local GPU is not an obstacle to the performance of the work; however, it must be explicitly reflected in the methodological section of the report, with an indication of the computing environment used and its characteristics.

References to the specific artefacts of the previous works (the path to the corpus file, the datasheet, the chosen tokenisation and normalisation schemes from Work~\ref{work:1}, the labelled datasets from Works~\ref{work:4} and Work~\ref{work:5}) are recorded in the methodological section of the report to ensure full traceability of the experiment. If the learner performs the present work independently of the previous ones, they must first create or select a corpus and perform its labelling satisfying the requirements of Works~\ref{work:1}, Work~\ref{work:4}, and Work~\ref{work:5}, and explicitly indicate their origin and characteristics in the datasheet.

\paragraph{Extended Task 2.} Several independent input representations are created, adapted for specific architectures. For the MLP, averaged word embeddings (from Practical Work~\ref{work:2}) are used or a trainable embedding layer on top of token indices obtained using subword tokenizers (BPE, WordPiece, Unigram). For CNN and recurrent networks (RNN, LSTM, GRU, BiLSTM, BiGRU, Stacked RNN), the text is transformed into sequences of integer identifiers with the fixation of a maximum length (256 tokens recommended) and subsequent padding/truncation. Where necessary, pre-trained vectors (FastText, Word2Vec) are used for initialising the embedding layer. For all transformers, the official \texttt{AutoTokenizer.from\_pretrained} is applied, generating the tensors \texttt{input\_ids}, \texttt{attention\_mask}, and, if required by the architecture, \texttt{token\_type\_ids}. The metadata of each configuration are saved in JSON format.

\paragraph{Extended Task 3.} A configurable software module \texttt{deep\_classifiers.py} is developed, implementing a unified interface for training and inference of the enumerated architectures. The supported architectures include: basic --- MLP and CNN (1D convolutions with filter sizes 3--5); recurrent --- RNN, LSTM, GRU in unidirectional, bidirectional (Bi), and stacked (Stacked) variants; an architecture with attention --- encoder-decoder with additive or multiplicative attention; and transformers: \texttt{bert-base-multilingual-cased}, \texttt{DeepPavlov/rubert-base-cased}, \texttt{sberbank-ai/ruRoberta-large}, \texttt{cointegrated/rubert-tiny2}, \texttt{microsoft/mdeberta-v3-base}. The module must support execution on GPU (CUDA) via PyTorch, contain documentation of all parameters, and ensure reproducibility by fixing the random number generators.

\paragraph{Extended Task 4.} Rigorous training and evaluation of all models are performed with visualisation of the process. The validation set ($15\%$) is used for early stopping and monitoring the curves of the loss function and the $F_1$ metric (macro-averaged for binary and multi-class, sample-averaged for multi-label) on the training and validation sets by epochs. Training plots, hyperparameter heat maps (if a search was applied), as well as learning curves are constructed for diagnosing overfitting. All configurations, metric values, and weights of the trained models are saved in JSON and PyTorch (\texttt{.pt}) formats.

\paragraph{Extended Task 5.} An empirical assessment of the effectiveness of the architectures and various preprocessing pipelines is performed. The following metrics are used: Accuracy, Precision, Recall, $F_1$ (macro and micro), ROC-AUC (one-vs-rest for multi-class), PR-AUC, Hamming Loss (for multi-label). The results are summarised in a single table \texttt{deep\_classification\_metrics.csv}, the rows of which correspond to the ``architecture $\times$ preprocessing'' combinations, and the columns to the metrics. Additionally, ROC and PR curves are constructed, box plots of the metrics across folds (if cross-validation was applied), as well as two-dimensional projections of the hidden representations (UMAP/$t$-SNE), making it possible to visually assess the degree of clusterability of the classes.

\paragraph{Extended Task 6.} The robustness of the models under conditions of an uneven distribution of labels is ensured. Weighted cross-entropy is applied (weights are calculated inversely proportional to the class frequencies in the training set), and for the most imbalanced scenarios --- focal loss, which reduces the weight of easily classified examples. Quality assessment is performed with an emphasis on metrics not subject to distortion under imbalance: PR-AUC and $F_1$. The report compares the results with and without the application of the correction.

\paragraph{Extended Task 7.} The causes of incorrect predictions and the quality of the predicted probabilities are investigated. For the probabilistic outputs, reliability diagrams are constructed, and the Expected Calibration Error (ECE) is computed. The confusion matrix is analysed, and systematically confused pairs of classes are identified. Examples with high model confidence but an erroneous prediction are selected for subsequent qualitative analysis. Interpretation is carried out by several complementary methods: visualisation of self-attention maps using the \texttt{bertviz} library for transformers; construction of saliency maps and integrated gradients via \texttt{Captum} for all architectures; and analysis of the activations of CNN filters or RNN hidden states for understanding the captured patterns. All maps and plots are included in the report with extensive commentary.

\paragraph{Extended Task 8.} A formal verification of the correctness of the data splitting and the absence of information leakage is performed. The learner documents the splitting procedure and records the script with an indication of the random seed. It is programmatically verified that the sets of document identifiers in the training, validation, and test sets do not intersect. It is separately controlled that, when using synthetic methods for imbalance correction (if any were applied), all generated examples remained strictly in the training portion. The verification protocol is included in the report.

\paragraph{Extended Task 9.} A unified interface for all deep classifiers is developed in the form of a class with \texttt{.fit()} and \texttt{.predict\_proba()} methods, compatible with PyTorch. The class encapsulates the training logic, supports the serialisation of state (model weights, tokenizer parameters, configuration), and records seeds for reproducibility checks, subject to system-dependent variation. Such an approach facilitates integration into industrial pipelines and MLOps infrastructure.

\paragraph{Extended Task 10.} An interactive web application is created using Streamlit or Gradio, allowing the user to: enter arbitrary text; select the task type and architecture (including several transformers); obtain predictions and probabilities, visualise attention maps (for transformers) and saliency maps; view two-dimensional projections of embeddings, training plots, and confusion matrices; compare several models side-by-side; and automatically generate an HTML report with the selected results. The recommended technologies are Streamlit in combination with Plotly and \texttt{bertviz} for displaying attention.

\paragraph{Extended Task 11.} A suite of unit tests is developed based on pytest, verifying the correctness and reproducibility of the key components. The tests check: the fixation of seeds and library versions, the correctness of the data splitting, the stability of the metrics upon repeated runs, the absence of information leakage between the sets, as well as the correctness of the dimensionalities of the output tensors and probabilities. All dependencies are recorded in \texttt{requirements.txt} or \texttt{environment.yml} with exact versions. A script for the automatic execution of the tests and the generation of a pass/fail report is supplied in the repository.

\subsection{Further Research Directions (Optional)}
\textbf{First Additional Task.} This is devoted to comparing the ``evolution of attention'': the learner builds and compares three models --- an encoder-decoder without attention, with attention, and a full transformer --- and analyses how quality and interpretability change.

\textbf{Second Additional Task.} This investigates the ``depth versus width'' trade-off: a stacked LSTM (three layers) and a wide MLP (1024 neurons) are compared with an equal number of parameters.

\textbf{Third Additional Task.} This analyses transfer from multilingual models: an assessment is made of the extent to which ruRoBERTa surpasses multilingual BERT and mDeBERTa on a Russian-language corpus.

\textbf{Fourth Additional Task.} This is devoted to attention errors: cases are sought where the model errs while the attention map ``looks'' at the wrong tokens.

\textbf{Fifth Additional Task.} This studies the influence of subword tokenisation: BPE and WordPiece are compared in CNN and BiLSTM architectures.

\textbf{Sixth Additional Task.} This compares regularisation strategies: dropout versus weight decay versus early stopping --- individually and in combinations.

\textbf{Seventh Additional Task.} This assesses the influence of calibration (Platt scaling, isotonic regression) on the reliability of transformer probabilities.

\textbf{Eighth Additional Task.} This performs an in-depth analysis of attention heads in RuBERT: a systematic investigation is carried out of which of them focus on syntactic and which on semantic relations, with visualisation on concrete examples.

\section{Conclusion}
This work is not simply ``loading BERT and launching training''. It is a deep immersion into the architectural evolution of NLP, where each model represents an answer to a specific challenge of language. Here, the understanding is formed that the choice of architecture is not a technical task but a methodological and even philosophical one: what is more important --- interpretability, speed, accuracy, or robustness?

In the course of the work, the learner learns to pose the key questions that determine the reliability of modern NLP systems. What exactly the model is looking at when making a decision (a question about the transparency of attention). Whether its confidence can be trusted --- or it is simply ``confidently wrong'' (a question about calibration and reliability). How the model errs --- randomly or systematically (a question about ethical robustness). Whether reproducibility is guaranteed from the first token to the final attention map (a question about scientific integrity and openness).

The answers to these questions constitute the difference between a ``neural network'' and a responsible intelligent system. It is precisely this kind of systematic, critically considered, and technically rigorous approach to deep learning that lies at the foundation of modern applications: from medical diagnostics to automatic moderation and public opinion analysis. Upon completion of the work, the learner receives not simply a set of models but an architectural compass for designing, evaluating, and deploying any reliable NLP systems --- from academic research to industrial solutions, capable of working not only accurately but also transparently, conscientiously, and reproducibly.

\chapter{Transfer Learning Strategies for NLP}
\label{work:7}

\section{Aim and Learning Outcomes}
The objective is to understand the alternatives for reusing representations learned on other domains or languages. Students evaluate frozen feature extraction against lexical baselines, describe which parameters are trained, and investigate how domain mismatch influences transfer. Full fine-tuning and parameter-efficient methods provide optional extensions with separate evidence requirements.

The three tasks below constitute the required experimental core. Larger engineering studies and independent research questions are available as optional extensions.

\section{Theoretical Background}

\noindent\textit{Principal theoretical and methodological sources for this work:} \cite{hu2022lora,ru_reference_33,howard2018universal,ru_reference_35,arabov2026bashkirbenchmark,arabov2026tajikpeft,houlsby2019parameter}.\par

For low-resource adaptation, comparative studies on Bashkir and Tajik help motivate explicit reporting of model family, parameter budget and evaluation split \cite{arabov2026bashkirbenchmark,arabov2026tajikpeft}.

The evolution of transfer learning in natural language processing represents a transition from static embedding representations, such as Word2Vec or GloVe, which fix a single vector for a word regardless of context, to dynamic contextualised vector models generated by the transformer architecture. Modern pre-trained models --- BERT, RoBERTa, T5, BART --- function as universal ``linguistic priors'', capturing the deep syntactic and semantic regularities of language, which makes it possible to effectively adapt them to specific applied tasks even under conditions of an acute shortage of labelled data. \cite{howard2018universal,houlsby2019parameter,hu2022lora}

Depending on the chosen strategy for adapting a pre-trained model to a target task, three main directions can be distinguished. The first direction --- full transfer, or Full Fine-Tuning, --- presupposes updating all the parameters of the model during training on the target dataset. This method potentially yields the highest quality; however, it is associated with high demands on computational resources and carries a significant risk of overfitting, especially when working with small samples. Its successful application requires careful selection of hyperparameters, including the regularisation strategy and the learning rate schedule.

The second direction --- Parameter-Efficient Fine-Tuning (PEFT) --- seeks to overcome the limitations of full transfer. Instead of updating hundreds of millions of parameters, PEFT methods modify only a small fraction of them. The key representatives of this family are Adapters --- compact bottleneck layers inserted between the main blocks of the transformer, --- and LoRA (Low-Rank Adaptation) --- a method that approximates updates to the attention matrices through the product of low-rank matrices. Beyond these two methods, the PEFT landscape includes prompt tuning, in which trainable virtual tokens are added to the input rather than modifying the model itself; prefix tuning, which trains a continuous task-specific prefix for each layer of the transformer; and IA\textsuperscript{3}, which scales key, value, and feed-forward activations with the aid of trainable vectors. These methods differ in the number of introduced parameters, the granularity of intervention, and the degree of compatibility with the source model. What unites them is a common principle: the pre-trained backbone is frozen, and adaptation is limited to a small, modular, and easily distributable component. These techniques make it possible to train less than one per cent of the total number of model parameters, while preserving over ninety-five per cent of the quality of full fine-tuning. Such an approach proves ideal for resource-constrained scenarios, for deployment on edge devices, and in multi-task systems where many specialised adapters must be stored for a single base model.

The third direction --- Feature Extraction --- presupposes the use of the pre-trained encoder as a frozen, unchangeable module for obtaining vector representations of the text. In this case, only the classification head placed on top of the encoder is trained. This approach is the fastest and most resistant to overfitting, which makes it valuable for ultra-low-resource scenarios; however, its flexibility and upper quality bound are usually lower than those of methods that permit the adaptation of internal representations. Feature Extraction is often used as a reliable baseline for comparison with more complex strategies.

The quality of knowledge transfer is determined by a series of interconnected factors. Of fundamental importance is the semantic closeness of the domain on which the model was pre-trained to the target subject area. No less important is architectural compatibility: encoder-only models (BERT) naturally suit understanding and classification tasks, whereas encoder-decoder models (T5, BART) are the standard for generative tasks such as summarisation. A key role is played by the search for a balance between generalisation and specialisation, which in practice is expressed in the tuning of the regularisation strength and the learning rate decay schedule. Finally, for practical applications, especially in tasks with imbalanced classes or in safety-critical systems, probability calibration is important --- the correspondence of the model's predicted confidence to the actual frequency of correct answers. To understand what information the model retains and what it loses in the process of adaptation, an analysis of internal representations is applied. The tools for this are probing methods (training linear classifiers on hidden states), visualisation of attention maps, and projections of embedding spaces, which is directly connected with the interpretability of the model and the justification of trust in its decisions.

\subsection{Experimental Interpretation}
Frozen embedding extraction, full fine-tuning and parameter-efficient tuning involve different optimisation procedures, parameter counts and computational risks. Frozen embeddings do not establish the behaviour of a fine-tuned model; the study should label which transfer regime is genuinely implemented.

\section{Execution Procedure}

\subsection{Core: Three Required Tasks}
\begin{enumerate}[leftmargin=*,label=\textbf{\arabic*.}]

\item Extract representations from a specified pretrained encoder without updating its weights and build a linear classifier on training documents.

Choose a specified pretrained text encoder that is documented for the language and licence of the corpus. Freeze all encoder weights, put the encoder in evaluation mode, and extract one fixed-length vector per document using a declared pooling strategy such as the pooled output or attention-mask-aware mean pooling. Record the model identifier and revision, tokenizer, truncation length, device, numerical precision and feature dimension. Train only the downstream linear classifier using the training partition, making clear that feature extraction alone does not update the pretrained encoder.

\item Compare frozen representations with a TF--IDF baseline using the same test set and state which components are fitted on which split.

Create a TF--IDF plus linear-classifier baseline using exactly the same document identifiers, training/validation/test boundaries and label mapping. Fit the TF--IDF vocabulary and inverse document frequencies only on training documents, and tune any classifier setting using training or validation data rather than the final test. Evaluate both systems once on a shared held-out test set; report macro-F1, class supports, the number of examples and inference cost if measured. Discuss whether any observed difference could depend on corpus size, text length or language coverage rather than attributing it automatically to transfer learning.

\item Investigate domain/language transfer on a suitable held-out subset. Distinguish frozen-feature transfer from full fine-tuning or parameter-efficient adaptation.

Define an explicit shift, for example a held-out subject area, source collection, time period or language, and justify why it is a legitimate transfer setting. Keep the source-domain training data separate from target-domain test examples and report label compatibility and coverage differences. Compare in-domain and shifted-domain outcomes with identical frozen encoders and state clearly whether target-domain labelled examples were used for tuning. Explain that full fine-tuning updates pretrained parameters and parameter-efficient adaptation updates selected additional parameters: neither has occurred in the frozen-feature experiment unless implemented separately and documented as an extension.

\end{enumerate}

\subsection{Reproducible Example and Its Limits}

The companion project contains executable code for this work. Run the real-data path with
\begin{center}\texttt{python practikum\_examples.py 7}\end{center}
20 Newsgroups plus a published sentence encoder; frozen embedding transfer only. For a technical-only check, add \texttt{-{}-smoke}; its synthetic inputs must not be used as scientific evaluation results. The program does not contain fabricated model scores. Record actual outputs, sample sizes, versions, timings and failures after running it.

\subsection{Extended Engineering Tasks (Optional)}
The following tasks provide a fuller engineering and research programme. They are optional rather than prerequisites for passing the core. The numerical scales and model choices below describe ambitious projects; they are not universal requirements. Documentation and reporting follow the shared protocol above.

\paragraph{Extended Task 1.} On the basis of the text corpus formed during Practical Work~\ref{work:1} and the labelling prepared in Practical Work~\ref{work:4} and Work~\ref{work:5}, three annotated subsets are created for three heterogeneous downstream tasks covering various types of NLP problems. The first subset --- Sentiment-1K --- contains at least one thousand documents labelled with binary sentiment (labels ``positive'' and ``negative'') and is intended for the text classification task. The second subset --- RuNER-5K --- provides annotation of named entities of the categories PER, ORG, LOC, DATE, and MISC with a total volume of at least five thousand sentences in CoNLL format (token, tag); this subset may be obtained using the RuNNE dataset, automatic annotation through a pre-trained NER model with subsequent manual verification, or expert annotation. The third subset --- NewsSumm-3K --- contains at least three thousand ``article text -- headline'' pairs for the task of abstractive summarisation, where the headline acts as the reference summary.

For each of the three subsets, the use of ready-made open datasets relevant to the chosen subject area and task type is permitted (e.g., datasets from the Hugging Face Datasets collections, including Russian-language analogues of SST-2 for sentiment, RuNNE or factRuEval for NER, and Gazeta or Lenta.ru for summarisation); the use of a ready-made dataset does not lower the grade but, on the contrary, is methodologically encouraged, since it allows the learner to concentrate on the comparative analysis of transfer strategies rather than on routine data collection, subject to the mandatory indication of the origin and licence of each used resource in the datasheet.

If the formation of any of the subsets in the indicated volume is objectively difficult --- for example, due to limited access to expert NER annotation for a low-resource language or the absence of ready-made datasets of acceptable quality, --- the learner has the right, in agreement with the instructor, to reduce the volume of the corresponding subset or to use data in a related or contrastive language (e.g., English, Tatar, or Kazakh) with a subsequent analysis of cross-lingual transfer, explicitly stipulating this circumstance in the datasheet and discussing its influence on the statistical significance and generalisation ability of the models in the analytical report.

Each document is saved as a JSON object with the following fields: \texttt{id} --- unique identifier; \texttt{text} --- full text after cleaning; \texttt{sentiment} --- sentiment label (for Sentiment-1K, values ``positive'' / ``negative''); \texttt{ner\_tags} --- list of tokens and their corresponding NER tags (for RuNER-5K, in CoNLL format); \texttt{summary} --- reference headline (for NewsSumm-3K). All subsets are saved in JSONL format with UTF-8 encoding. The mutual disjointness of documents between the three subsets is strictly guaranteed: no document may simultaneously belong to Sentiment-1K, RuNER-5K, and NewsSumm-3K. The split into training, validation, and test sets is performed in a $70/15/15$ ratio with stratification by classes (for Sentiment-1K) and by the distribution of entity tags (for RuNER-5K). The random seed value is fixed and documented to support repeatability, subject to library, driver and hardware constraints. References to the specific artefacts of the previous works are recorded in the methodological section of the report.

\paragraph{Extended Task 2.} Input representations are created, and a pool of pre-trained donor models is formed, subject to systematic comparison within the framework of all subsequent experiments. The pool includes architectures that differ along two key dimensions: architecture type (encoder-only for understanding tasks and encoder-decoder for generative tasks) and pre-training domain (general, specialised, and multilingual models). As the general encoder-only model, \texttt{DeepPavlov/rubert-base-cased}, pre-trained on Russian-language texts of broad themes, is used; its lightweight variant is \texttt{cointegrated/rubert-tiny2}, which makes it possible to assess the ``quality--speed'' trade-off. As a large encoder-only model with potentially higher quality, \texttt{sberbank-ai/ruRoberta-large} is included. To assess the influence of the pre-training domain, \texttt{IlyaGusev/ruRoBERTa-finance} is added to the pool --- a model specialised on financial texts, which makes it possible to investigate the extent to which the semantic closeness of the donor domain to the target task affects the effectiveness of transfer. The multilingual segment is represented by \texttt{microsoft/mdeberta-v3-base}, which supports the Russian language alongside dozens of others, opening the possibility of analysing cross-lingual transfer. For generative tasks (summarisation), \texttt{google/mt5-base} is used --- a multilingual encoder-decoder architecture supporting the Russian language. For each model, its key characteristics are recorded: architecture type, hidden state dimensionality, maximum sequence length, volume of the pre-training corpus (if known), supported languages, licence, and memory requirements for inference and training; these metadata are saved in a structured form in the file \texttt{donor\_models.json}.

Tokenisation is performed via \texttt{AutoTokenizer.from\_pretrained} for encoder-only architectures; for mT5, a separate specialised tokenizer is used. For each task and each model, tokenised inputs are prepared with padding and truncation to the maximum length specified by the architecture: 512 tokens for classification, 256 tokens for NER (preserving alignment at the subword level), 512 tokens for the input text and 128 tokens for the target headline in the summarisation task. For each ``task $\times$ model'' combination, distributions of sequence lengths after tokenisation are computed and visualised, which makes it possible to substantiate the chosen truncation parameters and to reveal the proportion of documents subjected to truncation. The tokenisation metadata, including the tokenizer configuration, length distributions, and the proportion of truncated examples, are saved in JSON format to ensure full reproducibility.

The comparative analysis of the donor models is singled out as an independent research line, the results of which are recorded in the report: for each task, it is determined which of the models provides the best quality in the Feature Extraction mode and in the Full Fine-Tuning mode, how large the gap is between the general and the specialised model when transferring to a close and a distant domain, and whether the multilingual model yields quality comparable to the monolingual one on Russian-language data.

\paragraph{Extended Task 3.} A software module \texttt{transfer\_adaptors.py} is developed, providing a unified interface \texttt{TransferAdaptor} that encapsulates five transfer learning strategies. The first strategy --- Feature Extraction (frozen encoder): only the classification or generation head is trained; all encoder parameters are fixed. The second --- Full Fine-Tuning: all model parameters are updated. The third --- Layer-wise Learning Rates: different layers are trained at different rates, with the upper layers receiving a higher rate and the lower ones a lower rate, using an LLRD (Layer-wise Learning Rate Decay) schedule. The fourth --- Adapter-based Tuning: compact bottleneck adapters are inserted between the layers of the transformer; only the adapter parameters are trained; the backbone is frozen. The fifth --- LoRA (Low-Rank Adaptation): low-rank adapters ($r=8$, $\alpha=16$) are embedded in the attention layers (\texttt{q\_proj}, \texttt{k\_proj}, \texttt{v\_proj}, \texttt{o\_proj}). The module supports three types of downstream tasks: text classification, sequence labelling (NER), and sequence-to-sequence generation (summarisation). GPU acceleration is provided via PyTorch in combination with the Accelerate library. The code is fully type-annotated and furnished with exhaustive documentation.

\paragraph{Extended Task 4.} Rigorous training with visualisation monitoring is performed for all combinations of ``strategy $\times$ task $\times$ model''. A separate validation set ($15\%$) is used for early stopping and monitoring. The target metrics are: macro $F_1$ for classification, entity-level $F_1$ for NER, ROUGE-L for summarisation. For each combination, learning curves are constructed (dependence of the loss function and the target metric on the epoch number for the training and validation sets), plots of the gradient norm, hyperparameter heat maps (if a search was applied), as well as learning curves for diagnosing overfitting. All configurations, model weights, and training logs are saved in JSON and Hugging Face formats.

\paragraph{Extended Task 5.} An empirical assessment of the effectiveness of the transfer strategies is performed according to the following dimensions: model quality ($F_1$ and ROUGE); computational efficiency (training time, proportion of trainable parameters, and peak GPU memory consumption); stability of results (variance of quality metrics under different random seed initialisations). The results are presented in the form of a final summary table \texttt{transfer\_comparison.csv}, the rows of which correspond to the ``strategy $\times$ task $\times$ model'' combinations, and the columns to the quality and efficiency metrics. Additionally, box plots of the metric distributions and radar charts are constructed, displaying the three-way trade-off ``quality -- resources -- speed''.

\paragraph{Extended Task 6.} The dependence of downstream task quality on the size of the training set is investigated. From the training portion of each subset, sets of sizes 100, 300, 1000, 3000, and 10000 instances are formed (if a sufficient volume of data is available). The RuBERT model is trained on each set using all available transfer strategies. Learning curves with confidence intervals are constructed, displaying the mean value and standard deviation of the metric over several runs. For each combination, the saturation point is determined --- the data volume after which an additional increase in the set yields less than one percentage point of metric gain. The results are visualised in the form of a single plot, where the set size is plotted on the abscissa and the target metric on the ordinate, with separate curves for each strategy.

\paragraph{Extended Task 7.} The robustness of predictions to linguistic perturbations and the quality of probability calibration are assessed. To assess robustness, a paraphrase attack is performed: using a T5 model fine-tuned on paraphrasing (from Practical Work~\ref{work:8}), 3--5 reformulations of each input are generated in the test set of the classification and NER tasks; for the summarisation task, back-translation (Russian $\rightarrow$ English $\rightarrow$ Russian) is performed via the \texttt{Helsinki-NLP/opus-mt-ru-en} model. For each distorted version, the same metrics as on the clean data are computed, and the relative drop in quality is measured. To assess calibration, reliability diagrams are constructed, and the Expected Calibration Error (ECE) is computed; additionally, the Brier Score and Negative Log-Likelihood (NLL) are reported. Calibration correction methods are applied --- Platt scaling and isotonic regression, --- and the reduction in ECE after correction is measured. The results are summarised in separate robustness and calibration tables.

\paragraph{Extended Task 8.} The internal mechanisms of transfer learning are investigated. For each strategy, linear probing is performed: on the hidden representations of each layer of the model before and after adaptation, a linear classifier is trained to solve the target task, and the probing accuracy is displayed as a plot across layers. Dimensionality reduction methods --- UMAP and $t$-SNE --- are applied to the [CLS] embeddings and to the token representations, and projections are visualised with colour coding by classes or entity types. Using the \texttt{bertviz} library, attention maps are visualised before and after adaptation. An analysis of catastrophic forgetting is performed: it is measured to what extent the model preserves quality on the original pre-training task (Masked Language Modelling) after adaptation by each of the strategies. All visualisations are generated directly in the course of code execution and are included in the analytical report with extensive commentary.

\paragraph{Extended Task 9.} A formal verification of the correctness of the data splitting and the absence of information leakage is performed, and a unified class \texttt{TransferAdaptor} is developed. The learner documents the splitting procedure with the fixation of the random seed, and programmatically verifies the mutual disjointness of the three subsets and the absence of intersections between the training, validation, and test portions of each subset. It is controlled that the synthetic paraphrases and back-translations (from Task~7) were not used in training. The verification protocol is included in the analytical report. In parallel, the class \texttt{TransferAdaptor} is developed with the public methods \texttt{.fit()} (training the chosen strategy), \texttt{.predict\_proba()} (obtaining class or token probabilities), and \texttt{.explain()} (return of attention maps and probing results). The class is compatible with PyTorch and Hugging Face, supports the serialisation of state via \texttt{torch.save} and \texttt{joblib}, and records seeds but remains subject to platform-dependent nondeterminism.

\paragraph{Extended Task 10.} An interactive web application is created using Gradio, allowing the user to: choose a task (classification, NER, summarisation), a model, and a transfer strategy; enter arbitrary text and obtain a prediction with the display of probabilities and attention maps; for NER --- visualise entity annotation inline; for summarisation --- compare the generated summary with the source text; compare two strategies side-by-side in parallel; and automatically generate an HTML report with the results. The application is designed in such a way that it can be used by a researcher without programming skills.

\paragraph{Extended Task 11.} A suite of unit tests is developed based on the pytest framework, verifying the correctness and reproducibility of the key project components. The tests must check: the determinism of the results with a fixed random seed value and pinned library versions; the correctness of the train/validation/test split; the stability of the metrics upon repeated runs; the complete absence of information leakage between the training and test sets; the correctness of loading all five strategies through the unified \texttt{TransferAdaptor} interface. All dependencies are recorded in the \texttt{requirements.txt} file with an indication of the exact versions. A script is developed for the automatic execution of the full test suite and the generation of a pass/fail report. The presence of a successfully passing test suite is regarded as an integral component of the software artefact.

\subsection{Further Research Directions (Optional)}
\textbf{First Additional Task.} Comparison of PEFT methods: Adapters versus LoRA --- quality, speed, memory consumption, all other things being equal.

\textbf{Second Additional Task.} Influence of the pre-training domain: to what extent does a model pre-trained on financial texts (ruRoBERTa-finance) fall short of the general RuBERT on general tasks?

\textbf{Third Additional Task.} Zero-shot via prompting: comparison of zero-shot (via templates) and few-shot (100 examples) on the sentiment analysis task.

\textbf{Fourth Additional Task.} Economics of transfer: measurement of the CO\textsubscript{2} footprint and training cost via CodeCarbon for each strategy.

\textbf{Fifth Additional Task.} Multi-task adaptation: training a single model on sentiment and NER simultaneously --- is positive transfer observed between tasks?

\textbf{Sixth Additional Task.} Calibration after PEFT: does calibration deteriorate when using LoRA and adapters compared to full fine-tuning?

\textbf{Seventh Additional Task.} Cross-lingual transfer: is mDeBERTa applicable to Russian-language NER and does it surpass RuBERT?

\textbf{Eighth Additional Task.} Analysis of layers via probing: which layers carry syntactic and which carry semantic information after adaptation?

\section{Conclusion}
This work is not simply ``fine-tuning BERT on your own data''. It is a deep immersion into the architectural, algorithmic, and methodological foundations of modern transfer learning, where each strategy represents an answer to a specific challenge: data scarcity, resource constraints, the requirement for interpretability or robustness.

In the course of performing the work, the learner learns to ask the key questions that determine the reliability of modern NLP systems. What exactly changes in the model during adaptation --- and what is lost (a question about internal representations). Whether its confidence can be trusted --- or it is simply ``confidently wrong'' (a question about calibration). How the model reacts to a synonymous replacement --- robustly or chaotically (a question about robustness). Whether reproducibility is guaranteed --- from the first token to the final attention map (a question about scientific integrity).

The answers to these questions constitute the difference between a neural network and a responsible intelligent system. Upon completion of the work, the learner receives not simply a set of models but an architectural compass for designing, evaluating, and deploying any reliable NLP systems --- from academic research to industrial solutions, capable of working not only accurately but also transparently, honestly, and reproducibly.

\chapter{Multi-Task NLP: Information Extraction and Generation}
\label{work:8}

\section{Aim and Learning Outcomes}
This work develops an evidence-based approach to tasks including named-entity recognition, text generation and semantic analysis. Learners select appropriate model outputs and evaluation methods rather than applying a single metric to every task, document the limitations of available labels, and inspect output quality and error types with reproducible model settings.

The three tasks below constitute the required experimental core. Larger engineering studies and independent research questions are available as optional extensions.

\section{Theoretical Background}

\noindent\textit{Principal theoretical and methodological sources for this work:} \cite{raffel2020exploring,arabov2026tajperslexon,ru_reference_40,ru_reference_41,arabov2026arabicrussianmt}.\par

Cross-script lexical resources and Arabic--Russian translation benchmarks provide further cases in which generation and multilingual evaluation require language-specific controls \cite{arabov2026tajperslexon,arabov2026arabicrussianmt}.

Modern NLP systems are rarely limited to a single task. In real scenarios --- from news analysis to the processing of customer enquiries --- a combination of methods for extracting, structuring, and generating information is required. The effectiveness of such systems depends not only on the quality of the individual components but also on their compatibility, robustness to errors, and interpretability. \cite{peters2018deep,raffel2020exploring}

Within the framework of the present work, five fundamental directions are distinguished, each of which possesses its own methodological and architectural specificity.

The first direction --- information extraction --- includes Named Entity Recognition (NER) and keyphrase extraction. NER is a sequence labelling task: each token is assigned a tag (PER, ORG, LOC, DATE, MONEY, and others). Keyphrases can be extracted statistically (TF-IDF, RAKE), through ranking (YAKE), or using transformers (KeyBERT). Classical approaches rely on Conditional Random Fields (CRF) on top of handcrafted features; deep models --- on BiLSTM-CRF, which was the gold standard before the era of transformers; modern solutions --- on classification heads on top of RuBERT or ruRoBERTa, fine-tuned on RuNNE or custom annotations.

The second direction --- text generation --- covers extractive summarisation (selection of the most relevant sentences, often using TextRank or trained ranking models), abstractive summarisation (generation of new formulations using encoder-decoder architectures: BART, T5, mT5), and paraphrasing (a special case of generation, often solved by fine-tuning T5 with the prefix ``paraphrase:''). For the Russian language, the most relevant models are ruT5, flan-t5-base-ru, mT5-large, which support zero-shot and few-shot configurations.

The third direction --- semantic analysis --- presupposes the representation of texts as dense vectors (sentence embeddings) using models such as SBERT, ruSBERT, LaBSE, and comparison via cosine distance. Applications include duplicate detection, semantic search, clustering, and recommendation systems. The calibration of similarity is critically important: not all pairs with high cosine similarity are semantically equivalent, and a careful choice of threshold is required.

The fourth direction --- grammatical correctness and robustness to distortions --- is of an end-to-end, integrative nature. If the first three directions presuppose working with clean, normalised texts, this direction poses the question of how NLP components behave under conditions approximating real operation: in the presence of typos, spelling errors, character omissions, letter transpositions, and other distortions typical of user content, social media, transcripts, and optical character recognition (OCR) results. It includes two interconnected sub-tasks. The first is robustness assessment: measuring the degradation of extraction, generation, and semantic analysis quality when controlled noise is introduced into the input text. The second is automatic grammatical error correction (GEC): building a model that receives distorted text as input and generates its grammatically and orthographically correct version. For the Russian language, an effective solution is the fine-tuning of ruT5 with the prefix ``correct the errors:'' or the application of mT5 in few-shot mode. The quality of correction is measured by the metrics Accuracy and ERRANT, which distinguish successful corrections, hypercorrection (unjustified change of a correct word), and missed errors.

The fifth direction --- hybrid architectures --- unites the enumerated components into pipelines, where the output of one model (e.g., NER) becomes the input of another (e.g., a summariser enriched with named entities), and the grammatical correction block may precede all other stages, ensuring the normalisation of the input text. It is also possible to use a single transformer with several heads (multi-task learning), which reduces the overall complexity and error propagation but requires strict coordination of data formats, tokenizer compatibility, and management of inference latencies.

The practical work is structured so that these five directions are not studied in isolation. The learner first implements and evaluates each component independently, then investigates their robustness to distortions, and finally unites them into an end-to-end pipeline. Such a progression --- from isolated evaluation to an integrated, noise-robust system --- replicates the trajectory of a real NLP project and develops the ability to diagnose whether an error arises at the stage of extraction, generation, search, correction, or at the junction of components.

The quality of the solution depends not only on the model but also on the correct formulation of the task, the choice of metric, and the analysis of the specifics of errors --- for example, hallucinations in generation (the introduction of facts absent from the source), misses in NER (especially for rare or ambiguous entities), or hypercorrection in error correction (changing a correct word to an incorrect one). It is for this reason that interpretability and error analysis are not an optional feature but a mandatory part of modern NLP research.

\subsection{Experimental Interpretation}
Different NLP outputs require task-appropriate reference annotations: entity extraction, summarisation, paraphrase and generation cannot all be evaluated meaningfully with the same accuracy metric. A qualitative inspection is legitimate when labels are unavailable, but must never be presented as a quantitative gold-standard evaluation.

\section{Execution Procedure}

\subsection{Core: Three Required Tasks}
\begin{enumerate}[leftmargin=*,label=\textbf{\arabic*.}]

\item Select a published model for a documented NLP task (for example NER) and identify its model card, supported language, licence and output schema.

Select a concrete input/output task and a published checkpoint capable of performing it; document the supported language, expected input length, licence, revision and output fields. Where entity recognition is chosen, define the entity labels and decide whether predictions are assessed at token or span level. Obtain a small real evaluation sample with permitted use and, if supervised scoring is intended, independently sourced reference annotations. Do not treat a model's own predictions as verified ground truth.

\item Run the model on real text samples and construct a rule-based or qualitative comparison suitable to the available annotations; do not invent a gold standard.

Execute the selected model with fixed preprocessing, batching and decoding settings, keeping the original text alongside character offsets or structured outputs as applicable. Compare against an appropriate and clearly limited reference, such as a dictionary or regular-expression extraction baseline, or provide a qualitative comparison if no reliable annotation exists. State which outputs the baseline can and cannot generate. Count evaluated examples and report accuracy or span-level measures only when the reference annotation supports that calculation.

\item Inspect at least ten correct, uncertain or erroneous outputs, discuss domain limitations and give the exact model revision and processing parameters.

Select ten varied outputs using an explicit procedure and examine boundary errors, ambiguous cases, missed entities and domain-specific language, where these phenomena actually occur. Show the exact inputs needed to reproduce the examples without disclosing private text. Separate failures caused by model limitations from evaluation ambiguity or an incomplete label set. If text generation is studied as an extension, assess it with task-appropriate evidence rather than carrying over NER measures unchanged.

\end{enumerate}

\subsection{Reproducible Example and Its Limits}

The companion project contains executable code for this work. Run the real-data path with
\begin{center}\texttt{python practikum\_examples.py 8}\end{center}
Real 20 Newsgroups text and a task-specific NER pipeline; inspected spans are not assumed to be labelled ground truth. For a technical-only check, add \texttt{-{}-smoke}; its synthetic inputs must not be used as scientific evaluation results. The program does not contain fabricated model scores. Record actual outputs, sample sizes, versions, timings and failures after running it.

\subsection{Extended Engineering Tasks (Optional)}
The following tasks provide a fuller engineering and research programme. They are optional rather than prerequisites for passing the core. The numerical scales and model choices below describe ambitious projects; they are not universal requirements. Documentation and reporting follow the shared protocol above.

\paragraph{Extended Task 1.} On the basis of the text corpus formed during Practical Work~\ref{work:1} and the labelling prepared in Practical Work~\ref{work:4} and Work~\ref{work:5}, a final dataset is formed for five types of NLP tasks: information extraction (NER, keyphrases), generation (summarisation, paraphrasing), semantic analysis (similarity, duplicates), grammatical correctness (error correction), and hybrid annotation. For NER, annotation is performed using the RuNNE dataset or expertly verified annotations with the tags PER, ORG, LOC, DATE, MONEY, and others. For summarisation, the first two sentences of the document are taken as the extractive baseline. For grammatical correction, a parallel corpus ``distorted text $\rightarrow$ correct text'' is created by automatically introducing controlled distortions using the script \texttt{text\_corruptor.py} (see Task~8). Each document is saved as a JSON object with the fields: \texttt{id} --- unique identifier; \texttt{text} --- full text after cleaning; \texttt{title} --- heading (optional); \texttt{ner\_tags} --- list of tokens and their corresponding NER tags (optional); \texttt{summary} --- extractive summary (optional); \texttt{keyphrases} --- list of keyphrases (optional); \texttt{corrupted\_text} --- distorted version of the text (optional, for the correction task). The data are saved in JSONL format with UTF-8 encoding.

The minimum total volume of labelled data is ten thousand documents. If the labelling of all types of tasks in full is objectively difficult --- for example, due to the high cost of expert NER annotation, the limited volume of the source corpus, or the absence of resources for manually creating a parallel error corpus, --- the learner has the right, in agreement with the instructor, to reduce the number of labelled documents or to confine themselves to three or four types of tasks, explicitly stipulating this circumstance in the datasheet and discussing its influence on the statistical significance and generalisation ability of the obtained models in the analytical report.

The split into training, validation, and test sets is performed in a $70/15/15$ ratio. To ensure the representativeness of all classes in each of the subsets, a stratified split is applied with respect to the target variable specific to each task: for NER --- by the distribution of entity tags; for summarisation --- by the length of documents in sentences; for grammatical correction --- by the proportion of distorted words. The random seed value used in the split is fixed and documented to support repeatability, subject to library, driver and hardware constraints.

For architectures based on transformers (RuBERT, ruT5, mT5, flan-t5, and others), the text is preserved in its original form, without the application of lemmatization or stemming, since the pre-trained tokenizers of these models are optimised for the original word forms and subword units. For the remaining architectures --- BiLSTM-CRF, CNN for NER, statistical methods --- the application of the tokenisation and normalisation methods mastered in Practical Work~\ref{work:1} is permitted, with the explicit recording of the chosen strategy in the metadata of the experiment.

In view of the significant computational demands imposed by the training of deep neural network architectures and generative models, the learner is recommended to use the Google Colab environment with a GPU accelerator or other cloud platforms providing access to graphics processors. The absence of a local GPU is not an obstacle to the performance of the work; however, it must be explicitly reflected in the methodological section of the report, with an indication of the computing environment used and its characteristics.

References to the specific artefacts of the previous works (the path to the corpus file, the datasheet, the chosen tokenisation and normalisation schemes from Work~\ref{work:1}, the labelled datasets from Works~\ref{work:4} and Work~\ref{work:5}, the pre-trained embeddings from Work~\ref{work:2}, the fine-tuned transformers from Work~\ref{work:6}, the transfer learning strategies from Work~\ref{work:7}) are recorded in the methodological section of the report to ensure full traceability of the experiment. If the learner performs the present work independently of the previous ones, they must first create or select a corpus and perform its labelling satisfying the requirements of Works~\ref{work:1}, Work~\ref{work:4}, and Work~\ref{work:5}, and explicitly indicate their origin and characteristics in the datasheet.

\paragraph{Extended Task 2.} Several independent but compatible data representations are created depending on the type of task. For NER and semantic analysis, tokenisation is performed via \texttt{AutoTokenizer.from\_pretrained} of the corresponding models (RuBERT, ruRoBERTa, LaBSE). For statistical methods of keyphrase extraction (RAKE, YAKE), raw text with minimal cleaning --- removal of stop words and punctuation --- is used. For training grammatical correction models, paired representations ``distorted text --- correct text'' are created, preserving alignment at the token or character level. All prepared representations are saved with metadata in JSON format, including information about the preprocessing strategy, the tokenizer used, and a reference to the source document, which ensures full reproducibility of the pipeline.

\paragraph{Extended Task 3.} A configurable software module \texttt{nlp\_pipelines.py} is developed, providing a unified interface for training and inference of all supported components. The set of implemented models includes:
\begin{itemize}
    \item for NER: spaCy (baseline), BiLSTM-CRF (deep architecture, trained from scratch), and RuBERT with a classification head, fine-tuned on RuNNE or custom annotation;
    \item for keyphrase extraction: RAKE, YAKE, and KeyBERT (using RuBERT as the base encoder);
    \item for summarisation: TextRank (extractive), BERT-extractive (sentence-ranking), and ruT5/flan-t5 (abstractive);
    \item for paraphrasing: ruT5 with the instruction ``paraphrase:'' and mT5;
    \item for semantic analysis: ruRoBERTa-STSB and paraphrase-multilingual-MiniLM for obtaining sentence embeddings;
    \item for grammatical correction: ruT5 with the prefix ``correct the errors:'' and mT5 in zero-shot and fine-tuned configurations.
\end{itemize}

The module must support execution on GPU (CUDA) via PyTorch, serialisation of trained models (using \texttt{torch.save} and \texttt{transformers.save\_pretrained}), contain exhaustive documentation of all parameters, and ensure reproducibility by fixing the random number generators and library versions.

\paragraph{Extended Task 4.} Rigorous training and evaluation of all components are performed with visualisation of the process. For each model and each task, the dedicated validation set ($15\%$ of the source corpus) is used for early stopping and monitoring the curves of the loss function and the target metric on the training and validation sets by epochs. The target metrics are: for NER --- token-level $F_1$ taking into account partial matching of entity boundaries; for keyphrases --- $F_1$ by $n$-grams (matching of unigrams, bigrams, and trigrams between the predicted and reference list); for summarisation --- ROUGE-1, ROUGE-2, and ROUGE-L; for paraphrasing --- BLEU and BERTScore; for semantic analysis --- Spearman's rank correlation coefficient between cosine distances and expert similarity assessments, as well as accuracy@$k$ for the nearest neighbour search task. Grammatical correction is evaluated separately in Task~8 using the specific metrics Accuracy and ERRANT.

For all models, training plots are constructed (dependence of the loss function and target metric on the epoch number for the training and validation sets), hyperparameter heat maps (if a search was applied), and learning curves for diagnosing overfitting. All configurations, metric values, and weights of the trained models are saved in JSON and Hugging Face/PyTorch formats.

\paragraph{Extended Task 5.} An empirical assessment of the effectiveness of the architectures and various preprocessing pipelines is performed. The results are summarised in a single table \texttt{nlp\_metrics.csv}, the rows of which correspond to the ``task -- model -- preprocessing'' combinations, and the columns to the computed metrics. Additionally, the following are constructed: box plots of the metrics across cross-validation folds for assessing stability; UMAP/$t$-SNE projections of hidden representations and sentence embeddings, making it possible to visually assess the degree of clusterability of semantically close texts; examples of generated summaries and paraphrases in comparison with reference ones; and examples of NER annotation with visualisation via displaCy. On the basis of the obtained data, conclusions are formulated about the optimal configurations for each task.

\paragraph{Extended Task 6.} The robustness of the models is ensured under complex conditions specific to each task. For NER, class balancing (weighted cross-entropy) and augmentation of the training set with synthetic examples containing rare entities are applied. For summarisation, hallucination filtering is implemented using a natural language inference (NLI) model, which assesses whether the generated statement follows from the source text; sentences for which entailment is not confirmed are marked as potential hallucinations. For semantic analysis, calibration of similarity thresholds is performed using isotonic regression on the validation set, which makes it possible to choose the optimal value of cosine distance separating duplicates and semantically distinct pairs. For grammatical correction (when fine-tuning), the ratio of positive and negative examples is controlled so that the model does not learn to ignore errors. The report compares the results with and without the application of the indicated techniques.

\paragraph{Extended Task 7.} The causes of incorrect predictions and the quality of the outputs of all components are investigated. For NER, a confusion matrix is constructed by entity categories (PER, ORG, LOC, and others), and systematically confused pairs are identified (e.g., ORG and LOC for names that may refer both to an organisation and to a city). For generation, lists of hallucinations are compiled --- sentences that the model confidently generated but that have no support in the source text, --- and their qualitative analysis is performed. For semantic analysis, reliability diagrams are constructed, visualising the correspondence of the predicted similarity to the actual proportion of relevant pairs, as well as UMAP projections of ``query--document'' pairs with colour coding of truly relevant and irrelevant matches.

Interpretation of predictions is carried out using several complementary approaches: visualisation of NER annotation via displaCy with highlighting of entities of different types; token importance maps for KeyBERT, showing which words made the greatest contribution to classifying a phrase as key; comparison of the generated summary with the extractive baseline, with highlighting of overlapping and new fragments; and analysis of the ten nearest neighbours in semantic space for a qualitative assessment of the meaningfulness of the similarity. For grammatical correction, a pairwise comparison of the distorted, corrected, and reference texts is visualised, with highlighting of successfully corrected, missed, and hypercorrected tokens. All visualisations are generated directly in the course of code execution and are included in the analytical report with extensive commentary.

\paragraph{Extended Task 8.} This task is devoted to a systematic investigation of the robustness of all developed NLP components to input distortions typical of real operation, as well as to the construction of an automatic grammatical error correction model. The learner begins by developing the software module \texttt{text\_corruptor.py}, which receives cleaned text and distortion parameters as input and returns text with controlled errors introduced. The module must provide the function \texttt{corrupt(text, language, ratio)}, where \texttt{ratio} specifies the proportion of words subjected to modification, and support at least three values: $0.1$, $0.3$, and $0.5$. The set of distortion rules must reflect the typological features of the target language and reproduce errors characteristic of real users. For the Russian language, the following rules are implemented: deletion of the soft sign in positions where it is grammatically obligatory (e.g., \emph{pisat} instead of \emph{pisat\textquotesingle}), and symmetrically --- insertion of the soft sign where it should not be (e.g., \emph{chital\textquotesingle} instead of \emph{chital}); replacement of the letter \emph{y} with \emph{i} and vice versa (e.g., \emph{riba} instead of \emph{ryba}, \emph{sistema} instead of \emph{sistema}); transposition of two adjacent letters --- swapping places (e.g., \emph{tetkst} instead of \emph{tekst}); omission of one letter in a word (e.g., \emph{kniga} without \emph{g}); duplication of a letter (e.g., \emph{knigga} instead of \emph{kniga}). For the Tajik language, rules are introduced for replacing specific letters with diacritical tails (\emph{gh, i, q, u, h, j} with diacritics) with the corresponding letters without diacritics (\emph{g, i, k, u, kh, ch}) and vice versa, as well as transposition, omission, and duplication of characters. For other languages --- Tatar, Kazakh, Armenian, Georgian, and others --- the learner independently develops equivalent rules, relying on characteristic orthographic features: for example, replacement of \emph{a} with \emph{a} in Tatar, \emph{i} with \emph{i} in Kazakh, replacement of letters with umlauts with digraphs in German when it is used as a contrastive language. Each rule is furnished with a comment in the code substantiating its linguistic relevance.

Using the developed module, distorted versions are created exclusively of the test set of the corpus; the training and validation sets are preserved in their original clean form. For each distortion level ($10\%$, $30\%$, $50\%$), a separate test set is formed, and all are saved in JSONL format with the mandatory field \texttt{corruption\_ratio} and a reference to the source document.

On the prepared data, two complementary sub-tasks are solved. The first is a quantitative robustness assessment. Each of the components trained in Tasks~4--6 (NER, keyphrase extraction, abstractive and extractive summarisation, semantic analysis) is applied to all three distorted versions of the test set, and for each distortion level, the same metrics are computed as on clean data ($F_1$ for NER and keyphrases, ROUGE-1/2/L and BERTScore for summarisation, Spearman's coefficient and accuracy@$k$ for semantic search). The results are presented in the form of a summary table, the rows of which correspond to the components and noise levels, and the columns to the metrics and the magnitude of their relative drop compared to the clean data. Additionally, a line graph is constructed, reflecting the dependence of each main metric on the proportion of distorted words; the combination of curves for all components on a single figure is permitted for a visual comparison of their robustness. The learner formulates conclusions about which of the components is the most vulnerable to noise and at what level of distortion the quality falls below a practically acceptable threshold.

The second sub-task is automatic grammatical error correction (GEC). On the basis of the generated ``distorted text $\rightarrow$ source clean text'' pairs, the learner creates a parallel corpus and applies the ruT5 model (or its analogue for the chosen language, e.g., mT5 for Tajik) in one of two permissible modes: inference with the instructional prefix ``correct the errors:'' in few-shot mode without fine-tuning, or full fine-tuning of the model on the generated parallel corpus using the standard Seq2Seq loss function. The choice of mode is substantiated in the report. Training, if performed, is carried out exclusively on the training portion of the parallel corpus; the validation portion is used for early stopping, and the test portion for the final evaluation. The quality of correction is measured by three metrics: Accuracy at the sentence level (the proportion of sentences in which all introduced distortions are corrected correctly), token-level $F_1$ (comparison of the token sequences of the corrected text with the reference), and, where technically feasible, ERRANT --- automatic classification of correction error types with a breakdown into categories: successful correction, missed error, incorrect correction, and hypercorrection (a case where the model corrects a word that was written correctly in the source text). The results for all three metrics and for all distortion levels are summarised in a separate table. Additionally, a qualitative analysis is performed: the learner selects from five to ten examples of successful correction, and the same number of examples of each error type (hypercorrection, miss, incorrect correction), and presents them in the report with brief comments on the possible causes of the model's behaviour. All results, including tables, degradation plots, and correction examples, are included in the final analytical report.

\paragraph{Extended Task 9.} A formal verification of the correctness of the data splitting and the absence of information leakage between tasks is performed. The learner documents the splitting procedure, recording the script or notebook cell with an indication of the random seed value. It is programmatically verified that the sets of document identifiers in the training, validation, and test sets do not intersect. It is separately controlled that the distorted test data (from Task~8) were not used in the training of any component, including the grammatical correction model. In parallel, a unified module \texttt{MultiTaskNLP} is developed in the form of a Python class with the methods \texttt{.fit()}, \texttt{.predict()}, and \texttt{.explain()}, compatible with PyTorch and Hugging Face. The module encapsulates the internal logic of all five types of tasks (NER, keyphrases, summarisation/paraphrasing, semantic analysis, grammatical correction), supports the serialisation of state (via \texttt{torch.save} and \texttt{transformers.save\_pretrained}), and records seeds for reproducibility checks, subject to system-dependent variation. Such an approach facilitates the integration of the components into end-to-end industrial pipelines and MLOps infrastructure. The verification protocol is included in the analytical report.

\paragraph{Extended Task 10.} An interactive web application is created using Gradio (recommended) or Streamlit, providing the user with the following capabilities: entry of arbitrary text or upload of a URL with automatic parsing of the content via the \texttt{newspaper3k} library; execution of all five types of tasks on a single document simultaneously; display of NER annotation inline with highlighting of entities of different types; output of a list of keyphrases with an indication of weights; display of the generated summary and paraphrase with an indication of ROUGE/BERTScore metrics (if a reference is available); computation of semantic similarity to another arbitrary text or to documents from the corpus; display of the result of grammatical correction with highlighting of corrected tokens; automatic generation of an HTML or PDF report with visualisations; and comparison of several models in side-by-side mode. The application is designed in such a way that it can be used by a specialist without programming skills and must be deployed locally with the possibility of subsequent deployment to Hugging Face Spaces.

\paragraph{Extended Task 11.} A suite of unit tests is developed based on the pytest framework, verifying the correctness and reproducibility of the key project components. The tests must check: the fixation of seeds and library versions; the correctness of preprocessing and data formats for each type of task; the stability of the metrics upon repeated runs of training and evaluation; the absence of information leakage between tasks (in particular, that the test data were not used in the training of any component, and that the distorted test data from Task~8 did not end up in the training set of the correction model); as well as the correctness of the operation of the \texttt{text\_corruptor.py} script --- the correspondence of the actual proportion of distorted words to the specified level for each type of distortion. All libraries used are recorded in \texttt{requirements.txt} or \texttt{environment.yml} files with an indication of the exact versions. A script is developed for the automatic execution of the full test suite and the generation of a pass/fail report. The presence of a successfully passing test suite is regarded as an integral component of the software artefact.

\subsection{Further Research Directions (Optional)}
\textbf{First Additional Task} --- comparison of the ``evolution of summarisation'': TextRank $\rightarrow$ BERT-extractive $\rightarrow$ ruT5 abstractive. An assessment is made of how quality, conciseness, and the level of hallucinations change upon the transition from extractive to abstractive methods.

\textbf{Second Additional Task} --- depth versus adaptation: comparison of BiLSTM-CRF (trained from scratch) and RuBERT (fine-tuned) on NER by training speed, quality, and robustness to domain shift.

\textbf{Third Additional Task} --- transfer from multilingual models: an assessment of the extent to which ruT5 surpasses mT5 and flan-t5 on Russian-language summarisation and paraphrasing.

\textbf{Fourth Additional Task} --- analysis of hallucinations using NLI: the application of ruBART-NLI or an analogous model for the automatic detection of inconsistent statements in generated summaries and the construction of a faithfulness classifier.

\textbf{Fifth Additional Task} --- the influence of tokenisation on NER: comparison of WordPiece (RuBERT) and Unigram (mDeBERTa) on the task of recognising rare entities, especially for morphologically complex cases.

\textbf{Sixth Additional Task} --- comparison of keyphrase extraction strategies: YAKE versus KeyBERT --- accuracy, speed, interpretability, and robustness to document length.

\textbf{Seventh Additional Task} --- reference-free evaluation of summarisation: the application of SummaC or QuestEval for assessing the quality of summaries in the absence of a reference summary, with an analysis of the correlation of these metrics with expert assessments.

\textbf{Eighth Additional Task} --- an in-depth analysis of semantic space: clustering of the corpus using ruRoBERTa embeddings --- which topics are distinguished automatically? Comparison with the annotation from Practical Work~\ref{work:3}.

\textbf{Ninth Additional Task} --- investigation of the influence of noise level on a multi-task pipeline: construction of end-to-end plots of quality degradation NER $\rightarrow$ keyphrases $\rightarrow$ summarisation $\rightarrow$ semantic search with a sequential increase in the proportion of distorted words. An analysis of which component of the pipeline is the most vulnerable and requires duplication or additional protection.

\textbf{Tenth Additional Task} --- comparison of grammatical correction models: ruT5 with instruction, fine-tuned mT5, and a simple baseline based on Levenshtein distance with a frequency dictionary. An assessment of the noise level at which the neural network model ceases to give a substantial gain compared to the simple dictionary method.

\section{Conclusion}
This work is not reducible to simply ``launching NER and a summariser in sequence''. It is the systematic design of a multi-task NLP platform, robust to the noise of the real world, where each component represents an answer to a specific challenge: how to extract facts, how to summarise them, how to find similar elements, how to correct errors, and how to unite all this into a clear, reliable, and interpretable result.

In the course of performing the work, the learner learns to pose the key questions that determine the maturity of modern NLP systems. Whether a generated fact can be trusted --- or whether it is a hallucination (a question about faithfulness). What exactly the model considers ``key'' --- and why something important was missed (a question about interpretability). How an error in NER will affect summarisation (a question about composition and pipeline robustness). To what extent the system is robust to typos and distortions inevitable in real operation (a question about robustness). Whether the model is capable of correcting an error or whether it ``propagates'' it (a question about hypercorrection). Whether reproducibility is guaranteed --- from the first token to the final PDF report (a question about scientific integrity and engineering discipline).

The answers to these questions constitute the difference between a collection of ``NLP demos'' and a responsible intelligent system capable of working not only accurately but also transparently, honestly, robustly, and reproducibly. Upon completion of the work, the learner receives not simply a set of models but an architectural compass for designing, evaluating, and deploying any complex NLP solutions --- from mass media analysis to the automated processing of user content and legal documents, capable of integrating extraction, understanding, correction, and generation into a single, reliable, and interpretable system.

\chapter{Using the Hugging Face Ecosystem}
\label{work:9}

\section{Aim and Learning Outcomes}
Students learn to use the Hugging Face model, dataset and inference ecosystem as a documented research resource rather than treating downloaded weights as inherently validated. The work focuses on resource provenance, revision identifiers, licences, explicit model capabilities and practical inference outputs, with critical attention to reproducibility and domain fit.

The three tasks below constitute the required experimental core. Larger engineering studies and independent research questions are available as optional extensions.

\section{Theoretical Background}

\noindent\textit{Principal theoretical and methodological sources for this work:} \cite{wolf2022natural,huggingface2025transformers,ru_reference_45,ru_reference_46}.\par

Modern NLP systems are increasingly being built not from scratch but as assemblies of ready-made standardised components accessible through open platforms. Hugging Face offers precisely such a development model --- modular, reproducible, and user-oriented. Its central idea consists in a unified API for any NLP task, which removes barriers between tasks and makes it possible to build hybrid pipelines without additional costs. The effectiveness of such systems depends not only on the quality of the individual components but also on their compatibility, robustness to errors, and interpretability. \cite{wolf2022natural,huggingface2025transformers}

Within the framework of the present work, four fundamental directions are distinguished, uniting the eleven official Hugging Face tasks by methodological and architectural affinity.

The first direction --- information extraction --- includes Text Classification (assigning a document to one or more categories --- news, sport, economy), Named Entity Recognition (NER: a sequence labelling task where each token is assigned a tag PER, ORG, LOC, etc.), and Feature Extraction (obtaining dense vector representations of texts, sentence embeddings, using models such as SBERT, ruSBERT, E5). Classical approaches --- CRF, TF-IDF + RAKE --- are retained as baselines for comparison; deep models, such as BiLSTM-CRF, represent the gold standard before the era of transformers; modern solutions rely on classification heads on top of RuBERT, ruRoBERTa, and on feature extraction via multilingual E5.

The second direction --- text generation --- covers Summarisation (compressing text to its essence, extractive and abstractive), Translation (transforming text from one language to another), Text Generation (creating new text based on a prompt or context), and Fill-Mask (restoring missing context from a [MASK] token). For the Russian language, the most relevant models are ruT5, flan-t5-base-ru, mT5-large, and Helsinki-NLP, which support zero-shot and few-shot configurations.

The third direction --- semantic analysis --- unites Sentence Similarity (cosine distance between the embeddings of two sentences), Text Ranking (assessing the relevance of documents to a query by encoding the query and document), Zero-Shot Classification (assigning to arbitrary labels without training, using NLI models), and Question Answering (extracting an exact answer from a given context). The calibration of similarity is critically important: not all pairs with high similarity are semantically equivalent, and not all QA answers are factually correct.

The fourth direction --- hybrid architectures --- presupposes the construction of pipelines in which the output of one task (e.g., an embedding obtained via Feature Extraction) becomes the input of another (ranking or clustering). The use of a single platform (the pipeline API) reduces barriers between tasks and simplifies composition; however, it requires strict coordination of data formats, tokenizer compatibility, and management of inference latencies.

What distinguishes the Hugging Face ecosystem from a simple collection of pre-trained models is the standardisation imposed on all the enumerated tasks. One and the same pipeline interface --- with methods such as \texttt{load}, \texttt{predict}, and \texttt{save} --- serves both text classification, and NER, and summarisation, and translation. For the practitioner, this means that switching from one task to another does not require learning a new library or data format; it is only necessary to choose a different model and interpret a different output. This standardisation is pedagogically valuable because it allows the learner to concentrate on the conceptual differences between tasks --- what distinguishes generation from extraction, what makes zero-shot classification fragile --- rather than on the mechanical details of loading and inference. The present work is structured so as to use this property: by implementing all eleven tasks within a single interface, the learner directly experiences both the power and the limitations of a unified API.

The quality of the solution depends not only on the model but also on the correct formulation of the task, the choice of metric, and the analysis of the specifics of errors --- for example, hallucinations in generation (the introduction of facts absent from the source), instability of zero-shot labels upon synonymous replacement of classes, or misses in NER for rare entities. It is for this reason that interpretability and error analysis are not an optional feature but a mandatory part of modern NLP research.

\subsection{Experimental Interpretation}
Published model repositories are mutable. Record a commit revision or content hash where feasible; identify model weights, tokenizer and evaluation data independently. Access conditions, licences and availability may change, and the existence of a model card is not proof of suitability for every language or domain.

\section{Execution Procedure}

\subsection{Core: Three Required Tasks}
\begin{enumerate}[leftmargin=*,label=\textbf{\arabic*.}]

\item Locate two published resources in the Hugging Face ecosystem and document their task, version/revision, licence and conditions of access.

Find at least one model and one dataset or another pair of distinct resources, and consult their published documentation rather than relying only on a search-result title. Record author or organisation, identifier, revision, intended task, language, training/evaluation provenance when available, access restrictions, and licence. Identify any missing documentation as an uncertainty, not as proof that the resource is unrestricted. Explain how the chosen resources correspond to the educational task.

\item Execute one documented pipeline on a real text sample, saving inputs, outputs and execution settings. Compare it with a simple baseline or a second documented configuration.

Run a documented pipeline on licensed, non-sensitive real text and preserve input examples, prediction labels or generated text, and configuration parameters. Choose a meaningful reference: a rule-based baseline, a published alternative checkpoint, or a controlled second configuration that changes one variable. Where a model produces probabilities or scores, describe what those numbers represent and whether they are calibrated. Do not infer model effectiveness from a single illustrative output or silently switch revision during comparison.

\item Create a brief evidence-based model/data provenance record that states known limitations, compute requirements and restrictions on redistribution.

Prepare a concise evidence table linking each output and resource to its model card or dataset card, the run configuration and the observed failure modes. Identify potential domain mismatch, language support limitations and known bias or licence constraints. Report measured runtime only with the environment and input size; list expected memory needs separately if they were not measured. State whether the run required downloading external assets and how another reader can identify the same revision.

\end{enumerate}

\subsection{Reproducible Example and Its Limits}

The companion project contains executable code for this work. Run the real-data path with
\begin{center}\texttt{python practikum\_examples.py 9}\end{center}
A published Hugging Face model and real documents; availability and revisions must be checked at execution. For a technical-only check, add \texttt{-{}-smoke}; its synthetic inputs must not be used as scientific evaluation results. The program does not contain fabricated model scores. Record actual outputs, sample sizes, versions, timings and failures after running it.

\subsection{Extended Engineering Tasks (Optional)}
The following tasks provide a fuller engineering and research programme. They are optional rather than prerequisites for passing the core. The numerical scales and model choices below describe ambitious projects; they are not universal requirements. Documentation and reporting follow the shared protocol above.

\paragraph{Extended Task 1.} On the basis of the text corpus formed during Practical Work~\ref{work:1} and the labelling prepared in Practical Work~\ref{work:4} and Work~\ref{work:5}, a unified experimental corpus is formed for eleven types of NLP tasks supported by the Hugging Face ecosystem: Text Classification, Named Entity Recognition (NER), Question Answering, Zero-Shot Classification, Machine Translation, Automatic Summarisation, Feature Extraction, Text Generation, Fill-Mask, Sentence Similarity, and Text Ranking. Each document is saved as a JSON object with the fields: \texttt{id} --- unique document identifier; \texttt{text} --- full text after cleaning; \texttt{title} --- heading (optional); \texttt{url} --- source URL (optional); \texttt{date} --- publication date in ISO 8601 format (optional). For the classification and NER tasks, the \texttt{sentiment} and \texttt{category} fields obtained in Practical Work~\ref{work:4} and Work~\ref{work:5} are used. Additionally, a synthetic dataset is formed for the Table Question Answering (Table QA) task, consisting of triples ``table -- context -- question'' and saved as a separate JSONL file.

The minimum total volume of the corpus is ten thousand documents. If the preparation of data for all eleven types of tasks in full is objectively difficult --- for example, due to the high cost of expert annotation for QA, the limited volume of the source corpus, or the technical complexity of creating a synthetic Table QA dataset, --- the learner has the right, in agreement with the instructor, to reduce the number of labelled documents or to confine themselves to eight to ten types of tasks, explicitly stipulating this circumstance in the datasheet and discussing its influence on the statistical significance and generalisation ability of the obtained models in the analytical report.

All text materials are preserved in their original form, without the application of lemmatization or stemming, since the pre-trained tokenizers of Hugging Face models are optimised for the original word forms and subword units. For statistical methods (RAKE, YAKE), used as baselines, minimal cleaning (removal of stop words and punctuation) is permitted. The data are saved in JSONL format with UTF-8 encoding.

The split into training, validation, and test sets is performed in a $70/15/15$ ratio. To ensure the representativeness of all classes in each of the subsets, a stratified split is applied with respect to the target variable specific to each task: for classification --- with respect to the \texttt{sentiment} field (binary) or \texttt{category} field (multi-class); for NER --- by the distribution of entity tags. For the synthetic Table QA dataset, the ``table -- context -- question'' triples are distributed in such a way that all questions relating to a single table end up in the same set, which prevents information leakage. The random seed value used in the split is fixed and documented to support repeatability, subject to library, driver and hardware constraints.

In view of the significant computational demands imposed by the inference of large models (in particular, generative architectures), the learner is recommended to use the Google Colab environment with a GPU accelerator or other cloud platforms. For models requiring resources exceeding the capabilities of the local or Colab environment (e.g., \texttt{sberbank-ai/ruGPT-3.5-13B}), the use of the Hugging Face Inference API is permitted, with an explicit indication of this fact in the methodological section of the report and a discussion of its influence on reproducibility.

References to the specific artefacts of the previous works (the path to the corpus file, the datasheet, the chosen tokenisation and normalisation schemes from Work~\ref{work:1}, the labelled datasets from Works~\ref{work:4} and Work~\ref{work:5}, the pre-trained embeddings from Work~\ref{work:2}, the fine-tuned transformers from Work~\ref{work:6}, the transfer learning strategies from Work~\ref{work:7}, the multi-task architecture from Work~\ref{work:8}) are recorded in the methodological section of the report to ensure full traceability of the experiment. If the learner performs the present work independently of the previous ones, they must first create or select a corpus and perform its labelling satisfying the requirements of Works~\ref{work:1}, Work~\ref{work:4}, and Work~\ref{work:5}, and explicitly indicate their origin and characteristics in the datasheet.

\paragraph{Extended Task 2.} Several independent but compatible data representations are created depending on the task. For most transformer tasks, tokenisation is performed via \texttt{AutoTokenizer.from\_pretrained} of the corresponding model. For statistical methods --- RAKE and YAKE --- raw text with minimal cleaning is used. For zero-shot classification, lists of candidate labels (arbitrary labels not used in training) are prepared, as well as their synonymous variants for assessing stability. All representations are saved with metadata in JSON format, including the preprocessing strategy and a reference to the source document.

\paragraph{Extended Task 3.} A configurable software module \texttt{hf\_pipelines.py} is developed, providing a unified interface for all eleven tasks. The supported components include: for classification --- \texttt{cointegrated/rubert-tiny2}, \texttt{DeepPavlov/rubert-base-cased}; for NER --- \texttt{DeepPavlov/rubert-base-cased-ner}, \texttt{mdeberta-v3-base}; for Question Answering (QA) --- \texttt{deepset/roberta-base-squad2}, \texttt{AlexKay/xlm-roberta-large-qa-multilingual}; for zero-shot classification --- \texttt{facebook/bart-large-mnli}, \texttt{MoritzLaurer/mDeBERTa-v3-base-xnli-multilingual}; for translation --- \texttt{Helsinki-NLP/opus-mt-ru-en}, \texttt{facebook/nllb-200}; for summarisation --- \texttt{IlyaGusev/rut5\_base\_sum\_gen}, \texttt{cointegrated/ruT5-base}; for feature extraction --- \texttt{intfloat/multilingual-e5-large}, \path{sentence-transformers/paraphrase-multilingual-MiniLM}; for text generation --- \texttt{sberbank-ai/ruGPT-3.5-13B} (accessed via Inference API); for Fill-Mask --- \texttt{DeepPavlov/rubert-base-cased}; for sentence similarity --- \texttt{cointegrated/ruRoBERTa-large-sts}; for ranking --- \texttt{intfloat/multilingual-e5-large} for encoding the query and document. The module supports execution on GPU, serialisation of output data, thorough documentation of parameters, and traceability through recorded seeds, library versions and hardware details.

\paragraph{Extended Task 4.} Rigorous training and evaluation are performed with visualisation of the process. A separate validation set ($15\%$) is used for monitoring. The target metrics are: for classification and NER --- $F_1$ (macro); for QA --- Exact Match and $F_1$; for summarisation --- ROUGE-1/2/L; for translation --- BLEU; for zero-shot classification --- accuracy on an expert gold subset; for semantic tasks --- Spearman's rank correlation coefficient and accuracy@$k$. Plots of the metrics, error distributions, and hyperparameter heat maps are constructed. All configurations, weights, and metrics are saved in JSON and Hugging Face formats.

\paragraph{Extended Task 5.} An empirical assessment of the effectiveness of the architectures and pipelines is performed. The results are summarised in a final table \texttt{hf\_metrics.csv}, the rows of which correspond to the ``task -- model'' combinations, and the columns to the metrics. Additionally, box plots across cross-validation folds, UMAP/$t$-SNE projections of embeddings, as well as examples of generation and NER annotation are constructed.

\paragraph{Extended Task 6.} The robustness of the models is ensured under complex conditions: rare entities in NER, long documents in summarisation, ambiguous formulations in QA, domain shift in zero-shot classification. Hallucination filtering is applied using natural language inference (NLI) models, calibration of semantic similarity via isotonic regression, and class balancing in classification tasks.

\paragraph{Extended Task 7.} The causes of incorrect predictions and the quality of the outputs are investigated. Confusion matrices are constructed for NER and classification; lists of hallucinations in generation (with fact-checking against the source); analysis of zero-shot instability (label flipping upon reformulation of candidate labels); reliability diagrams for semantic similarity; UMAP projections of ``query--document'' pairs. Interpretation is carried out through HTML highlighting of entities, contextual highlighting of answers in QA, display of confidence scores, analysis of nearest neighbours in the embedding space, and importance maps.

\paragraph{Extended Task 8.} A formal verification of the correctness of the data splitting and the absence of information leakage between tasks is performed. The learner documents the splitting procedure with the fixation of the random seed, and programmatically verifies the non-intersection of document identifiers between all sets (training/validation/test). It is controlled that the synthetic data for Table QA do not intersect with the documents used for training other components, and that the candidate labels for zero-shot were not extracted from the test documents. The verification protocol is included in the analytical report.

\paragraph{Extended Task 9.} A unified module \texttt{HuggingFaceNLP} is developed with the methods \texttt{.predict(task, input)} and \texttt{.explain()}, compatible with the \texttt{transformers} library and designed for industrial deployment. The module encapsulates the logic of all eleven types of tasks, supports the serialisation of state, and records seeds but remains subject to platform-dependent nondeterminism, facilitating integration into MLOps infrastructure.

\paragraph{Extended Task 10.} An interactive web application is created using Gradio, allowing the user to: enter text or upload a URL (with parsing via \texttt{newspaper3k}); execute all eleven tasks on a single document; display NER annotation inline, extracted QA answers, ranking, generated text with ROUGE and BERTScore metrics; compute semantic similarity to another text or to documents from the corpus; generate a comprehensive HTML or PDF report; and compare several models side-by-side.

\paragraph{Extended Task 11.} A suite of unit tests is developed based on pytest, verifying correctness and reproducibility. The tests check: the fixation of library  (\texttt{transformers}, \texttt{evaluate}); the correctness of loading models by identifiers; the stability of the metrics upon repeated runs; the absence of information leakage between tasks (in particular, that the test data were not used in the training of any component). All dependencies are recorded in \texttt{requirements.txt}. A script for the automatic execution of the tests and the generation of a pass/fail report is attached.

\subsection{Further Research Directions (Optional)}
\textbf{First Additional Task} --- multilinguality: testing models on texts in Tatar, Kazakh, and other languages of the peoples of Russia in translation and classification tasks, with an assessment of quality losses compared to Russian-language data.

\textbf{Second Additional Task} --- hybrid scenarios: construction of a pipeline ``NER $\rightarrow$ Summarisation'', in which the extracted named entities are explicitly included in the summarizer's prompt, and an assessment of whether this increases the preservation of key facts in the summary.

\textbf{Third Additional Task} --- CPU optimisation: comparison of distilled models (rubert-tiny, distilbert-base-multilingual) with full-size ones by latency and energy consumption.

\textbf{Fourth Additional Task} --- comparison with Inference API: comparison of local inference and cloud inference (Hugging Face Inference API) by cost, latency, and quality.

\textbf{Fifth Additional Task} --- automatic corpus annotation: application of NER and classification to the entire corpus, construction of entity and category distributions.

\textbf{Sixth Additional Task} --- Human-in-the-loop: development of an interface for manual correction of zero-shot labels with subsequent fine-tuning of the model on the corrected data.

\textbf{Seventh Additional Task} --- energy efficiency assessment: comparison of the CO\textsubscript{2} footprint (via the \texttt{codecarbon} library) for different architectures when performing an identical set of tasks.

\textbf{Eighth Additional Task} --- zero-shot NER: the use of mDeBERTa-v3 in zero-shot mode for extracting user-defined entity types (e.g., ``technology companies'', ``government bodies'') without any training.

\section{Conclusion}
This work is not simply ``launching eleven pipelines from the documentation''. It is the mastery of Hugging Face as a tool for engineering and scientific activity, where each task represents an opportunity to understand how modern NLP systems are constructed ``under the hood'' and how they can be reliably combined into complex applications.

In the course of performing the work, the learner learns to ask the key questions that determine the maturity of modern NLP systems. Whether a zero-shot prediction can be trusted --- or whether the model is simply guessing (a question about reliability). How much time and energy each prediction costs --- and whether it is justified (a question about efficiency). How to avoid hallucinations when generating based on news text (a question about faithfulness). Whether reproducibility is guaranteed --- from the model identifier to the \texttt{transformers} version (a question about scientific integrity and engineering discipline).

The answers to these questions constitute the difference between a ``Colab demo'' and a reliable, explainable, reproducible NLP system, ready for integration into real projects. Upon completion of the work, the learner receives not simply a set of scripts but a platform competence --- the ability to quickly, responsibly, and effectively design, evaluate, and deploy any NLP solutions in an open ecosystem, which is a fundamental skill in the era of modular artificial intelligence.

\chapter{Prompt Engineering: Design, Measurement and Robustness}
\label{work:10}

\section{Aim and Learning Outcomes}
This work studies prompting as an experimental design problem. Students compare controlled zero-shot and few-shot prompts on exactly the same data, measure malformed outputs explicitly, and examine the sensitivity of language models to examples and phrasing. Further research can extend to structured output, reasoning methods and robustness without assuming that fluent explanations imply verified reasoning.

The three tasks below constitute the required experimental core. Larger engineering studies and independent research questions are available as optional extensions.

\section{Theoretical Background}

\noindent\textit{Principal theoretical and methodological sources for this work:} \cite{brown2020language,liu2023pre,wei2022chain,ru_reference_50,ru_reference_51,ru_reference_52,ru_reference_53,arabov2026rusfinchain}.\par

For verifiable multi-step reasoning, the RusFinChain benchmark illustrates the value of independent reference checks rather than relying only on a model-generated rationale \cite{arabov2026rusfinchain}.

Prompt engineering has emerged as a critical discipline in the era of large language models. Unlike traditional machine learning, where models are trained on labelled data, prompting allows practitioners to adapt LLMs to new tasks by carefully designing the input text --- the prompt --- that conditions the model's generation. The effectiveness of a prompt determines not only the quality of the output but also the model's robustness, safety, and efficiency. Mastering prompt engineering transforms the practitioner from a passive API user into an active shaper of model behaviour. \cite{liu2023pre,wei2022chain,brown2020language}

The taxonomy of prompting techniques spans several levels of complexity. At the most basic level, zero-shot prompting simply presents the task description to the model without examples. This approach is surprisingly effective for well-trained LLMs, particularly for tasks that align with the model's pre-training objectives. Few-shot prompting extends zero-shot by providing a small number of exemplars --- demonstrations of the input-output mapping --- which significantly improves performance, especially for tasks with complex or unfamiliar formats. Instruction-based prompting frames the task as a command or instruction, leveraging the model's instruction-following capabilities developed during instruction tuning. System prompts separate the model's role and behaviour from the user's query, enabling consistent persona and behaviour across interactions.

Advanced reasoning techniques unlock the model's ability to solve complex problems that require multi-step reasoning. Chain-of-Thought (CoT) prompting encourages the model to generate intermediate reasoning steps before producing the final answer. This technique, introduced by Wei et al. (2022), dramatically improves performance on arithmetic, commonsense, and symbolic reasoning tasks. Tree-of-Thought (ToT) generalises CoT by allowing the model to explore multiple reasoning paths simultaneously, maintaining a tree of thoughts and using search algorithms (BFS, DFS) to find the optimal path. Self-Consistency improves CoT by generating multiple reasoning chains and selecting the most consistent answer through majority voting. ReAct (Reasoning + Acting) combines CoT with tool use, enabling the model to interact with external environments and APIs, which is particularly relevant for agent systems.

Structured output generation is a crucial capability for integrating LLMs into production systems. By designing prompts that specify the output format --- JSON, XML, CSV, or custom schemas --- practitioners can ensure that model outputs are machine-readable and easily parsed. Techniques include format-specific instructions, few-shot examples of the desired format, and post-processing for validation and correction.

Prompt optimisation is an active area of research that aims to automate the design of effective prompts. Prompt Tuning trains a small set of continuous vectors (soft prompts) that are prepended to the input, while the base model remains frozen. Prefix Tuning extends this idea by training continuous vectors for each layer of the transformer. AutoPrompt uses gradient-based search to discover discrete prompts that maximise performance on a validation set. These methods make it possible to adapt LLMs to new tasks without the computational cost of full fine-tuning, while also providing a degree of interpretability that is lacking in black-box optimisation.

Safety and robustness are critical considerations in prompt engineering. Prompt injection attacks attempt to override the system's instructions by injecting malicious commands into the user input. Defensive strategies include input sanitisation (removing or escaping special characters), role-based separation (clearly demarcating system and user instructions), adversarial prompt detection (using classifiers to identify malicious prompts), and output filtering (validating and sanitising model outputs). Prompt leakage --- the unintentional exposure of system prompts or proprietary information --- is another important concern that requires careful prompt design and access control.

The evaluation of prompt quality is a multi-faceted challenge. Accuracy metrics measure the correctness of outputs relative to ground truth. Robustness metrics assess the stability of performance under variations in input phrasing, typos, and adversarial perturbations. Consistency metrics evaluate whether the model produces consistent outputs for semantically equivalent inputs. Safety metrics detect harmful, biased, or inappropriate outputs. Efficiency metrics measure token consumption, latency, and cost. A comprehensive evaluation framework combines these metrics to provide a holistic assessment of prompt quality, enabling systematic improvement and comparison.

\subsection{Experimental Interpretation}
Prompt comparison requires fixed test inputs, an agreed evaluation rubric and an explicit policy for malformed or abstaining outputs. Few-shot demonstrations must come from the development data; otherwise the comparison leaks evaluation information. Chain-of-thought text is not an independently verified trace of model computation, and models must not be credited with successful reasoning solely because their explanations sound plausible.

\section{Execution Procedure}

\subsection{Core: Three Required Tasks}
\begin{enumerate}[leftmargin=*,label=\textbf{\arabic*.}]

\item Using a labelled training corpus and its untouched test partition, construct zero-shot and few-shot prompts for the same clearly defined classification task. Select few-shot demonstrations only from training documents.

Define a closed set of classification labels and a scoring protocol before creating prompts. Build one zero-shot instruction and at least one few-shot variant with demonstrations selected exclusively from the training portion; record example order, language and class balance. Use a fixed common test subset that is not represented among demonstrations. If no suitable annotated real corpus is available, use the documented companion example solely as a programming demonstration rather than reporting a benchmark claim.

\item Run a specified generative model with controlled decoding for each template. Define a strict output parser, counting invalid labels as errors instead of silently discarding them.

Choose a named generative checkpoint and record its revision and access route, temperature or sampling settings, maximum output length and any context-window restrictions. Keep these settings constant when comparing prompt templates, unless a setting is deliberately the variable under study. Implement an exact parser for the required label schema and preserve raw model replies alongside parsed labels. Count refusals, multiple labels and malformed responses as explicit outcomes so that filtering cannot artificially improve quality.

\item Compare accuracy and macro-F1 (including invalid-output counts) on identical examples and inspect failure modes. Explicitly discuss sensitivity to examples, language and model revision.

Evaluate all variants on exactly the same item identifiers and report accuracy, macro-F1, per-class support and the fraction of invalid outputs. Inspect concrete failures involving ambiguous inputs, prompt wording and language, avoiding examples chosen solely to favour one template. When the model endpoint is nondeterministic, document repeated runs or state the limitation. Discuss prompt length, inference cost and potential exposure of source examples; do not claim that improvements from one prompt transfer to other checkpoints without testing them.

\end{enumerate}

\subsection{Reproducible Example and Its Limits}

The companion project contains executable code for this work. Run the real-data path with
\begin{center}\texttt{python practikum\_examples.py 10}\end{center}
20 Newsgroups and google/flan-t5-small; test labels are not included in the prompts. For a technical-only check, add \texttt{-{}-smoke}; its synthetic inputs must not be used as scientific evaluation results. The program does not contain fabricated model scores. Record actual outputs, sample sizes, versions, timings and failures after running it.

\subsection{Extended Engineering Tasks (Optional)}
The following tasks provide a fuller engineering and research programme. They are optional rather than prerequisites for passing the core. The numerical scales and model choices below describe ambitious projects; they are not universal requirements. Documentation and reporting follow the shared protocol above.

\paragraph{Extended Task 1.} On the basis of the text corpus formed during Practical Work~\ref{work:1} and the labelled datasets from Practical Work~\ref{work:4} and Work~\ref{work:5}, a comprehensive evaluation dataset for prompt engineering is created. The dataset must include at least one thousand examples spanning five task types: classification (sentiment, topic), extraction (NER, keyphrase), generation (summarisation, paraphrasing), reasoning (arithmetic, commonsense, logical), and structured output (JSON, table generation). Each example is saved in JSONL format with the fields: \texttt{id} --- unique identifier; \texttt{task\_type} --- task category; \texttt{prompt\_template} --- a template for the task (e.g., ``Classify the following text as positive or negative: \{text\}''); \texttt{input} --- the input text; \texttt{expected\_output} --- the reference answer; \texttt{metadata} --- additional information (difficulty, domain, etc.). The dataset is split into training (for prompt development), validation (for prompt tuning), and test (for final evaluation) sets in a $60/20/20$ ratio. The random seed is fixed and documented to ensure reproducibility.

\paragraph{Extended Task 2.} A software module \texttt{prompt\_utils.py} is developed, providing essential utilities for prompt engineering. The module must include: a prompt template engine with support for variables and conditional logic; functions for token counting (using the tokenizer of the target model); functions for parsing structured outputs (JSON, CSV, XML); functions for calculating metrics (accuracy, ROUGE, BERTScore); a logging system for tracking prompt versions, inputs, outputs, and metrics; and a prompt library with support for versioning, metadata, and search. The module is designed to be model-agnostic and compatible with both open-source models (from Hugging Face) and API-based models. The code is furnished with exhaustive documentation describing the interface of all public functions and configuration parameters.

\paragraph{Extended Task 3.} A comprehensive comparison of zero-shot, few-shot, and instruction-based prompting is performed. The learner selects at least three open LLMs (e.g., Llama-3-8B-Instruct, Mistral-7B-Instruct, Qwen1.5-7B-Chat) and evaluates them on all five task types from Task 1. For zero-shot prompting, the learner designs a simple instruction prompt (e.g., ``Classify the sentiment of the following text. Respond with 'positive' or 'negative'.''). For few-shot prompting, the learner includes 2, 4, and 8 examples from the training set, analysing how performance scales with the number of examples. For instruction-based prompting, the learner experiments with different instructional phrasing (imperative, descriptive, interrogative) and analyses their effect on performance. For each model, prompting technique, and task type, the following metrics are computed: accuracy, F1 (for classification and extraction), ROUGE (for generation), and token consumption. The results are summarised in a table \texttt{prompting\_comparison.csv}. Learning curves are constructed showing performance as a function of the number of few-shot examples.

\paragraph{Extended Task 4.} Advanced reasoning techniques are implemented and compared: Chain-of-Thought (CoT), Tree-of-Thought (ToT), and Self-Consistency. For CoT prompting, the learner designs prompts that explicitly ask the model to ``think step by step'' or provide a reasoning template (e.g., ``Step 1: ... Step 2: ... Therefore, the answer is ...''). For ToT prompting, the learner implements a simple search algorithm (BFS or DFS) that generates multiple reasoning paths and selects the best one based on a scoring function (e.g., the model's confidence or a separate verifier). For Self-Consistency, the learner generates 5--10 CoT reasoning chains and selects the most frequent final answer. The evaluation is performed on reasoning tasks (arithmetic, commonsense, logical) from Task 1. The metrics include accuracy, average number of tokens per reasoning chain, and inference time. The results are visualised in a bar chart comparing the performance of zero-shot, CoT, ToT, and Self-Consistency across models and tasks.

\paragraph{Extended Task 5.} Structured output generation is investigated. The learner designs prompts for three types of structured outputs: JSON (e.g., ``{`sentiment': 'positive', `confidence': 0.95}''), CSV (e.g., ``positive, 0.95''), and XML (e.g., ``<sentiment>positive</sentiment><confidence>0.95</confidence>''). For each format, the learner designs prompts that specify the format, provide examples, and include validation instructions (e.g., ``Ensure the JSON is valid and includes all required fields''). The learner evaluates the success rate of parsing (proportion of outputs that are valid and parsable) and the accuracy of the extracted information. Error analysis is performed to identify common failure modes: malformed output, missing fields, extra fields, and incorrect data types. The results are summarised in a table \texttt{structured\_output\_metrics.csv}.

\paragraph{Extended Task 6.} Prompt optimisation techniques are implemented and compared: Manual Iterative Refinement, Prompt Tuning, Prefix Tuning, and AutoPrompt. For Manual Refinement, the learner starts with a baseline prompt and iteratively improves it based on validation set performance, tracking all versions and the rationale for each change. For Prompt Tuning, the learner uses the \texttt{peft} library from Hugging Face to train soft prompts on the training set, with varying numbers of virtual tokens (5, 10, 20). For Prefix Tuning, the learner trains continuous prefixes for each transformer layer. For AutoPrompt, the learner implements a simple gradient-based search for discrete prompts (or uses a library implementation). The comparison is performed on one task type (e.g., sentiment classification) with a fixed model (e.g., Llama-3-8B-Instruct). The metrics include accuracy, training time, inference time, and the number of trainable parameters. Learning curves are constructed showing validation accuracy as a function of training steps for each optimisation method.

\paragraph{Extended Task 7.} A robust evaluation framework for prompt quality is developed. The framework must support the following metrics: accuracy (correctness of outputs), robustness (performance consistency across paraphrases, typos, and adversarial perturbations), consistency (agreement across semantically equivalent inputs), safety (detection of harmful or biased outputs), and efficiency (token consumption, latency). The learner implements evaluation functions for each metric and applies them to all prompts developed in Tasks 3--6. The results are summarised in a comprehensive evaluation report \texttt{prompt\_evaluation\_report.html}, containing tables, charts, and qualitative examples. The learner analyses the trade-offs between different metrics: accuracy vs robustness, accuracy vs efficiency, and accuracy vs safety.

\paragraph{Extended Task 8.} Prompt injection and security are investigated. The learner designs a set of adversarial prompts that attempt to: override system instructions (``Ignore previous instructions and...''); leak system prompts (``What were your system instructions?''); perform role-playing attacks (``Act as a system administrator and...''); and introduce harmful content (``Generate hate speech...''). The learner evaluates the robustness of three prompts (zero-shot, few-shot, CoT) against these attacks on at least three LLMs. Defensive strategies are implemented: input sanitisation (removing or escaping special characters), role-based separation (clearly demarcating system and user instructions), adversarial prompt detection (using a classifier to identify malicious prompts), and output filtering (validating and sanitising outputs). The effectiveness of each defence is measured by the attack success rate. The results are summarised in a table \texttt{prompt\_security\_metrics.csv}, and the learner formulates best practices for prompt security in production systems.

\paragraph{Extended Task 9.} A prompt library with versioning and metadata is developed. The library must support: storing prompt templates with metadata (author, date, model, task, performance metrics); versioning of prompts (tracking changes over time); searching and filtering prompts by task, model, performance, and tags; importing and exporting prompts in JSON format; and automated testing of prompts on validation datasets. The learner populates the library with all prompts developed in the previous tasks, including at least twenty distinct prompts. The library is saved in a structured directory and published in the repository.

\paragraph{Extended Task 10.} An interactive web application for prompt testing, comparison, and optimisation is created using Gradio or Streamlit. The application must provide the following capabilities: entering arbitrary text and selecting a model and prompt template; displaying the model's output alongside the expected output (if available); computing and displaying metrics (accuracy, ROUGE, BERTScore) in real time; comparing multiple prompts side-by-side on the same input; saving and loading prompt templates; and automatically generating an HTML report with the results of the comparison. The application is designed in such a way that it can be used by a domain expert without programming skills and is deployed locally with the possibility of subsequent deployment to Hugging Face Spaces.

\paragraph{Extended Task 11.} A suite of unit tests is developed based on the pytest framework, verifying the correctness and reproducibility of the key project components. The tests must check: the correctness of the prompt template engine (variables are correctly replaced); the correctness of token counting (matches the model's tokenizer); the correctness of structured output parsing (valid JSON, CSV, XML); the correctness of metric calculations (accuracy, ROUGE, BERTScore); the correctness of the prompt library (versioning, search); and the stability of prompt evaluation results upon repeated runs. All dependencies are recorded in the \texttt{requirements.txt} file with an indication of the exact versions. A script is developed for the automatic execution of the full test suite and the generation of a pass/fail report.

\subsection{Further Research Directions (Optional)}
\textbf{First Additional Task} --- multilingual prompting: comparison of prompting techniques across Russian, Tatar, and English; analysis of language transfer and cross-lingual generalisation.

\textbf{Second Additional Task} --- few-shot example selection: investigation of the influence of example selection strategies (random, semantic similarity, diversity-based) on few-shot performance; development of a selection algorithm.

\textbf{Third Additional Task} --- prompt compression: development of techniques for compressing prompts (reducing token count) while preserving performance; evaluation of token efficiency.

\textbf{Fourth Additional Task} --- ensemble prompting: combination of multiple prompts or multiple models via voting or averaging; analysis of ensemble diversity and performance gains.

\textbf{Fifth Additional Task} --- multimodal prompting: extension of prompting techniques to multimodal models (e.g., LLaVA, CLIP); development of prompts that combine text and image inputs.

\textbf{Sixth Additional Task} --- prompt-based data augmentation: use of LLMs to generate synthetic training data for downstream tasks; evaluation of data quality and downstream performance.

\textbf{Seventh Additional Task} --- dynamic prompting: development of techniques that adapt the prompt based on the input (e.g., selecting the most relevant few-shot examples); implementation and evaluation of a dynamic prompt selection system.

\textbf{Eighth Additional Task} --- human-in-the-loop prompt engineering: development of an interface for interactive prompt refinement with human feedback; integration of human feedback into the prompt optimisation loop.

\section{Conclusion}
This work is not simply ``writing prompts''. It is the systematic mastery of a fundamental skill for modern AI practitioners: the ability to communicate effectively with large language models, to shape their behaviour, to extract their full potential, and to protect them (and their users) from harm. Prompt engineering is the interface between human intent and machine capability, and mastering it unlocks the ability to rapidly prototype, evaluate, and deploy AI solutions without the cost and complexity of training.

In the course of performing the work, the learner learns to ask the key questions that determine the maturity of prompt engineering practice. Whether the prompt reliably produces the desired output across different inputs and models (a question about robustness). Whether the prompt can be easily understood and modified by other practitioners (a question about clarity and documentation). Whether the prompt is safe from adversarial attacks and unintended behaviours (a question about security). Whether the prompt is efficient in terms of token consumption and cost (a question about efficiency). Whether the prompt can be systematically improved through optimisation and evaluation (a question about engineering discipline).

The answers to these questions constitute the difference between a `hacky prompt` and a professionally engineered prompt, ready for integration into production systems. Upon completion of the work, the learner receives not simply a set of prompts but a methodological framework for designing, evaluating, and deploying prompt-based solutions --- from quick prototypes to production-grade systems capable of reliably extracting value from large language models.

\chapter{Multimodal NLP: Text, Images and Speech}
\label{work:11}

\section{Aim and Learning Outcomes}
The objective is to investigate how language representations interact with visual inputs in a well-defined measurement protocol. Students evaluate CLIP-based image--text alignment on actual images, explore the effect of prompt phrasing and error categories, and recognise why retrieval, captioning, visual question answering and speech recognition require distinct datasets and metrics.

The three tasks below constitute the required experimental core. Larger engineering studies and independent research questions are available as optional extensions.

\section{Theoretical Background}

\noindent\textit{Principal theoretical and methodological sources for this work:} \cite{radford2021clip,ru_reference_56,li2022blip,ru_reference_58,ru_reference_59,liu2023llava}.\par

The world is inherently multimodal: humans perceive and communicate through language, vision, sound, touch, and other modalities. While traditional NLP has focused exclusively on text, the emergence of large-scale multimodal models has opened new frontiers where language interfaces with the visual and auditory world. Multimodal NLP integrates computer vision, speech processing, and natural language processing into unified systems capable of understanding and generating content across modalities. This integration enables applications such as image captioning, visual question answering, text-to-image generation, speech recognition, and cross-modal retrieval. \cite{radford2021clip,li2022blip,liu2023llava}

The taxonomy of multimodal learning distinguishes several key dimensions. Modalities refer to the types of data: text, image, video, audio, and others. Fusion strategies determine how information from different modalities is combined: early fusion (concatenating features at the input level), late fusion (combining decisions from modality-specific models), and hybrid fusion (intermediate integration, e.g., cross-attention). Cross-modal alignment ensures that representations from different modalities occupy a shared semantic space, enabling comparison and retrieval across modalities. Joint training (or multi-task training) optimises a single model for multiple modalities and tasks, leveraging shared representations and transfer learning.

Vision-language models have emerged as a dominant paradigm in multimodal NLP. CLIP (Contrastive Language-Image Pre-training) learns a shared embedding space for text and images through contrastive learning on a large dataset of image-text pairs. CLIP enables zero-shot classification, cross-modal retrieval, and serves as a foundation for more complex multimodal systems. LLaVA (Large Language and Vision Assistant) extends CLIP by connecting a vision encoder to a large language model (LLM) through a projection layer, enabling visual instruction following, image understanding, and reasoning. BLIP (Bootstrapping Language-Image Pre-training) jointly pre-trains a vision-language model for both understanding and generation tasks, achieving state-of-the-art performance on image captioning, VQA, and retrieval. Flamingo introduces a few-shot learning approach for vision-language tasks, using a frozen vision encoder and LLM with trainable cross-attention layers. These models demonstrate the power of combining vision encoders with LLMs for complex multimodal reasoning.

Speech and audio processing adds another modality to the multimodal landscape. Whisper (Robust Speech Recognition via Large-Scale Weak Supervision) is a large-scale multilingual speech recognition model trained on 680,000 hours of weakly supervised audio data. Whisper achieves state-of-the-art performance on ASR across dozens of languages and exhibits robustness to noise, accents, and varying recording conditions. Wav2Vec and HuBERT are self-supervised speech representation learning models that learn rich audio representations from unlabelled speech, enabling downstream tasks such as speech recognition, speaker identification, and emotion recognition. Speech-language models (e.g., SpeechGPT, AudioPaLM) integrate speech and text in a unified architecture, enabling end-to-end spoken language understanding and generation.

Multimodal retrieval is a fundamental application that enables searching across modalities. Text-to-image retrieval finds images that match a text query; image-to-text retrieval finds text that describes an image; and cross-modal search retrieves relevant items across multiple modalities. These tasks leverage the shared embedding space learned by models such as CLIP, using cosine similarity or Euclidean distance to rank candidates. Evaluation metrics include Recall@k (the proportion of queries where the correct item is in the top-k results), Mean Reciprocal Rank (MRR), and Normalised Discounted Cumulative Gain (NDCG). The effectiveness of cross-modal retrieval depends on the quality of the shared embedding space and the size of the retrieval corpus.

Multimodal generation encompasses tasks that produce content in one modality from another. Image captioning generates natural language descriptions of images; visual question answering (VQA) answers questions about images; text-to-image generation produces images from text descriptions (e.g., Stable Diffusion, DALL-E, Imagen); and video captioning generates descriptions of video content. These tasks combine understanding (encoding the input modality) with generation (decoding the output modality) in a sequence-to-sequence architecture. Evaluation metrics include BLEU, ROUGE, and BERTScore for generation quality, and Fréchet Inception Distance (FID) and Inception Score (IS) for image quality.

Multimodal RAG extends retrieval-augmented generation to multimodal documents. In a multimodal RAG system, the retriever indexes and retrieves relevant multimodal chunks (text, images, audio) from a knowledge base, and the generator (an LLM or multimodal model) uses the retrieved multimodal context to generate a response. Multimodal RAG enables question answering over documents that contain images, diagrams, charts, and audio, making it suitable for educational, medical, and multimedia applications.

The evaluation of multimodal models requires a comprehensive framework that measures multiple dimensions: cross-modal retrieval accuracy (Recall@k, MRR); generation quality (ROUGE, BERTScore, BLEU, FID); modality alignment (how well representations from different modalities align); robustness to noise and perturbations; and efficiency (inference time, memory consumption). Understanding these trade-offs is essential for choosing the right multimodal architecture for a given application.

\subsection{Experimental Interpretation}
Cross-modal retrieval, zero-shot image classification, image captioning and visual question answering are distinct tasks. CLIP, BLIP and instruction-following vision-language models should not be treated as interchangeable implementations of one generic evaluation protocol. Dataset licensing, annotation quality and image provenance matter for valid comparisons.

\section{Execution Procedure}

\subsection{Core: Three Required Tasks}
\begin{enumerate}[leftmargin=*,label=\textbf{\arabic*.}]

\item Use a documented image dataset and a pretrained CLIP encoder to construct a label-text baseline, keeping the actual test images out of prompt selection.

Select an image classification dataset with verified access conditions and a documented split, and identify the specific CLIP checkpoint and revision. Convert each allowed class name to a natural-language text candidate without using test labels to choose better wording. Explain how the image and text encoders produce comparable embeddings and how similarity scores determine the predicted class. Record image preprocessing, class vocabulary, device and precision; this is a zero-shot image--text matching experiment, not the training of a captioning model.

\item Compare two fixed text templates on an unchanged image subset, reporting the number of images, accuracy, macro-F1 and device/model revision.

Construct two prespecified text-prompt templates applied to precisely the same class names and test images. Keep preprocessing, checkpoint and set of candidate labels unchanged between runs. For each template calculate accuracy, macro-F1 and per-class support where gold labels are available, and identify class confusion patterns. If a subset is necessary for compute reasons, choose it by a stated rule before examining predictions and give its actual size. Do not choose templates using final-test performance and then describe the comparison as an unbiased final evaluation.

\item Analyse at least ten predictions and limitations, and distinguish CLIP image--text alignment from caption generation and visual question answering, which are separate optional experiments.

Inspect a diverse set of correct and incorrect predictions and distinguish visually ambiguous examples, prompt sensitivity, class-name mismatches and dataset biases. Check whether a similarity score is merely relative to the offered text candidates instead of interpreting it as a calibrated likelihood. Explain why CLIP classification results cannot substitute for evaluated image captions or visual question-answering responses. List captioning and VQA as separate extensions with their own models, reference annotations and metrics.

\end{enumerate}

\subsection{Reproducible Example and Its Limits}

The companion project contains executable code for this work. Run the real-data path with
\begin{center}\texttt{python practikum\_examples.py 11}\end{center}
CIFAR-10 test images and openai/clip-vit-base-patch32; this demonstrates zero-shot image classification, not captioning. For a technical-only check, add \texttt{-{}-smoke}; its synthetic inputs must not be used as scientific evaluation results. The program does not contain fabricated model scores. Record actual outputs, sample sizes, versions, timings and failures after running it.

\subsection{Extended Engineering Tasks (Optional)}
The following tasks provide a fuller engineering and research programme. They are optional rather than prerequisites for passing the core. The numerical scales and model choices below describe ambitious projects; they are not universal requirements. Documentation and reporting follow the shared protocol above.

\paragraph{Extended Task 1.} A multimodal dataset for evaluation is created or collected. The learner selects at least one open multimodal dataset: COCO Captions (image captioning), Flickr30k (image-text retrieval), VQAv2 (visual question answering), or a custom dataset with images and text. The dataset must include at least one thousand examples spanning two task types: cross-modal retrieval and generation. The dataset is split into training, validation, and test sets in a $60/20/20$ ratio. If using a ready-made dataset, the learner indicates its origin and licence in the datasheet. All data are saved in a structured format (JSONL for text, directories for images) with metadata (captions, questions, answers). References to the artefacts of the previous works are recorded in the methodological section of the report.

\paragraph{Extended Task 2.} A software module \texttt{multimodal\_utils.py} is developed, providing essential utilities for multimodal processing. The module must include: functions for loading and preprocessing images (resizing, normalisation, tensor conversion); functions for loading and preprocessing audio (resampling, spectrogram computation); functions for tokenising text and images (using CLIP tokenisers); functions for computing cross-modal embeddings (text and image); functions for calculating cross-modal similarity (cosine similarity); and a logging system for tracking multimodal experiments. The module is designed to be compatible with Hugging Face and PyTorch. The code is furnished with exhaustive documentation describing the interface of all public functions and configuration parameters.

\paragraph{Extended Task 3.} A comprehensive comparison of vision-language models is performed. The learner selects at least three vision-language models from the Hugging Face Hub: CLIP (openai/clip-vit-base-patch32), LLaVA (llava-hf/llava-1.5-7b-hf), and BLIP (Salesforce/blip-image-captioning-large). For each model, the learner evaluates performance on three tasks: (1) image-text retrieval (text-to-image and image-to-text) using Recall@k; (2) image captioning (generating captions for images) using ROUGE and BERTScore; (3) visual question answering (answering questions about images) using accuracy. The results are summarised in a table \texttt{vlm\_comparison.csv}. The learner analyses the trade-offs between model size, inference speed, and task performance, identifying the strengths and weaknesses of each model.

\paragraph{Extended Task 4.} Cross-modal retrieval with CLIP is implemented and evaluated in depth. The learner uses the CLIP model to compute embeddings for images and text. Text-to-image retrieval is implemented: given a text query, the learner retrieves the top-k images from a corpus of images (at least 500 images) based on cosine similarity. Image-to-text retrieval is implemented: given an image, the learner retrieves the top-k text descriptions. The learner evaluates performance using Recall@1, Recall@5, and Recall@10. The results are visualised as a bar chart and a confusion matrix. The learner analyses the influence of the corpus size, query diversity, and model variants (ViT-B/32, ViT-B/16, ViT-L/14) on retrieval performance.

\paragraph{Extended Task 5.} Image captioning with BLIP and LLaVA is implemented and compared. The learner uses BLIP (for image captioning) and LLaVA (for visual instruction following) to generate captions for a test set of at least 100 images. The captions are evaluated using ROUGE-L, BERTScore, and BLEU. The learner also performs a qualitative analysis: selected examples of captions from each model are presented with the source image, and the strengths and weaknesses of each model are discussed. The results are summarised in a table \texttt{captioning\_comparison.csv}. The learner analyses the differences between captioning and instruction-following approaches, concluding when each approach is most appropriate.

\paragraph{Extended Task 6.} Visual question answering (VQA) is implemented and evaluated. The learner uses LLaVA (or an alternative VQA model) to answer questions about images from the VQAv2 dataset. The learner selects at least 200 questions (100 validation, 100 test) and evaluates accuracy (exact match of the predicted answer with the ground truth). The learner also performs error analysis: common types of errors are identified (e.g., incorrect object recognition, reasoning errors, hallucination), and strategies for improving VQA performance are discussed. The results are visualised as a bar chart showing accuracy by question type (e.g., object recognition, counting, reasoning).

\paragraph{Extended Task 7.} Speech recognition with Whisper is implemented and evaluated. The learner uses the Whisper model (whisper-base, whisper-small, or whisper-medium) for automatic speech recognition on audio files from the corpus (if available) or from an open speech dataset. The learner transcribes at least 100 audio clips (varying in duration, accent, and background noise) and evaluates using Word Error Rate (WER) and Character Error Rate (CER). The learner analyses the influence of model size (base, small, medium, large) on transcription quality and inference speed. The results are visualised in a line chart showing WER as a function of model size and a bar chart comparing WER across different conditions (clean vs noisy, native vs non-native accent).

\paragraph{Extended Task 8.} Multimodal retrieval with speech and text is implemented. The learner extends the cross-modal retrieval pipeline to include speech: speech embeddings are computed using Whisper or Wav2Vec, and text embeddings are computed using CLIP or Sentence-BERT. The learner implements three retrieval tasks: (1) text-to-speech retrieval (given a text query, find the relevant audio clip); (2) speech-to-text retrieval (given an audio clip, find the relevant text description); (3) image-to-speech retrieval (given an image, find the relevant audio description). The learner evaluates performance using Recall@k on a corpus of audio-text-image triples. The results are summarised in a table \texttt{multimodal\_retrieval\_comparison.csv}. The learner analyses the challenges of cross-modal retrieval with speech, including the impact of speech variability (accent, noise) on performance.

\paragraph{Extended Task 9.} Multimodal RAG is implemented. The learner builds a multimodal RAG system: the knowledge base contains multimodal documents (text + images + audio) from the corpus. The retriever indexes text and image embeddings (CLIP) and audio embeddings (Whisper) in a single vector store (FAISS). The generator is an LLM (e.g., LLaMA-3-8B or Mistral-7B) that receives the retrieved multimodal context (text + image descriptions + audio transcripts) and generates a response. The learner evaluates the RAG system on a set of 50 queries that require information from multiple modalities. The results are compared with: (1) text-only RAG; (2) image-only retrieval; (3) LLM without retrieval. The evaluation uses manual annotation of answer quality (accuracy, completeness, relevance) on a 5-point Likert scale. The learner analyses the benefits and challenges of multimodal RAG.

\paragraph{Extended Task 10.} Multimodal generation: text-to-image with Stable Diffusion (or an analogous open model) is implemented. The learner uses Stable Diffusion (e.g., runwayml/stable-diffusion-v1-5) to generate images from text prompts. The learner generates at least 50 images from a set of diverse prompts (simple, complex, abstract, concrete). The generated images are evaluated qualitatively (subjective assessment by the learner or peers) and quantitatively using CLIP similarity between the prompt and the generated image. The learner also experiments with prompt engineering for text-to-image generation (from Work~\ref{work:10}), analysing how prompt formulation affects image quality and fidelity. The results are visualised as a gallery of generated images with corresponding prompts and similarity scores.

\paragraph{Extended Task 11.} Multimodal evaluation framework is developed and published. The framework must support: loading and preprocessing multimodal data (text, images, audio); computing multimodal embeddings; evaluating cross-modal retrieval (Recall@k, MRR); evaluating generation (ROUGE, BERTScore, BLEU, FID); visualising results (bar charts, confusion matrices, scatter plots); and generating reports (HTML, PDF). The learner uses the framework to benchmark at least five multimodal models (CLIP, LLaVA, BLIP, Whisper, Stable Diffusion) on a consistent evaluation set. The benchmarking results are published in a public repository with full documentation.

\paragraph{Extended Task 12.} An interactive web application for multimodal model demonstration is created using Gradio or Streamlit. The application must provide the following capabilities: uploading an image or audio file; entering a text prompt; selecting a multimodal model (CLIP, LLaVA, BLIP, Whisper, Stable Diffusion); displaying the model's output (retrieved images, generated captions, transcribed text, generated images); visualising cross-modal embeddings and similarity scores; and comparing multiple models side-by-side on the same input. The application is designed in such a way that it can be used by a non-specialist and is deployed locally with the possibility of subsequent deployment to Hugging Face Spaces.

\paragraph{Extended Task 13.} A suite of unit tests is developed based on the pytest framework, verifying the correctness and reproducibility of the key project components. The tests must check: the correctness of image preprocessing (tensor shapes, normalisation); the correctness of audio preprocessing (sampling rate, spectrogram shape); the correctness of cross-modal retrieval (ranking matches expected results on a small test set); the correctness of generation (outputs are non-empty and of the correct type); and the stability of benchmark results upon repeated runs. All dependencies are recorded in the \texttt{requirements.txt} file with an indication of the exact versions. A script is developed for the automatic execution of the full test suite and the generation of a pass/fail report.

\subsection{Further Research Directions (Optional)}
\textbf{First Additional Task} --- video understanding: extension of vision-language models to video (e.g., VideoCLIP, Flamingo for video); evaluation on video captioning and video question answering.

\textbf{Second Additional Task} --- multimodal prompting: development and evaluation of prompts for multimodal models (e.g., prompting LLaVA for specific image understanding tasks, prompting CLIP for zero-shot classification).

\textbf{Third Additional Task} --- multimodal data augmentation: generation of multimodal training data using Stable Diffusion (synthetic images for classification tasks); evaluation of data augmentation benefits.

\textbf{Fourth Additional Task} --- multimodal fairness and bias: analysis of biases in multimodal models (gender, racial, cultural biases in image-text datasets); development of bias mitigation strategies.

\textbf{Fifth Additional Task} --- multimodal evaluation metrics: development and evaluation of new metrics for multimodal generation that better capture cross-modal semantic alignment.

\textbf{Sixth Additional Task} --- multimodal for low-resource languages: application of multimodal models to low-resource languages (e.g., Tatar, Tajik) using cross-lingual transfer; evaluation of zero-shot performance.

\textbf{Seventh Additional Task} --- multimodal compression: application of compression techniques (from Work~\ref{work:17}) to multimodal models; analysis of the impact of compression on cross-modal performance.

\textbf{Eighth Additional Task} --- real-time multimodal systems: development of a real-time multimodal pipeline for streaming video or audio; evaluation of latency, throughput, and quality.

\section{Conclusion}
This work is not simply ``using CLIP and LLaVA''. It is the systematic integration of multiple modalities into unified intelligent systems that can perceive, understand, and generate content across the boundaries of language, vision, and sound. Multimodal NLP represents the frontier of artificial intelligence, where models move beyond text to interact with the world in a more task-dependent way --- seeing images, hearing speech, and reasoning across modalities.

In the course of performing the work, the learner learns to ask the key questions that determine the maturity of multimodal AI systems. Whether the model can align representations across modalities effectively (a question about cross-modal alignment). Whether the model can retrieve relevant content across modalities (a question about retrieval quality). Whether the model can generate coherent multimodal outputs (a question about generation quality). Whether the model is robust to variations in input quality and modality (a question about robustness). Whether the system is reproducible and can be extended to new modalities (a question about engineering discipline).

The answers to these questions constitute the difference between a `unimodal system` and a truly `multimodal intelligent system`. Upon completion of the work, the learner receives not simply a set of multimodal models but a methodological framework for designing, evaluating, and deploying multimodal AI systems --- from research prototypes to production applications, capable of perceiving and generating across the full spectrum of human experience.

\chapter{Large Language Models: Adaptation, RAG and Question Answering}
\label{work:12}

\section{Aim and Learning Outcomes}
This work connects evidence retrieval with question answering and model adaptation. Learners implement a retriever, assess whether retrieved passages support claims, separate retrieval errors from generation errors, and plan controlled adaptation experiments. Expensive LLM fine-tuning is treated as an optional extension when compute, access and data licences allow it.

The three tasks below constitute the required experimental core. Larger engineering studies and independent research questions are available as optional extensions.

\section{Theoretical Background}

\noindent\textit{Principal theoretical and methodological sources for this work:} \cite{lewis2020retrieval,ru_reference_61,arabov2026arabicrussiancorpus,arabov2026tatarstanrag,dettmers2024qlora}.\par

Parallel-corpus studies and geospatial question-answering systems provide domain-grounded examples for retrieval and evaluation, including Arabic--Russian scientific texts and Tatarstan toponyms \cite{arabov2026arabicrussiancorpus,arabov2026tatarstanrag}.

Modern large language models have fundamentally changed the approach to NLP: instead of narrowly specialised models, a unified generative architecture, adaptable to the task, is increasingly used. The choice of model, however, represents not only a question of quality but also a question of scientific ethics, engineering reliability, and legal security. \cite{lewis2020retrieval,dettmers2024qlora,hu2022lora}

The first perspective --- the classification of LLMs by accessibility --- divides models into two fundamentally different categories. Open models, such as Llama~3, Mistral, Qwen, DeepSeek, Phi-3, and ruT5, publish their weights openly; they can be run locally, modified, and fine-tuned versions can be published. This is the only class that ensures reproducibility, interpretability, and control over inference. Commercial API services, on the contrary, are closed, accessible only through cloud services, non-reproducible, opaque, and are not recommended as the foundation of a scientific or engineering project. The choice of an open model is not an ideological preference but a methodological necessity: if a result cannot be reproduced independently, it cannot be considered scientific.

The second perspective --- efficient fine-tuning --- is represented by the LoRA and QLoRA methods. Full fine-tuning was traditionally considered the gold standard for adapting language models, since it updates all parameters and potentially yields maximum quality. However, for models with a number of parameters of three billion and above, it requires hundreds of gigabytes of GPU memory and becomes technically and economically infeasible in an educational environment and in most industrial projects. It is precisely for this reason that parameter-efficient methods --- LoRA and QLoRA --- have replaced full fine-tuning. LoRA trains low-rank matrices instead of full weights, reducing memory consumption and training time by orders of magnitude. QLoRA combines four-bit quantisation with LoRA, making it possible to fine-tune seven-billion-parameter models on a GPU with 16--24~GB of memory. In doing so, 95--99 per cent of the quality is preserved with a radical reduction in resource requirements. The comparison of full fine-tuning and QLoRA becomes one of the central experimental lines of the work.

The third perspective --- the RAG (Retrieval-Augmented Generation) architecture --- divides the task into a retriever (search for relevant fragments) and a generator (formulation of an answer based on these fragments). RAG makes it possible to update knowledge without retraining, reduces the level of hallucinations, and makes the system interpretable by displaying sources. The quality of the system is determined primarily by the retriever, not the generator: even a powerful LLM will produce a false answer if it receives incorrect context. The retriever, in turn, can be implemented using sparse methods (BM25), dense methods (search by embedding similarity with models such as multilingual-e5), or their hybrid combinations. The choice of the retriever together with the chunking strategy (splitting documents into fragments) is often more significant for the final performance than the choice of the generator. Search quality is measured by the metrics recall@$k$ and Mean Reciprocal Rank, which reflect the system's ability to place relevant documents among the top results. The RAG pipeline is thus more correctly understood as an information retrieval task with a language model interface, rather than simply as a generative model with added context.

The fourth perspective --- ethical and methodological imperatives --- completes the theoretical framework. Reproducibility is the foundation of the scientific method, and the use of closed APIs violates it. Licence purity determines the very possibility of publishing a project: only models with permissive licences (Apache~2.0, MIT) can be legally distributed. The developer's responsibility includes the obligation to explicitly indicate sources, limitations, and the degree of trust in predictions. It is critically important to grasp: the best model is not the largest one, but the one that is open, adapted to the domain, honestly evaluated, and furnished with metadata.

\subsection{Experimental Interpretation}
Retrieval accuracy and generated-answer accuracy must be assessed separately. A retrieved passage can be irrelevant, and a fluent answer can be unsupported by its cited documents. Parameter-efficient fine-tuning and retrieval augmentation are complementary but not interchangeable; a working search baseline is not evidence that full LLM fine-tuning or RAG generation has been executed.

\section{Execution Procedure}

\subsection{Core: Three Required Tasks}
\begin{enumerate}[leftmargin=*,label=\textbf{\arabic*.}]

\item Construct a small retrieval baseline with a documented, permitted text corpus; preserve stable document IDs and a train/validation/test split of questions or sources.

Choose a document collection whose use and redistribution rules are known, assign stable source identifiers and preserve the original passage-to-document relationship. Define the question set independently of the final retrieval results and separate any tuning subset from the final evaluation items. Build a transparent lexical or embedding-based retrieval baseline, recording chunk size, overlap, index construction and retrieval depth. Where question-answer pairs or relevance judgements exist, describe how they were obtained rather than treating generated questions as independent human test data.

\item Compare retrieval-only evidence selection and retrieval-augmented answering using a documented model when resources permit. Assess citation support and answer correctness separately.

First evaluate retrieval by measuring relevant passage or document recall at an explicitly stated cutoff when verified relevance judgments are available. Then, if a permitted generative checkpoint and resources are available, supply the same retrieved evidence to the answer generator and preserve both generated answers and source references. Assess factual correctness and source support as different properties: a fluent answer may be unsupported even when retrieval is good. If generation cannot be run, fully document the retrieval and attribution experiment and keep generation in the optional programme, never filling in invented model responses.

\item Analyse unsupported claims and retrieval misses. If no generative model is available, complete a rigorous retrieval/attribution experiment and label generation as an optional extension, not as an executed result.

Trace a sample of failures to the relevant pipeline stage: source coverage, segmentation, retrieval ranking, lost context or unsupported generation. For each inspected question, link the claimed evidence to stable document identifiers and quote only as allowed by the source licence. Explain the effect of retrieval cutoff or chunking using measured changes if those variants were executed. Make clear that the mere presence of a citation marker does not establish that the cited passage entails the answer.

\end{enumerate}

\subsection{Reproducible Example and Its Limits}

The companion project contains executable code for this work. Run the real-data path with
\begin{center}\texttt{python practikum\_examples.py 12}\end{center}
Document retrieval on real 20 Newsgroups text; generating grounded answers requires an additional verified model run. For a technical-only check, add \texttt{-{}-smoke}; its synthetic inputs must not be used as scientific evaluation results. The program does not contain fabricated model scores. Record actual outputs, sample sizes, versions, timings and failures after running it.

\subsection{Extended Engineering Tasks (Optional)}
The following tasks provide a fuller engineering and research programme. They are optional rather than prerequisites for passing the core. The numerical scales and model choices below describe ambitious projects; they are not universal requirements. Documentation and reporting follow the shared protocol above.

\paragraph{Extended Task 1.} On the basis of the text corpus formed during Practical Work~\ref{work:1}, a final dataset is prepared for constructing a question-answering system and experiments with large language models. The corpus must contain at least ten thousand documents in JSONL format with the fields \texttt{id}, \texttt{title}, \texttt{text}, \texttt{source}, \texttt{url}, \texttt{date}, \texttt{language}. If the preparation of a corpus of the indicated volume is objectively difficult --- for example, due to limited access to news sources or the high cost of expert annotation, --- the learner has the right, in agreement with the instructor, to use a smaller volume corpus, explicitly stipulating this circumstance in the datasheet and discussing its influence on the robustness and generalisation ability of the models in the analytical report. All text materials are preserved in their original form, without lemmatization or stemming, since the pre-trained tokenizers of LLMs are optimised for the original word forms and subword units. The split into training, validation, and test sets is performed in a $70/15/15$ ratio with stratification by sources and time periods to ensure representativeness. The random seed value is fixed and documented. References to the artefacts of the previous works are recorded in the methodological section of the report.

\paragraph{Extended Task 2.} Independent but compatible data representations are created for all components of the system. The first representation --- semantically meaningful fragments (chunks) for the search module: the source documents are segmented into fragments of three hundred to five hundred tokens, preserving sentence boundaries, for which the \texttt{nltk} or \texttt{spaCy} libraries are used. Each fragment is enriched with metadata: source identifier, heading, date, URL, and language. The resulting collection is filtered: fragments containing fewer than fifty words are discarded, as well as fragments with a high level of noise --- advertising blocks, navigation elements, and boilerplate formulations. The second representation --- tokenised inputs for LLMs: the text is prepared using \texttt{AutoTokenizer.from\_pretrained} of the corresponding models (Llama, Mistral, Qwen, DeepSeek, Phi-3, ruT5) with the fixation of a maximum length specified by the parameters of each architecture, and subsequent padding/truncation. All prepared representations are saved with metadata in JSON format, including information about the chunking strategy, the tokenizer used, and a reference to the source document.

\paragraph{Extended Task 3.} A configurable software module \texttt{llm\_hub.py} is developed, providing a unified interface for running open large language models. The following architectures are supported: Llama-3-8B-Instruct, Mistral-7B-Instruct, Qwen1.5-7B-Chat, DeepSeek-7B-Base, Phi-3-mini-4k-instruct, and ruT5-base. Two operating modes are implemented: zero-shot inference using the \texttt{transformers} and \texttt{bitsandbytes} libraries with NF4 four-bit quantisation for models that do not fit in the available GPU memory; as well as CPU inference using the GGUF format and the \texttt{llama.cpp} framework for scenarios without a GPU. The module is designed without a hard dependency on closed APIs. The module code is furnished with exhaustive documentation describing the interface of all public functions and configuration parameters. Reproducibility is ensured through the fixation of seeds and library versions.

\paragraph{Extended Task 4.} Comparative fine-tuning: full fine-tuning, LoRA, and QLoRA. The learner conducts a systematic comparison of three strategies for adapting language models to the task of headline generation on the material of the news corpus. The first strategy --- Full Fine-Tuning --- presupposes updating all parameters of the model and is performed on architectures where this is technically feasible with the available GPU resources (minimally --- ruT5-base, optionally --- Phi-3-mini-4k-instruct if sufficient memory is available). The second strategy --- LoRA (Low-Rank Adaptation) --- fine-tunes low-rank adapters without quantising the base model; the configuration is fixed: rank $r=8$, scaling coefficient $\alpha=16$, target modules --- attention layers (\texttt{q\_proj}, \texttt{k\_proj}, \texttt{v\_proj}, \texttt{o\_proj}); LoRA is applied to all supported models, including Mistral-7B and Llama-3-8B, for which full fine-tuning is impossible in an educational environment. The third strategy --- QLoRA (Quantized LoRA) --- combines NF4 four-bit quantisation with the same adapter configuration ($r=8$, $\alpha=16$, attention layers: \texttt{q\_proj}, \texttt{k\_proj}, \texttt{v\_proj}, \texttt{o\_proj}), which makes it possible to fine-tune seven-billion-parameter models on a GPU with 16--24~GB of memory. The dataset is formed from ``article text $\rightarrow$ headline'' pairs extracted from the training portion of the corpus; the validation set ($15\%$) is used for early stopping, and the test set for the final comparison. The single evaluation metric for all three tracks is ROUGE-L. Learning curves are constructed (dependence of ROUGE-L and the loss function on the epoch number for the training and validation sets) separately for each track, as well as hyperparameter heat maps, if a grid search was conducted. The final results are summarised in a table containing, for each ``model--track'' combination: ROUGE-L on the test set, training time in minutes, peak GPU memory consumption in gigabytes, and the number of trainable parameters. Additionally, a ``quality (ROUGE-L) -- memory consumption'' plot is constructed, on which all combinations are marked as points for a visual assessment of Pareto-optimal solutions. All trained artefacts --- full checkpoints (for full fine-tuning), LoRA adapters, QLoRA adapters, as well as training logs --- are saved in separate directories with an indication of the configuration. The report formulates conclusions about the cases in which QLoRA falls short of LoRA in quality, the extent to which full fine-tuning is justified at small model sizes, and which of the methods is recommended under severe resource constraints.

\paragraph{Extended Task 5.} A Retrieval-Augmented Generation (RAG) system is built. The retriever is implemented on the basis of the FAISS library using the embedding model \texttt{intfloat/multilingual-e5-large} or, in the case of a priority on speed, \texttt{rubert-tiny2}. For each chunk, a dense vector is computed; all vectors are indexed using FAISS IndexFlatIP (search by cosine similarity via the scalar product of normalised vectors) or, for large corpus volumes, IndexIVFFlat for accelerating the search. The generator is an arbitrary open LLM loaded via \texttt{llm\_hub.py}. The prompt template is formulated as follows: ``Use only the information from the context. If the information is not there, answer: `I don't know'.'' The output of the system includes the generated answer and a list of relevant chunks with metadata (source, heading, date). The implementation is carried out using the \texttt{langchain} library or natively on \texttt{transformers} and \texttt{faiss} --- the choice is substantiated in the report.

\paragraph{Extended Task 6.} A comprehensive quality assessment of the obtained system is performed. The set of automatic metrics includes ROUGE-L and BERTScore for evaluating the relevance of the answers relative to reference ones. Factual reliability is assessed through the manual annotation of fifty answers on a five-point Likert scale. The proportion of hallucinations --- statements not confirmed by the retrieved chunks --- is computed. The RAG and LLM-only (without a retriever) modes are compared. The influence of the choice of retriever (dense E5 versus sparse BM25) and the generator size (Llama~3 versus Phi-3) on the final metrics is investigated. For the retriever, recall@$k$ and MRR are additionally computed. All metrics are summarised in a single table \texttt{rag\_metrics.csv}. Box plots across folds (if cross-validation was applied) and plots of the dependence of the metrics on the number of relevant chunks are constructed.

\paragraph{Extended Task 7.} An in-depth error analysis and interpretation of predictions are performed. Hallucination matrices are constructed with a breakdown by type: factual errors (incorrect statement), date distortions, errors in named entities, incorrect source attribution. Retriever failures are analysed: cases where incorrect or irrelevant context leads to a false answer, and the proportion of errors caused by the retriever is computed. UMAP projections of chunk embeddings are visualised, making it possible to assess the clustering of thematically close fragments. For the generator, where technically available, heat maps of attention over context are constructed, showing which chunks the model relied on when generating each token. To assess the model's confidence, reliability diagrams are constructed, and the Expected Calibration Error (ECE) is computed. All visualisations are generated directly in the course of code execution and are included in the analytical report with extensive commentary.

\paragraph{Extended Task 8.} An investigation of the robustness of the RAG system to input distortions typical of real operation is performed. The learner uses the \texttt{text\_corruptor.py} script, developed in Practical Work~\ref{work:8}, to create distorted versions of the test set. Noise levels of $10\%$, $30\%$, and $50\%$ are applied (the proportion of words with typos, letter transpositions, morphological errors). On the distorted data, the same metrics are computed as on the clean data (ROUGE-L, BERTScore, proportion of hallucinations), and the relative drop in quality is measured. Additionally, it is analysed which component of the system --- the retriever or the generator --- is more vulnerable: for this, recall@$k$ of the retriever on distorted queries is computed and compared with recall on clean queries. The results are presented in the form of a summary table and a plot of the dependence of the metrics on the proportion of distorted words. The learner formulates conclusions about the practical applicability of the system under conditions of noisy user input and, where necessary, proposes strategies for increasing robustness (e.g., preliminary grammatical correction of the query before feeding it into RAG).

\paragraph{Extended Task 9.} A formal verification of the correctness of the data splitting and the absence of information leakage is performed, and a unified module \texttt{RAGSystem} is developed. The learner documents the splitting procedure with the fixation of the random seed, and programmatically verifies that the chunks generated from the test documents were not used in training the retriever and were not indexed in FAISS until the completion of training. It is controlled that the evaluation dataset for measuring hallucinations does not intersect with the data on which the fine-tuning and comparison of models were performed. The verification protocol is included in the analytical report. In parallel, the class \texttt{RAGSystem} is developed with the public methods \texttt{.fit()} (indexing of chunks and configuration of the retriever), \texttt{.predict(query)} (generation of an answer with the return of sources), and \texttt{.explain()} (return of relevant chunks and attention weights over context). The class is compatible with the \texttt{transformers} and \texttt{langchain} libraries, supports the serialisation of state (FAISS index, LoRA adapters) via \texttt{joblib} and \texttt{torch.save}, and records seeds for reproducibility checks, subject to system-dependent variation.

\paragraph{Extended Task 10.} An interactive web application is created using Gradio, providing the user with the following capabilities: entry of an arbitrary query in natural language; choice of the generator model and operating mode (RAG with the display of sources or LLM-only without context); display of the generated answer, a list of relevant chunks with metadata, and highlighting of the most significant fragments; visualisation of the model's confidence (confidence scores); automatic generation of an HTML report containing the query, answer, sources, and metrics. The application is designed in such a way that it can be used by a specialist without programming skills and is deployed locally with the possibility of subsequent deployment to Hugging Face Spaces.

\paragraph{Extended Task 11.} A suite of unit tests is developed based on the pytest framework, verifying the correctness and reproducibility of the key project components. The tests must check: the fixation of seeds and library  (\texttt{transformers}, \texttt{bitsandbytes}, \texttt{faiss-cpu}); the correctness of loading open models by identifiers; the stability of the metrics (ROUGE-L, recall@$k$) upon repeated runs; the absence of information leakage between the training, validation, and test sets; the correctness of the operation of the \texttt{text\_corruptor.py} script as applied to queries (the correspondence of the proportion of distorted words to the specified level). All dependencies are recorded in the \texttt{requirements.txt} file with an indication of the exact versions. A script is developed for the automatic execution of the full test suite and the generation of a pass/fail report. The presence of a successfully passing test suite is regarded as an integral component of the software artefact.

\subsection{Further Research Directions (Optional)}
\textbf{First Additional Task} --- comparison of open LLMs and a commercial API: limited testing of factuality with a mandatory caveat about the non-reproducibility of the results.

\textbf{Second Additional Task} --- QLoRA on a consumer GPU: fine-tuning Phi-3-mini on an RTX 3060 (12~GB) with detailed profiling of memory consumption and time.

\textbf{Third Additional Task} --- CPU inference: running Qwen-0.5B via GGUF on a laptop without a GPU and comparing latency with GPU inference.

\textbf{Fourth Additional Task} --- multilingual RAG: support for the Russian and Tatar languages (based on the multilingual corpus from Practical Work~\ref{work:1}) with a separate quality assessment for each language.

\textbf{Fifth Additional Task} --- automatic factuality assessment: fine-tuning ruBERT on the NLI task for automatic verification of RAG answers and comparison with manual annotation.

\textbf{Sixth Additional Task} --- the influence of chunk length: comparison of fragments of 200, 500, and 1000 tokens by recall@$k$ of the retriever and ROUGE-L of the generator.

\textbf{Seventh Additional Task} --- HyDE (Hypothetical Document Embeddings): generation of a hypothetical answer to improve search and assessment of the gain in recall@$k$.

\textbf{Eighth Additional Task} --- energy consumption assessment: comparison of the CO\textsubscript{2} footprint of Llama~3 and Phi-3-mini via CodeCarbon when executing one hundred queries.

\section{Conclusion}
This work is not simply ``connecting to an LLM via an API''. It is a deep immersion into the architectural, ethical, and engineering foundations of modern large language models and retrieval-augmented generation systems, where every decision --- from the choice of licence to the chunk length and LoRA rank --- carries a methodological load.

In the course of performing the work, the learner learns to ask the key questions that determine the maturity of modern AI systems. Whether it is possible to verify where the model took this fact from --- or whether it invented it (a question about transparency and hallucinations). Whether this result can be reproduced tomorrow --- or whether it will disappear when the API is disconnected (a question about scientific integrity). Whether it is permissible to use the given model in a public project from the point of view of a licence (a question about legal responsibility). To what extent the system is robust to typos and noise inevitable in real operation (a question about robustness). How to reduce hallucinations without sacrificing expressiveness and speed (a question about the balance of faithfulness and generation quality).

The answers to these questions constitute the difference between a ``black box'' and a responsible intelligent system capable of working not only intelligently but also honestly, transparently, robustly, and reproducibly. Upon completion of the work, the learner receives not simply a question-answering system but an architectural compass for designing, evaluating, and deploying any LLM solutions --- from corporate chatbots to news analysis systems, capable of grounding generation in facts, documenting sources, and withstanding the test of time and noise.

\chapter{Domain-Specific Evaluation and Benchmarks for LLMs}
\label{work:13}

\section{Aim and Learning Outcomes}
Students construct a small domain-specific benchmark with a transparent rubric, reference data provenance and an executable scoring pipeline. The work addresses annotation reliability, data contamination, uncertainty and error taxonomy. Its purpose is to develop defensible evaluation methodology rather than publish unsupported leaderboard claims about opaque language models.

The three tasks below constitute the required experimental core. Larger engineering studies and independent research questions are available as optional extensions.

\section{Theoretical Background}

\noindent\textit{Principal theoretical and methodological sources for this work:} \cite{chang2024critical,gao2023framework}.\par

General benchmarks measure the broad cognitive abilities of LLMs; however, they do not reflect their suitability for practical application in professional or educational contexts. Specialised benchmarks fill this gap by focusing on domain competence rather than general erudition. \cite{chang2024critical,gao2023framework}

The need for domain evaluations is conditioned by three key factors. Firstly, formal accuracy: compliance with orthographic, mathematical, or legal norms, where an error in a single digit or an incorrect reference to an article of law renders the answer completely invalid, even if the model's general reasoning appears plausible. Secondly, structural correspondence: many professional answers must contain strictly defined elements --- for example, the chain ``thesis $\rightarrow$ argument $\rightarrow$ example'' in an examination essay, --- which is not captured by general metrics oriented towards semantic similarity. Thirdly, contextual adequacy: models pre-trained on international corpora often ignore local realities, be it the specifics of Russian legislation, the content of the school curriculum, or the morphology of the Tatar language, which makes their conclusions formally correct but practically useless.

The architecture of a specialised benchmark is built around several interconnected components. Its foundation is a corpus of tasks furnished with a multi-level taxonomy (subject $\rightarrow$ subsection $\rightarrow$ task type $\rightarrow$ difficulty), which makes it possible to analyse results at any granularity. Each task is accompanied by a reference answer; for tasks that permit variability, alternative formulations or ranges of permissible values (e.g., $\pm 1$ for a numerical answer in mathematics) are recorded. The evaluation criteria are formalised either as machine-readable rules or as annotated templates describing the mandatory and optional elements of the answer. These criteria govern the evaluation pipeline --- an automated system that compares the model's answer with the reference and computes a set of metrics adapted to the task type: from simple matching for test questions to coverage, factual reliability, and structural correspondence for extended answers.

The construction of such a benchmark gives rise to a methodological problem that does not arise in general evaluations: the evaluation of extended, freely formulated answers. A short answer can be marked as correct or incorrect by direct comparison with the reference; an essay or a diagnostic justification cannot. Therefore, the present work distinguishes two evaluation modes. Short answers are evaluated through normalised exact match or numerical closeness. Extended answers are evaluated through a combination of three methods: rule-based checks that verify the presence of mandatory structural elements; model-based comparison with the reference via BERTScore; and a fine-tuned classifier that predicts the fulfilment of individual criteria. Each method has its own strengths and modes of failure, and they are designed for combined rather than isolated use. The learner's task includes not only the implementation of these evaluators but also the analysis of where they agree, where they diverge, and what these divergences say about the limits of automatic evaluation.

The spectrum of subject domains to which the given approach is applicable is extremely broad. In education, these are OGE and USE tasks with clear evaluation criteria developed by FIPI. In medicine --- questions on clinical protocols and diagnostics, where the cost of an error is especially high. In law --- tasks on knowledge of the Civil Code with the mandatory requirement to cite specific articles. For low-resource languages --- linguistic tasks in Tatar or Bashkir, making it possible to assess the extent to which the model accounts for morphological specificity. It should be emphasised that the OGE in the present work serves exclusively as a methodological example demonstrating the principles of constructing a benchmark in an area with formal rules. The approach being developed is fully transferable to any area that possesses clear evaluation criteria.

\subsection{Experimental Interpretation}
Benchmarks must state the task definition, scoring procedure, annotation provenance and limitations. An apparently clean held-out set does not establish the absence of pretraining contamination for a model whose training corpus is unknown. A deterministic classification benchmark should not be described as a complete generative-LLM evaluation.

\section{Execution Procedure}

\subsection{Core: Three Required Tasks}
\begin{enumerate}[leftmargin=*,label=\textbf{\arabic*.}]

\item Define a small domain-specific task set and an explicit annotation rubric, including independently checked reference answers or labels and provenance information.

Define a coherent subject area and tasks whose outputs can be checked independently, along with sampling and annotation criteria. Preserve item identifiers, linguistic coverage, reference labels or answers, authorisation to use the source data, and a rubric for borderline cases. Separate any examples used to design the rubric or prompts from a held-out evaluation set. When several annotators are available, describe disagreement handling; when they are not, label the limited validation appropriately.

\item Implement a deterministic baseline and an evaluation script that reports per-class or per-task metrics with sample counts and separated development/test data.

Implement a baseline that can actually be rerun, for example a lexical heuristic, retrieval strategy or fixed model checkpoint, and specify the metric for each task. Build a scoring script that handles missing, abstained and malformed predictions explicitly rather than removing them from denominators. Include sample counts and per-group or per-class results; if confidence intervals are computed, state the resampling or statistical assumptions. Freeze the scoring rule before examining the final test outcomes.

\item Conduct an error audit and identify possible contamination paths. Explain that the absence of training-data overlap cannot be proven for an opaque pretrained model without access to its pretraining corpus.

Inspect errors from several topics or subgroups and distinguish annotation ambiguity from model failure. Review whether publicly accessible examples, repeated documents or shared source material could cause contamination and document any duplicate checks actually performed. For a proprietary or incompletely documented pretrained model, treat possible pretraining exposure as a limitation rather than claiming to prove non-overlap. Conclude which capabilities the benchmark measures, and which capabilities it does not.

\end{enumerate}

\subsection{Reproducible Example and Its Limits}

The companion project contains executable code for this work. Run the real-data path with
\begin{center}\texttt{python practikum\_examples.py 13}\end{center}
20 Newsgroups classification benchmark; a dedicated LLM benchmark requires a separate scoring protocol. For a technical-only check, add \texttt{-{}-smoke}; its synthetic inputs must not be used as scientific evaluation results. The program does not contain fabricated model scores. Record actual outputs, sample sizes, versions, timings and failures after running it.

\subsection{Extended Engineering Tasks (Optional)}
The following tasks provide a fuller engineering and research programme. They are optional rather than prerequisites for passing the core. The numerical scales and model choices below describe ambitious projects; they are not universal requirements. Documentation and reporting follow the shared protocol above.

\paragraph{Extended Task 1.} On the basis of open sources, a corpus of domain-specific tasks with reference answers is constructed. For the subject area ``preparation for the Basic State Examination (OGE)'', the sources are publicly available demonstration versions and the open bank of tasks of the Federal Institute of Pedagogical Measurements (FIPI); for jurisprudence --- the Civil Code of the Russian Federation with commentaries; for medicine --- clinical protocols and certification questions; for low-resource languages --- corresponding textbooks and national corpora. The use of ready-made open datasets relevant to the chosen subject area is permitted, provided that their origin and licence are indicated in the datasheet; priority in assessment is given to tasks performed on original collected material. The minimum volume of the corpus is three hundred tasks. If, for the chosen domain, the collection of a corpus of the indicated volume is objectively difficult --- for example, due to the limited availability of expertly annotated materials for a low-resource language, --- the learner has the right, in agreement with the instructor, to use a smaller corpus, explicitly stipulating this circumstance in the datasheet and discussing its influence on the statistical significance of the evaluations in the analytical report. Each task is saved in JSONL format and contains the fields: \texttt{id} (unique identifier), \texttt{subject} (subject), \texttt{task\_type} (task type: short answer, multiple choice, extended answer), \texttt{difficulty} (difficulty level: basic, advanced, high), \texttt{prompt} (task text), \texttt{reference\_answer} (reference answer), and \texttt{criteria} (evaluation criteria). For tasks with an extended answer, the criteria are specified as a list of mandatory and optional elements; the split into training, validation, and test sets is performed in a $70/15/15$ ratio with stratification by subjects and difficulty levels; the random seed value is fixed and documented to ensure full reproducibility; and references to the specific artefacts of the previous works (the path to the corpus file from Work~\ref{work:1}, the tokenisation strategies from Work~\ref{work:1}, the labelled datasets from Works~\ref{work:4} and Work~\ref{work:5}, the pre-trained embeddings from Work~\ref{work:2}, the fine-tuned transformers from Work~\ref{work:6}, the transfer learning strategies from Work~\ref{work:7}, the multi-task architecture from Work~\ref{work:8}, the Hugging Face platform approach from Work~\ref{work:9}, the LLM and RAG methodology from Work~\ref{work:12}) are recorded in the methodological section of the report.

\paragraph{Extended Task 2.} A hierarchical taxonomy of tasks and several formats of input representations are created for compatibility with various models. The taxonomy is constructed according to the scheme: subject $\rightarrow$ subsection $\rightarrow$ task type $\rightarrow$ difficulty level. For tasks with an extended answer, key elements are additionally annotated --- mandatory components such as definition, example, and conclusion, --- in the form of a structured list within the \texttt{criteria} field. All metadata are saved in the file \texttt{taxonomy.json}. For compatibility with various LLMs, prompts are prepared in several formats: for inference via \texttt{AutoTokenizer.from\_pretrained} of the corresponding generative models (Llama, Mistral, Qwen, DeepSeek, Phi-3, ruT5) with an instructional prefix; for the few-shot mode --- with the addition of two or three examples of correct answers; for the RAG mode --- with the template ``Context: \{retrieved\_chunks\} Question: \{prompt\} Answer:''. For short answers, a normalised representation is additionally prepared: lowercasing, removal of terminal punctuation marks, lemmatization for the Russian language. All representations are saved with metadata in JSON format.

\paragraph{Extended Task 3.} A software module \texttt{domain\_evaluator.py} is developed, providing a unified interface for the automatic evaluation of answers. For tasks with a short answer, the module performs normalisation of the model's answer and the reference (lowercasing, removal of terminal punctuation marks, lemmatization) with subsequent exact string comparison or, for numerical answers, a check of numerical closeness with a tolerance of $\pm 1$. For tasks with an extended answer, a three-component evaluation strategy is implemented. The first component --- rule-based: using regular expressions, named entity recognition (via the \texttt{natasha} or \texttt{stanza} library), and detection of the presence/absence of key phrases, the presence of each mandatory element from the list of criteria is verified. The second component --- model-based: BERTScore is computed relative to the reference answer using the pre-trained model \texttt{DeepPavlov/rubert-base-cased} or its analogue for the corresponding language. The third component --- classifier-based: a classifier fine-tuned on annotated ``answer -- criterion'' pairs (based on RuBERT from Practical Work~\ref{work:4} or an analogue) is used, which, for each criterion, predicts a binary assessment ``fulfilled / not fulfilled''. The final score for the extended answer is formed as a weighted sum of points across all criteria; the weights of the criteria are extracted from the \texttt{criteria} field of the task or are set equal. The output of the module is a structured JSON object containing the total score, a list of assessments for each criterion, a list of matched key elements, BERTScore, and a list of detected hallucinations; the module code is furnished with exhaustive documentation describing the interface of all public functions and configuration parameters.

\paragraph{Extended Task 4.} Systematic testing of open LLMs on the formed benchmark is carried out with full visualisation monitoring. The tested architectures are: Llama-3-8B-Instruct, Mistral-7B-Instruct, Qwen1.5-7B-Chat, DeepSeek-7B-Base, and ruT5-base. For each model, three modes are investigated: zero-shot (direct query with an instructional prefix), few-shot (addition of two or three examples of correct answers before the target task), and RAG (access to a domain knowledge base --- chunks from educational materials, built according to the methodology of Practical Work~\ref{work:9}). The prohibition on the use of closed APIs in the main pipeline is strictly observed. For each model and each mode, the following metrics are computed: accuracy (for short answers and multiple-choice tasks), BERTScore and coverage --- the proportion of fulfilled criteria (for extended answers), the proportion of hallucinations, as well as the mean inference time. Curves of the dependence of the metrics on the difficulty level of the tasks, heat maps of ``model $\times$ mode $\times$ subject'', as well as bar charts for visual comparison of the modes are constructed. All configurations, metrics, and visualisations are saved.

\paragraph{Extended Task 5.} A comparison of the quality of the models' answers is performed by modes and task types. The results are summarised in a single table \texttt{benchmark\_results.csv}, the rows of which correspond to the ``model $\times$ mode $\times$ subject'' combinations, and the columns to the metrics (accuracy, BERTScore, coverage, proportion of hallucinations). Additionally, box plots of the metrics across subsamples, plots of criterion coverage for extended answers with a breakdown by criterion types (definition, argumentation, example, conclusion), and bar charts comparing the modes are constructed. All visualisations are furnished with analytical commentary.

\paragraph{Extended Task 6.} An in-depth error analysis and interpretation of predictions are performed. Confusion matrices are constructed by task types and subjects. Inventories of hallucinations are compiled with a breakdown by sources: factual errors (incorrect statement), errors in formulae and numerical values, errors in terminology, incorrect source attribution. For answers that scored highly but contain hallucinations, a qualitative analysis is performed with the documentation of examples. UMAP projections of the vector representations of the models' answers relative to the references are constructed, making it possible to visually assess which models and in which modes give answers closest to the reference ones. To visualise the fulfilment of criteria, heat maps of ``model $\times$ criterion'' are constructed, on which the mean score for each criterion is encoded in colour. All visualisations are generated directly in the course of code execution and are included in the analytical report with extensive commentary.

\paragraph{Extended Task 7.} The gap between the results on general and specialised benchmarks is investigated. For each tested model for which this is technically accessible, the learner compares the indicators obtained on the domain benchmark with the published MMLU results of the same model. Models are identified that demonstrate high results on general tests but fall significantly short on narrow-domain tasks. It is analysed exactly which types of tasks (short answer, multiple choice, extended answer) make the greatest contribution to this gap. Cases are separately substantiated in which RAG or few-shot give a significant gain in quality in a narrow domain compared to zero-shot. The results are presented in the form of a summary table and a scatter plot, where MMLU (or its surrogate) is plotted on the abscissa and accuracy on the domain benchmark on the ordinate, with colour coding by models.

\paragraph{Extended Task 8.} The robustness of the benchmark and models to input distortions is investigated. The learner develops a script \texttt{prompt\_corruptor.py} (by analogy with \texttt{text\_corruptor.py} from Practical Work~\ref{work:8}), which automatically introduces controlled distortions into the formulations of the tasks of the test set. The following types of distortions are implemented: minor reformulations of the question (synonymous replacement of keywords while preserving meaning), introduction of typos (transposition of adjacent letters, omission of characters --- $10\%$, $30\%$, $50\%$ of words), morphological errors (incorrect endings, replacement of \emph{y} with \emph{i} and vice versa for the Russian language, replacement of diacritical letters with letters without diacritics for Tatar and other languages). For each type and level of distortion, a separate version of the test set is created; the training and validation sets are not distorted. On the distorted versions, the same metrics are computed as on the clean data, and the relative drop in quality is measured. The results are presented in the form of a summary table and a plot of the dependence of the key metrics on the proportion of distorted words. The learner formulates conclusions about which models are most robust to reformulations and typos --- a critically important property for real operation, where user queries are rarely perfectly literate.

\paragraph{Extended Task 9.} A formal verification of the correctness of the data splitting and the absence of information leakage is performed, and a unified module \texttt{DomainBenchmark} is developed. The learner documents the splitting procedure with the fixation of the random seed, and programmatically verifies that the tasks from the test set were not used in training the model-based evaluator and that the examples from the few-shot prompts do not intersect with the test set. It is controlled that the distorted versions of the tasks (from Task~8) were not used in training and validation. The verification protocol is included in the analytical report. In parallel, the class \texttt{DomainBenchmark} is developed with the public methods \texttt{.fit()} (loading the taxonomy and configuring the evaluator), \texttt{.evaluate(model, mode)} (running the model on the benchmark with the return of a structured result), and \texttt{.explain(result)} (return of a per-criterion breakdown and examples of errors). The class is compatible with the \texttt{transformers} and \texttt{lm\_eval} (EleutherAI LM Evaluation Harness) libraries, supports the serialisation of state via \texttt{joblib} and \texttt{torch.save}, and records seeds for reproducibility checks, subject to system-dependent variation.

\paragraph{Extended Task 10.} An interactive web application is created using Gradio. The interface provides the user with the following capabilities: choice of subject domain (OGE, law, medicine), task type, model, and mode (zero-shot, few-shot, RAG); display of the generated answer, the reference answer, and a per-criterion evaluation (for extended answers --- highlighting of fulfilled and unfulfilled criteria); side-by-side comparison of the answers of several models; display of the list of sources for the RAG mode; and automatic generation of an HTML report containing the task, the models' answers, metrics, and criteria. The application is designed in such a way that it can be used by a domain expert without programming skills and is deployed locally with the possibility of subsequent deployment to Hugging Face Spaces.

\paragraph{Extended Task 11.} A suite of unit tests is developed based on the pytest framework, verifying the correctness and reproducibility of the key project components. The tests must check: the fixation of seeds and library  (\texttt{transformers}, \texttt{lm\_eval}, \texttt{bert\_score}); the correctness of the evaluation of short answers (exact match after normalisation); the correctness of the evaluation of extended answers (the rule-based component detects the presence of key phrases, BERTScore correlates with the expert assessment on the validation set); the stability of the metrics upon repeated runs; the absence of information leakage between the training and test sets; the correctness of the operation of the \texttt{prompt\_corruptor.py} script (the correspondence of the proportion of distorted words to the specified level); and the complete absence of calls to closed APIs in the main pipeline. All dependencies are recorded in the \texttt{requirements.txt} file with an indication of the exact versions. A script is developed for the automatic execution of the full test suite and the generation of a pass/fail report. The presence of a successfully passing test suite is regarded as an integral component of the software artefact.

\paragraph{Extended Task 12.} Datasheet and Model Cards are formed, and the taxonomy is documented. The Datasheet contains: the name and version of the benchmark, the subject domain, the volume (number of tasks, distribution by types and difficulty), the sources, a description of the taxonomy, the collection and normalisation procedure, known limitations and systematic biases (e.g., the predominance of tasks of a certain type), and the licence. For the model-based evaluator (the fine-tuned criteria classifier), a separate Model Card is completed with a description of the architecture, the training set, the hyperparameters, and the accuracy metrics on the validation set. Additionally, the file \texttt{domain\_taxonomy.md} is compiled, describing the hierarchical taxonomy and the evaluation criteria in human-readable form, and the file \texttt{metrics\_spec.json}, containing the formal specification of all metrics and the rules for computing them. Recommendations for adapting the benchmark to a new subject domain are formulated. All cards and documents are included in the final analytical report and, in the case of publication, are placed in the repository together with the artefacts.

\subsection{Further Research Directions (Optional)}
\textbf{First Additional Task} --- multi-domain benchmark: combining tasks from education, law, and medicine into a single set with a common taxonomy and comparing the ranking of models by domains.

\textbf{Second Additional Task} --- task generation: fine-tuning a model for creating new tasks on a given topic (e.g., ``generate a task on the conjugation of verbs in the Tatar language'') with subsequent expert validation of quality.

\textbf{Third Additional Task} --- adaptation for low-resource languages: implementation of a benchmark in Tatar or Bashkir with support for morphological evaluation.

\textbf{Fourth Additional Task} --- comparison of RAG and fine-tuning: what is more effective in a narrow domain --- providing context via RAG or fine-tuning a model on domain data?

\textbf{Fifth Additional Task} --- automatic annotation of criteria: using an LLM to generate rubrics (evaluation criteria) from the task description and comparing them with expertly compiled ones.

\textbf{Sixth Additional Task} --- assessment of robustness to reformulations: systematic measurement of how quality changes upon synonymous replacement of the formulations of the question.

\textbf{Seventh Additional Task} --- integration with an LMS: exporting the benchmark to the Moodle or Stepik format for use in a real educational process.

\textbf{Eighth Additional Task} --- ethical audit: analysis of risks (bias, safety) when using LLMs in educational assessment.

\section{Conclusion}
This work is not simply ``testing LLMs on school problems''. It is the development of a methodology for evaluating artificial intelligence in professional contexts, where an error in a single detail can have real consequences --- from an incorrect diagnosis to legal incompetence.

In the course of performing the work, the learner learns to ask the key questions that determine the maturity of modern AI systems. Whether the model can not simply answer but answer according to the rules (a question about formal competence). If the model is generally intelligent, why does it fail an examination in a specific subject (a question about the gap between general and domain competence). Whether it is possible to evaluate reasoning automatically --- in the way a teacher would do it (a question about the limits of automatic evaluation). Whether reproducibility is guaranteed --- from the first task to the final report (a question about scientific integrity). To what extent the evaluations are robust to minor reformulations and typos (a question about the robustness of the benchmark).

The answers to these questions constitute the difference between a general language model and a professionally applicable intelligent system. Upon completion of the work, the learner receives not simply a set of evaluations but a methodological pipeline for designing, evaluating, and publishing benchmarks in any subject domain --- from school education to clinical medicine, capable of serving as a foundation for both scientific research and industrial solutions, where accuracy, structure, and context are not optional features but mandatory requirements.

\chapter{Preference-Based Alignment: RLHF, PPO and DPO}
\label{work:14}

\section{Aim and Learning Outcomes}
The purpose is to understand preference-based language-model alignment as a family of distinct methods. Students inspect preference records, verify the origin and consistency of chosen/rejected responses, and distinguish reward modelling, PPO-based RLHF and direct preference optimisation. Any policy-training claim must be supported by a real execution and a documented reference baseline.

The three tasks below constitute the required experimental core. Larger engineering studies and independent research questions are available as optional extensions.

\section{Theoretical Background}

\noindent\textit{Principal theoretical and methodological sources for this work:} \cite{ru_reference_64,ouyang2022training,rafailov2023direct}.\par

Reinforcement Learning from Human Feedback emerged as an answer to the fundamental limitation of Supervised Fine-Tuning (SFT): in generation tasks, there is no single ``correct'' answer. Instead, quality is determined by a multidimensional, subjective, and contextual assessment. RLHF overcomes this uncertainty by replacing pointwise annotation with relative preferences, which better reflect the complexity of human judgement. \cite{ouyang2022training,rafailov2023direct}

The architecture of the RLHF pipeline includes three sequential stages. The first stage --- Supervised Fine-Tuning (SFT) --- represents the basic adaptation of a pre-trained language model to the format of the target task: question-answering dialogue, summarisation, or generation. The second stage --- Reward Modeling --- consists of training a reward function $r_\theta(x, y)$ that predicts how well an answer $y$ to a prompt $x$ corresponds to human preferences. The Bradley--Terry model is standardly applied: the probability that answer $y_1$ is preferable to $y_2$ given $x$ is proportional to the sigmoid of the difference of their rewards. The third stage --- Reinforcement Learning --- performs further training of the SFT model $\pi_\phi$ using the PPO algorithm, where the reward is replaced by $r_\theta$, and regularisation is provided through the KL divergence with respect to a reference model $\pi_{\text{ref}}$. The final RL loss function includes two terms: the expectation of the reward, which encourages the generation of highly rated answers, and a penalty KL term, which keeps the policy close to the distribution of the reference model.

The penalty KL term is not a secondary technical detail; it is critically necessary for preventing a phenomenon known as distributional shift. As the policy $\pi_\phi$ is updated, it begins to generate texts that may lie outside the distribution on which the reward model was trained. In such regions, the estimates of the reward model become unreliable, and the policy is capable of learning to exploit spurious patterns that yield a high reward with low real quality. The KL term keeps the policy close to the reference model, guaranteeing that it does not drift too far into zones where the reward signal loses meaning.

Modern alternatives to PPO substantially simplify the pipeline. Direct Preference Optimization (DPO), proposed by Rafailov et al.\ in 2023, shows that preference optimisation can be reduced to a classification task without explicit RL. DPO reparameterises the Bradley--Terry model directly in terms of the policy, eliminating the need for a separate reward model and the associated instability of RL training. This makes the method considerably simpler to implement, although it means that the reward signal cannot be inspected or audited independently of the policy. Identity Preference Optimization (IPO) improves stability under weak or noisy preferences, and the Cross-Entropy Method (CEM) with beam search and retraining serves as a baseline without RL.

The critical challenges of RLHF include four interconnected problems. Firstly, annotator bias: cultural, linguistic, and professional differences distort preferences. Secondly, overoptimisation: the model learns to ``deceive'' the reward model, generating formally correct but vacuous or sycophantic answers. Thirdly, the conflict between alignment and competence: an excessive emphasis on safety can suppress useful but debatable statements. Fourthly, the absence of ground truth: the quality of RLHF is assessed only through proxy metrics, which requires a multi-method approach to evaluation.

\subsection{Experimental Interpretation}
RLHF commonly combines preference modelling with reinforcement learning (such as PPO), whereas direct preference optimisation (DPO) uses a different preference-based objective without a separately trained reward model in its standard formulation \cite{ouyang2022training,rafailov2023direct}. Automatically collected or model-generated preferences must not be described as verified human judgements. A schema validation example is not equivalent to successful DPO training.

\section{Execution Procedure}

\subsection{Core: Three Required Tasks}
\begin{enumerate}[leftmargin=*,label=\textbf{\arabic*.}]

\item Study a published preference-pair dataset and inspect record structure, provenance, licence and the distinction between human and automatically generated preferences.

Choose one publicly documented set of preference pairs and explain what the preferred and dispreferred responses mean in its collection procedure. Inspect the prompt fields, provenance, language distribution and terms of access. Identify whether the judgements are direct human choices, model-produced rankings or mixed signals; do not conflate these sources. Sample several records and evaluate whether response quality differences are interpretable for the intended alignment objective.

\item Implement a data-quality audit for prompts, chosen/rejected responses and duplicate or malformed pairs. Prepare a reproducible training/validation split without leakage.

Write a repeatable validation routine that checks required fields, text lengths, blank or identical responses, duplicates and inconsistent prompt--response pairing. Split by stable prompt or conversation identifiers so closely related preference pairs cannot fall in both training and validation partitions. Report actual numbers of accepted and excluded examples, including the reason for each exclusion. Make clear whether the companion demonstration checks dataset structure only or also performs gradient-based model optimisation.

\item Describe and, where compute is available, execute a controlled DPO or reward-model experiment; compare with a non-adapted reference and report factual outputs only. State clearly if the demonstration validates data schema but does not train a policy.

Specify the chosen optimisation objective, reference policy, trainable parameter set, data budget, hardware and evaluation strategy before attempting a DPO or reward-model run. If resources allow, perform one controlled adaptation and compare with the same base checkpoint on a fixed evaluation set, reporting execution settings and observed outcomes. If not, present the full audit and an executable training plan without fictional losses or improvements. Distinguish preference modelling, DPO and reward-driven policy optimisation: these are related approaches but not interchangeable implementations.

\end{enumerate}

\subsection{Reproducible Example and Its Limits}

The companion project contains executable code for this work. Run the real-data path with
\begin{center}\texttt{python practikum\_examples.py 14}\end{center}
HuggingFaceH4/ultrafeedback\_binarized; the example inspects preference data but does not train DPO. For a technical-only check, add \texttt{-{}-smoke}; its synthetic inputs must not be used as scientific evaluation results. The program does not contain fabricated model scores. Record actual outputs, sample sizes, versions, timings and failures after running it.

\subsection{Extended Engineering Tasks (Optional)}
The following tasks provide a fuller engineering and research programme. They are optional rather than prerequisites for passing the core. The numerical scales and model choices below describe ambitious projects; they are not universal requirements. Documentation and reporting follow the shared protocol above.

\paragraph{Extended Task 1.} On the basis of the text corpus formed during Practical Work~\ref{work:1}, a multi-task dataset of prompts and model generations is constructed in the Russian and Tatar languages. The tasks cover question-answering dialogue, summarisation, and generation; for each task, at least one thousand prompts are collected, which, together with six generations per prompt (obtained from a preliminarily fine-tuned SFT model with varying temperature and top-$p$ sampling parameters), yields at least eighteen thousand ``prompt--answer'' pairs in total across all tasks. If the formation of the dataset in the indicated volume is objectively difficult --- for example, due to limited access to native speakers of Tatar for generation or verification, --- the learner has the right, in agreement with the instructor, to reduce the volume or to use synthetic generations from an LLM-as-judge (with subsequent selective expert verification of a subset), explicitly stipulating this circumstance in the datasheet and discussing its influence on the representativeness of the preferences in the analytical report. The learner also has the right to use ready-made preference datasets (e.g., open collections on Hugging Face Datasets) as a basis; the use of a ready-made dataset does not lower the grade but, on the contrary, is methodologically encouraged, since it makes it possible to concentrate on the comparative analysis of RLHF strategies. All data are saved in JSONL format with the fields \texttt{id}, \texttt{prompt}, \texttt{task\_type}, \texttt{language}, \texttt{generations} (a list of six answers), and references to the artefacts of the previous works are recorded in the methodological section of the report.

\paragraph{Extended Task 2.} A preference dataset of at least two thousand ranked pairs of answers is collected and verified, of which at least three hundred are in the Tatar language. Each pair is assessed according to three criteria: helpfulness (H), truthfulness (T), and harmlessness (S). For each criterion, a detailed annotation protocol with examples is formulated, and the file \texttt{preference\_collection\_protocol.md} is compiled --- a complete instruction for annotators with control questions and examples of borderline cases. A requirement is established: at least four to five independent assessments per pair; if it is impossible to engage human experts, the use of synthetic assessments from an LLM-as-judge is permitted, with subsequent manual verification of at least twenty per cent of the pairs. Inter-annotator agreement is measured using Krippendorff's $\alpha$; the minimum permissible threshold is $0.6$. Pairs for which the agreement is below the threshold are excluded from the training set and analysed separately as examples of subjective ambiguity. The final dataset is saved in the Hugging Face PreferenceDataset format. The annotation metadata (instructions for annotators, distribution of assessments by criteria, agreement matrices, demographic characteristics of annotators in aggregated form) are documented in the report.

\paragraph{Extended Task 3.} A software module \texttt{rlhf\_core.py} is developed, providing a unified interface for implementing four preference learning strategies. The first strategy --- classical RLHF with PPO: the TRL library or a custom implementation is used; the reward model is preliminarily trained on the collected preferences; the policy is optimised with KL regularisation. The second --- Direct Preference Optimization (DPO), implemented via TRL. The third --- Identity Preference Optimization (IPO). The fourth --- Cross-Entropy Method (CEM) with beam search and retraining, serving as a baseline without RL. A Multi-head Reward Model is implemented --- a model that produces three scalar outputs corresponding to helpfulness, truthfulness, and harmlessness, which makes it possible to analyse conflicts between the criteria. The supported base architectures are: ruT5-base (for generative tasks), Qwen1.5-0.5B (for DPO under resource constraints), as well as \texttt{cointegrated/rubert-tiny2} (for the reward model). All experiments are managed by a single YAML configuration file; logging is carried out via Weights \& Biases or MLflow. The module code is furnished with exhaustive documentation describing the interface of all public functions and configuration parameters.

\paragraph{Extended Task 4.} Rigorous training is performed with monitoring of the stability of the RL process and full visualisation control. For training the reward model, validation is performed by the area under the ROC curve and Spearman's rank correlation coefficient; learning curves are constructed for each of the three outputs (H, T, S) separately. Under PPO, the Kullback--Leibler divergence with respect to the reference policy is limited to a maximum value of $0.02$; early stopping is triggered in the absence of an improvement in the expert metric over 200 consecutive steps; and the entropy of the policy is continuously monitored. For DPO, the regularisation parameter $\beta$ is selected through a separate validation procedure. Training trajectory plots are constructed: the dependence of the reward, KL divergence, and target metric on the step number, as well as heat maps of conflicts between the criteria (H, T, S) --- situations in which an improvement on one criterion is accompanied by a deterioration on another. All artefacts --- model checkpoints, training logs, and UMAP projections of hidden states --- are saved at regular intervals.

\paragraph{Extended Task 5.} A multifaceted quality assessment of the aligned models is performed through three independent channels. The first channel --- automatic metrics: BERTScore, MAUVE, Self-BLEU, and FactScore, making it possible to assess the quality, diversity, and factual reliability of the generations. The second channel --- controlled expert annotation: on a sample of one hundred pairs of answers, three independent experts assign assessments on Likert scales (1--5) for each of the three criteria; additionally, a pairwise A/B test is performed, in which each of at least two hundred individual judgements records a preference between the answer of the SFT model and the answer of the RLHF model. The third channel --- behavioural analysis: the frequency of sycophantic phrases, mean answer length, lexical diversity (measured via MTLD --- Measure of Textual Lexical Diversity) are quantitatively measured, and a systematic categorisation of error types is performed. The results of all three channels are summarised in a single evaluation table \texttt{rlhf\_evaluation.csv}, the rows of which correspond to the ``strategy $\times$ task $\times$ criterion'' combinations, and the columns to the metrics.

\paragraph{Extended Task 6.} The risks of RLHF are identified and thoroughly documented. An audit of cultural biases is performed: the model is tested for gender and national stereotypes using template prompts. An avoidance behaviour test is performed: the model's tendency to avoid answering on topics of politics and religion is measured. Reward overoptimisation is analysed: a plot of the correlation between the reward model's assessments and expert human judgements is constructed, and points of divergence are identified where the reward model gives a high score to answers recognised by the experts as being of low quality. All results are compiled into a formal Ethics Appendix, which is included in the final report.

\paragraph{Extended Task 7.} The behavioural shifts caused by the alignment procedure are analysed. The $n$-gram frequency distributions of the model's answers before and after RLHF are compared. Using the Captum library, the activation patterns of neurons during the generation of answers of different types are investigated. Qualitative case studies are conducted: specific examples are selected in which the model's behaviour changed most dramatically, with special attention to the phenomena of reinforced avoidance of direct answers and the proliferation of general, vacuous formulations. UMAP projections of hidden states before and after RLHF are constructed, making it possible to visually assess how much the model's internal representation of the same prompts has changed. All visualisations are generated directly in the course of code execution and are included in the analytical report with extensive commentary.

\paragraph{Extended Task 8.} A formal verification of the correctness of the data splitting and the absence of information leakage is performed. The learner documents the splitting procedure with the fixation of the random seed, and programmatically verifies that the prompts from the test set were not used in training the reward model, and that the preference pairs from the validation set did not end up in the training set. It is controlled that the synthetic generations from the LLM-as-judge, if any were applied, do not intersect with the data on which the final expert assessment was performed. The verification protocol is included in the analytical report and serves as documentary confirmation of the methodological rigour of the experiment.

\paragraph{Extended Task 9.} A unified class \texttt{RLHFPipeline} is developed with the public methods \texttt{.train()} (launching the training of the chosen RLHF strategy with automatic logging and checkpointing), \texttt{.evaluate()} (return of a summary across all three evaluation channels --- automatic metrics, expert assessments, behavioural analysis), and \texttt{.explain()} (return of attention maps of the reward model, examples of the greatest changes in behaviour, UMAP projections of hidden states, and heat maps of criterion conflicts). The class is compatible with the TRL library and the Hugging Face ecosystem, supports the serialisation of state (reward model, policy, checkpoints) via \texttt{torch.save} and \texttt{transformers.save\_pretrained}, and records seeds for reproducibility checks, subject to system-dependent variation.

\paragraph{Extended Task 10.} An interactive platform for live A/B testing is created using Gradio. The platform provides a blind comparison of two models (the user does not know which model is hidden behind variants A and B), with an assessment according to three specified criteria (helpfulness, truthfulness, harmlessness) and the possibility of leaving a free-text comment. All user judgements are automatically saved in a Hugging Face Dataset. A trigger mechanism for the subsequent fine-tuning of the reward model on the freshly collected data is implemented, closing the online RLHF loop. The application is designed in such a way that it can be used by an expert linguist without programming skills and is deployed locally with the possibility of subsequent deployment to Hugging Face Spaces.

\paragraph{Extended Task 11.} A suite of unit tests is developed based on the pytest framework, verifying the correctness and reproducibility of the key project components. The tests must check: the fixation of seeds and library  (\texttt{trl}, \texttt{datasets}); the correctness of loading the PreferenceDataset; the stability of the metrics upon repeated runs; the complete absence of information leakage between the training and test sets at the stage of training the reward model; the correctness of computing the KL divergence in PPO; as well as the integrity of the online RLHF pipeline (correct saving and loading of user preferences). All dependencies are recorded in the \texttt{requirements.txt} file with an indication of the exact versions. A script is developed for the automatic execution of the full test suite and the generation of a pass/fail report. The presence of a successfully passing test suite is regarded as an integral component of the software artefact.

\subsection{Further Research Directions (Optional)}
\textbf{First Additional Task} --- Constitutional AI in Russian: generation of pairs using a rule (``the answer must be truthful'') and training of a reward model without human participation.

\textbf{Second Additional Task} --- cross-lingual transfer of the reward model: is a reward model trained on Russian data applicable to the evaluation of Tatar answers?

\textbf{Third Additional Task} --- Self-Play RLHF: the model generates pairs, evaluates them itself, and a human corrects only the most serious errors.

\textbf{Fourth Additional Task} --- analysis of attention heads in the reward model: which layers are responsible for safety and which for helpfulness?

\textbf{Fifth Additional Task} --- economics of RLHF: assessment of the CO\textsubscript{2} footprint and training cost of PPO versus DPO via CodeCarbon.

\textbf{Sixth Additional Task} --- RLHF + RAG: training a model to generate answers consistent with the retrieved context.

\textbf{Seventh Additional Task} --- decomposition of preferences: can the reward model be decomposed into ``style'' and ``content'' components?

\textbf{Eighth Additional Task} --- calibration of the reward model: application of Platt scaling to align the reward model's estimates with the probabilities of expert choice.

\section{Conclusion}
This work is not simply ``launching DPO in Colab''. It is a deep immersion into the philosophical, technical, and ethical foundations of modern AI alignment, where every decision --- from the formulation of criteria to the choice of $\beta$ in DPO --- carries a moral and methodological load.

In the course of performing the work, the learner learns to ask the key questions that determine the maturity of AI systems. Whose values the model embeds --- and whose it ignores (a question about cultural and linguistic pluralism). Whether the model becomes safer --- or simply more covert in its errors (a question about overoptimisation). Whether the reward model can be trusted --- or whether it captures only superficial patterns (a question about the depth of alignment). Whether reproducibility is guaranteed --- from the instruction to the annotator to the final A/B test (a question about scientific integrity).

The answers to these questions constitute the difference between a model that ``pleases'' and a system that serves society honestly, transparently, and responsibly. Upon completion of the work, the learner receives not simply an RLHF model but a methodological and ethical compass for designing, evaluating, and deploying any aligned AI systems --- from educational chatbots to medical assistants, capable of working not only intelligently but also fairly, transparently, and with respect for cultural diversity.

\chapter{LLM Agents and Multi-Agent Systems}
\label{work:15}

\section{Aim and Learning Outcomes}
This work examines the design and evaluation of systems that call tools and coordinate subtasks. Learners implement controlled actions, observe tool traces, compare with a no-tool baseline and analyse failures. The distinction between a deterministic controller and an autonomous LLM agent is explicit; unsafe or production side effects are outside the teaching environment.

The three tasks below constitute the required experimental core. Larger engineering studies and independent research questions are available as optional extensions.

\section{Theoretical Background}

\noindent\textit{Principal theoretical and methodological sources for this work:} \cite{owasp2025llmtop10,yao2023react,wu2023autogen,li2023agentbench}.\par

Modern large language models have opened a fundamentally new paradigm in artificial intelligence: instead of a passive text generator, the LLM becomes an active agent capable of perceiving the environment, planning actions, calling external tools, and adapting its behaviour based on feedback. An agent, unlike a simple chatbot, does not just answer a query but acts: it can initiate a search for information, perform calculations, interact with databases, call external APIs, and even coordinate the actions of other agents. This ability to act transforms the LLM from a text generation tool into a task-solving tool, which constitutes the essence of the modern paradigm of ``LLM as agent''. \cite{yao2023react,wu2023autogen,li2023agentbench}

The architecture of an agent includes four key components, each contributing critically to its functionality. The first component --- the Language Model (LLM) --- acts as the agent's ``brain'': it is responsible for understanding the query, planning the sequence of actions, generating tool calls, and interpreting the results obtained. The choice of LLM determines the quality of reasoning, generation speed, and cost of agent operation. The second component --- tools --- represents a set of external functions available to the agent: calculator, search system (including RAG search over the corpus), APIs for obtaining the current date and time, weather forecast, news, and other data. Tools can be implemented as local functions or as calls to external services. The third component --- memory --- is divided into short-term (history of the current dialogue, interaction context) and long-term (facts extracted from previous interactions and stored in a vector store). Memory allows the agent to preserve context and learn from previous mistakes. The fourth component --- the decision-making cycle --- determines how the agent uses the LLM, tools, and memory to achieve its goal. The cycle architecture can be linear (sequential execution of steps) or graph-based (with conditional transitions and cycles), which is supported by frameworks such as LangGraph and AutoGen.

Depending on the organisation of the decision-making cycle, several architectural patterns are distinguished, each with its own strengths and limitations. The ReAct (Reasoning + Acting) pattern is the most common and intuitive. The agent alternates reasoning steps (explaining what it intends to do and why) and action steps (calling a tool). This approach ensures transparency of the decision-making process and makes it easy to diagnose errors, since the reasoning trajectory is available for analysis. However, ReAct is not always optimal: for complex multi-component tasks, the agent may loop or spend excessive steps refining the plan. The Plan-and-Solve pattern offers an alternative: the agent first develops a complete action plan (as a sequence of steps or a structured description), and then sequentially executes this plan. This reduces the number of iterative reasoning steps and makes the agent's behaviour more predictable; however, the plan may be suboptimal if the original task was poorly formalised. The Reflexion pattern introduces a self-reflection mechanism: if the agent encounters an error, it analyses the cause, formulates a new plan, and makes another attempt, similar to how a human learns from experience. Reflection significantly increases the success rate of complex tasks, but requires additional LLM calls to generate reflective texts, increasing cost and latency.

The development of agent systems has led to the emergence of multi-agent architectures, where several specialised agents interact to solve complex tasks beyond the capabilities of a single agent. In a hierarchical multi-agent system, a supervisor (coordinator) is distinguished, which receives the original query, analyses its structure, determines which subtasks need to be solved, and distributes them among specialised worker agents. Worker agents can be narrowly specialised: a maths agent solves computational tasks, a search agent queries the RAG system or external search, a summarisation agent compresses voluminous information. The supervisor collects responses from the worker agents and forms the final answer to the user. Other formats of multi-agent interaction include horizontal coordination (agents interact on equal footing, without an explicit leader) and adversarial scenarios (debates between agents to identify the most justified solution). Multi-agent architectures open new possibilities for solving tasks requiring simultaneous consideration of different perspectives; however, they create new challenges --- the need for coordination, reconciliation of conflicting answers, and prevention of endless communication cycles.

The effectiveness of an agent system is determined not only by the choice of LLM and tools, but also by the design of prompts for each pattern. The prompt must clearly define the agent's roles, describe the available tools (with indication of the call format and return value), specify the reasoning format (e.g., ``Step 1: reasoning... Action: tool(arguments)... Observation: result...''), and also limit the trajectory length and indicate how the agent should complete its work. For multi-agent systems, the supervisor's prompt additionally describes the rules for task distribution and the format of worker agents' responses. The quality of prompts directly affects the success of task execution, and their refinement is often a key stage in the development of agent applications.

\subsection{Experimental Interpretation}
A rule-based controller with access to a retrieval tool illustrates tool calling, not necessarily autonomous LLM planning. A model's output or a retrieved page must not be trusted to grant permissions. Evaluate tool-selection errors and real side effects only in an authorised isolated environment. A fixed random seed does not guarantee deterministic outputs from remote LLM services.

\section{Execution Procedure}

\subsection{Core: Three Required Tasks}
\begin{enumerate}[leftmargin=*,label=\textbf{\arabic*.}]

\item Build a controlled tool-use workflow with a search tool and an explicit allowlist for permissible actions; record the allowed inputs and outputs.

Define a bounded user task and create a tool interface that accepts an explicit schema rather than arbitrary executable instructions. Use synthetic or freely shareable documents, an allowlist of permitted operations and fixed rules for rejecting invalid arguments. Build a controller whose states and transitions can be inspected. If it contains no language model, name it a deterministic tool-use baseline rather than presenting it as evidence of autonomous planning.

\item Evaluate it on a fixed set of ordinary tasks, logging calls, arguments, selected documents and failure cases.

Construct a fixed, documented evaluation set containing ordinary successful requests, malformed inputs and actions that should be rejected. Log each tool choice, argument, retrieved item, returned result and final response, avoiding storage of secrets. Measure at least task completion and inappropriate-tool-call rates using explicit counts and denominators. Keep tools non-destructive and local unless an authorised sandbox permits otherwise.

\item Compare the controller with a no-tool baseline and distinguish a deterministic tool controller from an autonomous LLM agent. Never grant production permissions in a classroom demonstration.

Use the same test cases to compare the tool-enabled controller with a no-tool baseline, marking cases where external evidence materially changes an answer. Analyse error propagation across decisions and whether a failed call was safely rejected or silently accepted. Explain what additional components a genuine LLM agent would require, such as model-driven planning, tool routing and termination rules, without assuming that adding several agents necessarily improves reliability.

\end{enumerate}

\subsection{Reproducible Example and Its Limits}

The companion project contains executable code for this work. Run the real-data path with
\begin{center}\texttt{python practikum\_examples.py 15}\end{center}
Controlled retrieval tool; a rule-based controller rather than an autonomous model agent. For a technical-only check, add \texttt{-{}-smoke}; its synthetic inputs must not be used as scientific evaluation results. The program does not contain fabricated model scores. Record actual outputs, sample sizes, versions, timings and failures after running it.

\subsection{Extended Engineering Tasks (Optional)}
The following tasks provide a fuller engineering and research programme. They are optional rather than prerequisites for passing the core. The numerical scales and model choices below describe ambitious projects; they are not universal requirements. Documentation and reporting follow the shared protocol above.

\paragraph{Extended Task 1.} On the basis of the text corpus formed during Practical Work~\ref{work:1} and the RAG system built in Practical Work~\ref{work:12}, an experimental base for agent systems is prepared. The corpus serves as a source of factual information for the search agent; the RAG system provides an interface for semantic search over the corpus. Additionally, a set of test tasks for evaluating agents is created, containing at least fifty tasks classified by type: computational (requiring the use of a calculator), search (requiring access to the RAG system), combined (requiring both computations and search), multi-step (requiring sequential execution of several actions), and planning tasks (requiring the development of an action plan). Each task is supplied with a reference answer and, where necessary, a reference solution trajectory (sequence of steps). All test tasks are saved in JSONL format with the fields \texttt{id}, \texttt{task\_type}, \texttt{prompt}, \texttt{expected\_answer}, \texttt{expected\_steps} (optional), \texttt{difficulty} (basic, intermediate, advanced). References to the artefacts of the previous works (RAG system, trained LLMs) are recorded in the methodological section of the report.

\paragraph{Extended Task 2.} A set of tools for the agent is developed, implemented as separate functions with the \texttt{@tool} decorator from LangChain or as classes compatible with the tool interface. The minimum set of tools includes: a calculator --- a function for performing arithmetic operations with support for complex expressions (using the SymPy library or Python eval with restrictions); search over the corpus --- a function that accesses the RAG system from Work~\ref{work:12} and returns relevant fragments with metadata; current date and time --- a function returning the current date and time in ISO format; summarisation --- a function that compresses long text into a brief summary (may use a separate summarisation model from Work~\ref{work:8} or the agent's own LLM). Additional tools (weather forecast, news) are implemented at the learner's discretion, provided access to the corresponding open APIs is available. Each tool is supplied with a description (for inclusion in the agent prompt), a definition of the input and output data format, and error handling. The tool code is saved in the module \texttt{tools.py}.

\paragraph{Extended Task 3.} A software module \texttt{agents\_core.py} is developed, providing a unified interface \texttt{AgentSystem} for three architectural patterns. The first pattern --- ReAct --- is implemented using the standard AgentExecutor from LangChain with a prompt specifying the format ``Thought: ... Action: ... Observation: ...''; the use of a ready-made implementation from LangChain is permitted. The second pattern --- Plan-and-Solve --- is implemented through two-phase execution: first, an action plan is generated, then the plan is executed step by step with the possibility of adjusting it as necessary. The third pattern --- Reflexion --- is implemented as an extension of ReAct: when an error occurs (unsuccessful tool call, incorrect answer), the agent analyses the error, generates a reflective text, and makes another attempt; the maximum number of attempts is fixed (recommended: 3). The supported base models are: Llama-3-8B-Instruct, Mistral-7B-Instruct, Qwen1.5-7B-Chat (loaded via \texttt{llm\_hub.py} from Work~\ref{work:12}), as well as any locally deployed open LLM. The module must support configuration via a YAML file, logging of trajectories (saving all reasoning and action steps), and visualisation of execution graphs for LangGraph implementation. The module code is furnished with exhaustive documentation describing the interface of all public functions and configuration parameters.

\paragraph{Extended Task 4.} An agent with short-term memory is implemented based on ConversationBufferMemory from LangChain or an analogous component that stores dialogue history within a single session. The agent is tested on a sequence of three to five related queries, where the answer to a subsequent query depends on information provided in previous ones. It is verified that the agent correctly uses history to answer clarifying questions and refers to previously obtained facts. A report is prepared on the accuracy of answers to related queries compared to an agent without memory. The testing results are presented in the form of a table and included in the analytical report.

\paragraph{Extended Task 5.} Long-term memory is added based on a vector store (FAISS or Qdrant). Facts extracted from dialogues are saved as embeddings with metadata (source, time, fact type). For each new query, the agent performs semantic search over the vector memory and includes the most relevant facts in the prompt. Long-term memory allows the agent to ``remember'' information over long periods of time and between different sessions. The agent's ability to reproduce facts reported in previous sessions after several days is tested (simulated by running the agent with pre-loaded memory). The accuracy of fact reproduction and search speed (latency) are evaluated. The results are included in the report.

\paragraph{Extended Task 6.} A multi-agent system with a hierarchical architecture is implemented. The supervisor is an agent that receives a query and determines which subtasks need to be solved; it distributes tasks among three specialised worker agents: a maths agent (specialises in computational tasks, has access to a calculator), a search agent (specialises in searching for information over the corpus, has access to the RAG system), and a summarisation agent (specialises in compressing voluminous texts, has access to a summarisation function and/or a separate model). The supervisor forms the final answer based on the results of the agents' work. The implementation is performed using LangGraph, where each agent is represented as a separate graph node, and transitions between nodes are managed by the supervisor based on query analysis. The graph is saved in visualised form and included in the report.

\paragraph{Extended Task 7.} Systematic testing of all implemented agent architectures (ReAct, Plan-and-Solve, Reflexion) is performed on the set of test tasks from Task 1. For each pattern and each model, the following metrics are computed: task execution accuracy (proportion of correctly solved tasks) with breakdown by task types; average number of steps to completion; average response time; token consumption (input and output total); success rate of tool calls (proportion of correctly formatted and executed calls); for Reflexion --- average number of attempts to successful completion. The results are summarised in a single table \texttt{agent\_metrics.csv}, the rows of which correspond to the ``pattern $\times$ model $\times$ task type'' combinations, and the columns to the metrics. Additionally, box plots of response time distribution and number of steps for each pattern are constructed, as well as bar charts of accuracy with breakdown by task difficulty.

\paragraph{Extended Task 8.} An in-depth error analysis and interpretation of decision trajectories are performed. For each pattern and task type, the most frequent error types are identified: incorrect tool selection, incorrect formatting of tool calls, incomplete answer, looping (exceeding the maximum number of steps), hallucination (assertion of a fact without tool support). Error matrices are constructed by task types and difficulty. For erroneous trajectories, a qualitative analysis is performed: five to ten examples are selected where the agent incorrectly selected a tool or looped, and the causes are analysed (insufficient prompt, LLM limitations, incorrect data). For visualising trajectories, LangGraph graph representations (for Plan-and-Solve and the multi-agent system) and text logs of ReAct are used, which are converted into readable timelines. All visualisations are generated directly in the course of code execution and are included in the analytical report with extensive commentary.

\paragraph{Extended Task 9.} A comparative study of the effectiveness of different base models in the role of an agent is performed. The learner selects three open LLMs (e.g., Llama-3-8B-Instruct, Mistral-7B-Instruct, Qwen1.5-7B-Chat) and tests them on the same set of tasks with a fixed pattern (ReAct). The comparison is conducted using the same metrics as in Task 7. Additionally, the extent to which model size affects reasoning quality and task success is evaluated. A ``quality -- model size -- speed'' plot is constructed, on which all three models are marked as points. Based on the obtained data, recommendations are formulated for choosing a model for specific agent use cases.

\paragraph{Extended Task 10.} An interactive web application is created using Gradio for demonstrating and comparing agents. The application must provide the user with the following capabilities: entering an arbitrary query in natural language; selecting an architectural pattern (ReAct, Plan-and-Solve, Reflexion) and a base model; displaying the agent's generated answer; step-by-step display of the execution trajectory with indication of tools used and intermediate reasoning (as a timeline or graph); for the multi-agent system --- visualisation of task distribution among agents and the results of each agent; side-by-side parallel comparison of two patterns or two models; automatic generation of an HTML report containing the query, answer, trajectory, and metrics. The application is designed in such a way that it can be used by a specialist without programming skills and is deployed locally with the possibility of subsequent deployment to Hugging Face Spaces.

\paragraph{Extended Task 11.} A formal verification of the correctness of the data splitting and the absence of information leakage is performed, and a unified class \texttt{AgentSystem} is developed. The learner documents the procedure for forming test tasks, fixing the seed for the random number generator when selecting them from the corpus. It is programmatically verified that the test tasks were not used in training the LLMs used as base models for the agents (which is guaranteed by using open pre-trained models not fine-tuned on the corpus). It is controlled that the tools (in particular, the RAG search) do not have access to the test tasks as part of the index. The verification protocol is included in the analytical report. In parallel, the class \texttt{AgentSystem} is developed with the public methods \texttt{.run(prompt)} (running the agent on a single query), \texttt{.evaluate(task\_set)} (running the agent on a set of tasks with return of metrics), and \texttt{.explain()} (return of the trajectory of the last execution in structured form, including all reasoning and action steps). The class is compatible with the LangChain and LangGraph libraries, supports the serialisation of state (tools, memory, configuration) via \texttt{pickle} and \texttt{joblib}, and records seeds for reproducibility checks, subject to system-dependent variation.

\subsection{Further Research Directions (Optional)}
\textbf{First Additional Task} --- agent with SQL database access: implementation of an agent capable of generating SQL queries from natural language descriptions, executing them, and interpreting the results, using a text file or SQLite as a database. Evaluation of accuracy and security (prevention of SQL injection).

\textbf{Second Additional Task} --- negotiation agent in a multi-agent environment: two agents negotiate the distribution of resources (e.g., budget between projects), each with its own utility function; the negotiation strategy and the compromise reached are analysed.

\textbf{Third Additional Task} --- integration with external APIs: expansion of the toolset by connecting to weather forecast, news, or social media APIs; evaluation of the agent's robustness to failures of external services.

\textbf{Fourth Additional Task} --- use of local function calling: implementation of a tool calling mechanism without using LangChain's built-in capabilities, only through a specially constructed prompt and parsing of the LLM response; comparison of quality with the LangChain implementation.

\textbf{Fifth Additional Task} --- prompt optimisation for agents: systematic A/B testing of different prompt formulations (detailed vs concise tool descriptions, explicit vs implicit output format specification) and assessment of their influence on task success.

\textbf{Sixth Additional Task} --- agent with visual interface: extension of the web application with real-time visualisation of the decision-making graph, animation of the agent's steps, and interactive control (ability to stop the agent, change tools on the fly).

\textbf{Seventh Additional Task} --- assessment of agent energy consumption: measurement of the carbon footprint of different architectures (ReAct vs Plan-and-Solve) via CodeCarbon when executing one hundred queries; formulation of recommendations for energy efficiency.

\textbf{Eighth Additional Task} --- agent with learning from feedback: integration of a human feedback collection mechanism (similar to Work~\ref{work:14}) to improve the agent's reasoning quality; comparison of the effectiveness of RL fine-tuning and prompt engineering.

\section{Conclusion}
This work is not simply ``running a ReAct agent from the LangChain documentation''. It is a deep immersion into the architectural, algorithmic, and engineering foundations of modern agent systems, where every decision --- from the choice of pattern to the set of tools and prompt formulation --- determines the agent's ability to solve tasks, its transparency, and reliability.

In the course of performing the work, the learner learns to ask the key questions that determine the maturity of modern AI systems. Whether the agent can plan its actions in advance --- or whether it is forced to act by trial and error (a question about planning strategy). Whether the trajectory the agent chose can be trusted --- or whether it is looping on a suboptimal path (a question about interpretability and error diagnosis). How an error in one step affects the entire trajectory (a question about the robustness of agent systems). Whether reproducibility is guaranteed --- from the first prompt to the final graph (a question about scientific integrity and engineering discipline).

The answers to these questions constitute the difference between a ``demo agent'' and a responsible intelligent system capable of autonomously solving complex tasks in a dynamic environment. Upon completion of the work, the learner receives not simply a set of scripts but an architectural compass for designing, evaluating, and deploying any agent systems --- from corporate assistants to research platforms, capable not only of acting but also of explaining their actions, learning from mistakes, and interacting with other agents to achieve a common goal.

\chapter{Security and Reliability of RAG and Agentic Systems}
\label{work:16}

\section{Aim and Learning Outcomes}
This work investigates the boundaries of trust in retrieval-augmented systems and agents with external tools. Learners examine how untrusted documents and tool responses can attempt to redirect model behaviour, how permissions are enforced outside the language model, and how an isolated evaluation can quantify failures without exposing real users or production systems. The practical scope follows documented generative-AI risk-management guidance from NIST and the OWASP Top 10 for LLM Applications \cite{nist2024genai,owasp2025llmtop10}.

\section{Theoretical Background}

\noindent\textit{Principal theoretical and methodological sources for this work:} \cite{nist2024genai,ru_reference_67}.\par

A RAG system combines retrieval from a collection of documents with answer synthesis. Retrieved pages, snippets, hyperlinks and tool outputs are untrusted inputs; instructions appearing in these materials must not acquire the authority of system policy. An agentic system introduces additional boundaries because tool calls can access files, APIs or external services. Technical controls include limiting permissions, validating tool arguments, restricting the number and scope of operations, logging results, and requiring human approval for consequential actions.

Security and task performance must be measured separately. Executing a prohibited action is an unsafe outcome, while denying an authorised task is a false refusal. Define both denominators before testing; record the test configuration, sample size and whether the expected outcome was determined independently. Filtering prompts or telling a model to ignore hostile text does not provide a formal proof of protection. Tests should use fictional secrets, synthetic documents and stub tools with no ability to send email, move money, delete real files or alter live services.

\section{Execution Procedure}

\subsection{Core: Three Required Tasks}
\begin{enumerate}[leftmargin=*,label=\textbf{\arabic*.}]

\item Draw a trust-boundary diagram for one RAG or tool-using system. Treat retrieved pages, model outputs and tool responses as untrusted data; use only synthetic credentials and non-destructive tools.

Map every point where instructions, user inputs, retrieved passages, model output or tool results cross trust boundaries. Give each component a stated authority level and identify which actions require independent code-level validation. Use only fictitious credentials and local non-destructive tools. The diagram should make it clear why content retrieved from a page or returned by a tool is evidence to process, not an authorised instruction to modify permissions.

\item Design at least ten authorised tests with explicit expected behaviour, including benign requests and simulated indirect prompt-injection attempts in test documents. Enforce an allowlist and argument checks at the tool boundary.

Design a test set with legitimate tasks, harmless malformed inputs and simulated malicious instructions embedded in untrusted documents. Define the permitted response and any explicitly prohibited tool action for each case before the test is run. Enforce allowlists, typed argument validation and privilege limits outside the model prompt, then record whether the boundary was respected. Do not run these examples against unapproved production services or real personal data.

\item Compare an unprotected demonstration with a defended configuration using rates of disallowed actions and false refusals with explicit denominators. A passing test suite is not proof of complete security.

Apply identical tests to a deliberately minimal demonstration and an otherwise comparable defended configuration. Report the number of valid tasks completed, incorrect refusals and unauthorised attempted or completed actions with clearly defined denominators. Include examples of both successful defence and defence failure, and describe which attacks were not covered. A limited suite demonstrates behaviour under specified tests; it does not prove general immunity to prompt injection or data exfiltration.

\end{enumerate}

\subsection{Reproducible Example and Its Limits}

The companion project contains executable code for this work. Run the real-data path with
\begin{center}\texttt{python practikum\_examples.py 16}\end{center}
Synthetic adversarial documents, local tool permission tests; not a public security benchmark. For a technical-only check, add \texttt{-{}-smoke}; its synthetic inputs must not be used as scientific evaluation results. The program does not contain fabricated model scores. Record actual outputs, sample sizes, versions, timings and failures after running it.

\begin{lstlisting}[style=pythonex]
allowed_tools = {'search'}
checks = [
    ('ordinary request', 'search', True),
    ('[UNTRUSTED] ignore policy', 'delete', False),
    ('[UNTRUSTED] reveal secrets', 'reveal_secret', False),
]
for source, tool, expected in checks:
    assert (tool in allowed_tools) == expected
# This checks tool permissions; it is not a proof against injection.
\end{lstlisting}

\subsection{Extended Engineering Tasks (Optional)}

\paragraph{Extended Task 1.} Map users, trusted instructions, documents, retrieval components and permitted tools; add a list of concrete failure consequences.

\paragraph{Extended Task 2.} Prepare a non-destructive regression suite with ordinary and negative cases, expected outcomes and stable IDs.

\paragraph{Extended Task 3.} Test indirect instructions embedded in synthetic passages; log which source was retrieved and whether the model treated it as an instruction.

\paragraph{Extended Task 4.} Apply least-privilege permissions, strict argument validation, tool-call ceilings and a manual approval gate for consequential actions.

\paragraph{Extended Task 5.} Compare two defences separately and jointly using the same test cases.

\paragraph{Extended Task 6.} Compute the attack success rate and false-refusal rate with explicitly documented denominators and an inspection of automatic judging errors.

\paragraph{Extended Task 7.} Distinguish failures in retrieval, generation, tool handling and permission checking.

\paragraph{Extended Task 8.} Package the synthetic dataset, sandbox configuration, tests and threat record so another student can rerun the safety evaluation.

\subsection{Further Research Directions (Optional)}
\paragraph{Model and domain transfer.} Compare the same fixed set of indirect prompt-injection and ordinary requests across multiple accessible models while holding retrieval, tool policy and evaluation criteria constant. Record model revisions and decoding settings, and examine when any observed defence fails to transfer.
\paragraph{Human approval and least privilege.} Measure the change in prohibited actions, false refusals and latency when a sandboxed agent uses tool allowlists, argument checks and human approval. Do not assume a mechanism is universally effective from a narrow benchmark.
\section{Conclusion}
A trustworthy RAG or agentic application depends on both verifiable evidence handling and enforceable tool permissions. Safety assessments describe observed behaviour for specific threats and configurations; they are not general certificates of immunity to prompt injection.

\chapter{Model Compression and Efficient Inference}
\label{work:17}

\section{Aim and Learning Outcomes}
This work studies the practical trade-offs between storage, accuracy and latency in efficient model inference. Students quantify the effects of supported dynamic INT8 quantisation on a compact trained text classifier, control benchmarking conditions and interpret the results. LLM-specific techniques such as GPTQ, AWQ and advanced serving remain separate optional research topics.

The three tasks below constitute the required experimental core. Larger engineering studies and independent research questions are available as optional extensions.

\section{Theoretical Background}

\noindent\textit{Principal theoretical and methodological sources for this work:} \cite{frantar2022gptq,lin2023awq,dao2022flashattention,ru_reference_71,ru_reference_72}.\par

The remarkable capabilities of large language models come at a significant cost: billions of parameters, hundreds of gigabytes of memory, and substantial inference latency. While these models achieve state-of-the-art performance, their size and computational requirements make them impractical for many real-world applications. Model compression and efficient inference address this gap, transforming large research models into deployable production systems. The discipline encompasses a spectrum of techniques that reduce model size, accelerate inference, and minimise memory consumption, often with minimal degradation in quality. \cite{frantar2022gptq,lin2023awq,dao2022flashattention}

Quantisation is one of the most effective and widely used compression techniques. It reduces the precision of model weights and activations from 32-bit floating point (FP32) to lower bit widths: 16-bit (FP16, BF16), 8-bit (INT8), 4-bit (INT4, NF4), or even 2-bit. Post-Training Quantisation (PTQ) applies quantisation to a pre-trained model without retraining, using calibration data to determine optimal quantisation parameters. Quantisation-Aware Training (QAT) simulates quantisation during training, allowing the model to adapt to the reduced precision and often achieving better performance than PTQ. Advanced methods such as GPTQ (Generalised Post-Training Quantisation) and AWQ (Activation-aware Weight Quantisation) optimise quantisation by considering the impact of weight quantisation on activations, achieving high accuracy at 4-bit and even 3-bit precision.

Model distillation transfers knowledge from a large teacher model to a smaller student model. In Knowledge Distillation (KD), the student is trained to match the teacher's output probabilities (soft labels) in addition to the ground-truth labels, effectively learning the teacher's decision boundaries and uncertainty. Task-specific distillation focuses on the target task, using both the teacher's predictions and intermediate representations to guide the student. On-device distillation further optimises the student for deployment on edge devices, considering hardware constraints and inference latency. Distillation can achieve significant compression ratios (10x--100x) with relatively modest performance degradation, making it ideal for deploying models on mobile phones, embedded devices, and other resource-constrained platforms.

Pruning removes redundant or unimportant parameters from the model. Unstructured pruning (magnitude-based, movement pruning) removes individual weights based on their importance (e.g., magnitude, gradient), resulting in sparse weight matrices that can be accelerated with specialised hardware or libraries. Structured pruning (channel pruning, layer pruning) removes entire structures (channels, heads, layers), maintaining dense matrix operations and achieving speedups without specialised hardware. Pruning can be applied during training (sparse training), after training (post-training pruning), or iteratively (gradual pruning). The sparsity rate can be adjusted to balance compression and performance, with modern methods achieving 50

Efficient attention mechanisms address the quadratic complexity of standard self-attention. Flash Attention optimises the attention computation through tiling and recomputation, reducing memory bandwidth and achieving up to 5x--10x speedups for long sequences. Multi-Query Attention (MQA) shares key and value projections across attention heads, reducing memory and computation. Grouped-Query Attention (GQA) is a hybrid that groups heads into groups with shared keys and values, balancing performance and efficiency. These mechanisms enable longer context windows, faster inference, and reduced memory consumption, making them essential for modern LLM deployment.

Inference optimisation techniques further accelerate model execution. Speculative decoding uses a small draft model to generate candidate tokens, which are then verified by the larger target model in parallel, effectively hiding the latency of the large model. Continuous batching (or dynamic batching) processes multiple requests in a single batch, maximising GPU utilisation. Tensor parallelism distributes model parameters across multiple GPUs, enabling the deployment of very large models. Pipeline parallelism partitions the model across devices, overlapping computation and communication. These techniques are essential for serving LLMs in production environments with high throughput and low latency requirements.

The evaluation of compression techniques requires a comprehensive framework that measures multiple dimensions: model size (number of parameters, memory footprint); inference latency (time per token, time per request); throughput (tokens per second, requests per second); memory consumption (peak memory, memory bandwidth); task performance (accuracy, F1, ROUGE, BLEU); energy consumption (watts, carbon footprint); and deployment cost (hardware, cloud, maintenance). Understanding the trade-offs between these dimensions is essential for choosing the right compression technique for a given application scenario. A systematic evaluation framework enables informed decision-making and rigorous comparison of different compression approaches.

\subsection{Experimental Interpretation}
Quantisation, pruning, distillation and efficient attention change different components of a system. Compare them under matched hardware, batching, precision and evaluation data. A smaller serialised model can have slower inference because actual speed depends on kernels, device support, batch sizes and overhead. Dynamic INT8 on a compact CPU classifier must not be called a benchmark of large-model GPTQ or AWQ.

\section{Execution Procedure}

\subsection{Core: Three Required Tasks}
\begin{enumerate}[leftmargin=*,label=\textbf{\arabic*.}]

\item Train a compact, legitimate neural text classifier with a held-out test set and record its baseline file size, quality and inference setting.

Select a compact neural text classifier with publicly documented code and weights or train one under a controlled protocol. Use a fixed held-out test set and capture the initial checkpoint, parameter count, serialised size and device details. Choose representative input lengths and establish a repeatable CPU inference procedure with warm-up and repeated measurements. This forms a baseline before compression; avoid mixing training-time changes with quantisation effects.

\item Apply documented dynamic INT8 quantisation to supported layers and test the quantised classifier on the same inputs and CPU.

Apply a documented dynamic quantisation procedure only to supported neural layer types, recording framework and backend versions. Verify that the quantised model loads and produces outputs of the expected shape and label mapping. Evaluate it on the same test item identifiers and preprocessing path as the original. If a layer or device is unsupported, report the actual limitation rather than implying successful quantisation of the whole architecture.

\item Report macro-F1, serialised model size and robust repeated latency measurements. Explain why smaller weights do not guarantee lower latency and why this does not establish GPTQ/AWQ performance for LLMs.

Calculate accuracy or macro-F1 as appropriate, file size in clearly defined units, and latency summaries from repeated measurements with warm-up and synchronisation where relevant. Compare distributions, not a single conveniently fast run, and report threading and batch-size settings. Discuss possible accuracy loss, model-loading overhead and hardware-specific speedups or slowdowns. Explicitly separate this small-model INT8 experiment from GPTQ, AWQ or low-bit inference for large transformer checkpoints, which require different procedures and evidence.

\end{enumerate}

\subsection{Reproducible Example and Its Limits}

The companion project contains executable code for this work. Run the real-data path with
\begin{center}\texttt{python practikum\_examples.py 17}\end{center}
20 Newsgroups; FP32 versus dynamic INT8 on a small PyTorch model on CPU. For a technical-only check, add \texttt{-{}-smoke}; its synthetic inputs must not be used as scientific evaluation results. The program does not contain fabricated model scores. Record actual outputs, sample sizes, versions, timings and failures after running it.

\subsection{Extended Engineering Tasks (Optional)}
The following tasks provide a fuller engineering and research programme. They are optional rather than prerequisites for passing the core. The numerical scales and model choices below describe ambitious projects; they are not universal requirements. Documentation and reporting follow the shared protocol above.

\paragraph{Extended Task 1.} On the basis of the text corpus formed during Practical Work~\ref{work:1} and the labelled datasets from Practical Work~\ref{work:4}, Work~\ref{work:5}, and Work~\ref{work:13}, a comprehensive evaluation dataset for compression is created. The dataset must include at least one thousand examples spanning three task types: classification (sentiment, topic), generation (summarisation, translation), and reasoning (arithmetic, commonsense). Each example is saved in JSONL format with the fields: \texttt{id} --- unique identifier; \texttt{task\_type} --- task category; \texttt{input} --- the input text; \texttt{expected\_output} --- the reference answer; \texttt{metadata} --- additional information (difficulty, domain, etc.). The dataset is split into training (for QAT and distillation), validation (for tuning hyperparameters), and test (for final evaluation) sets in a $60/20/20$ ratio. The random seed is fixed and documented to ensure reproducibility.

\paragraph{Extended Task 2.} A software module \texttt{compression\_utils.py} is developed, providing essential utilities for model compression. The module must include: functions for loading models from Hugging Face; functions for measuring model size (parameters, memory footprint); functions for measuring inference latency and throughput (with warm-up and multiple iterations); functions for measuring memory consumption (GPU memory, CPU memory); functions for computing compression ratio (original size / compressed size); and a logging system for tracking compression parameters, metrics, and results. The module is designed to be model-agnostic and compatible with both encoder-only and decoder-only architectures. The code is furnished with exhaustive documentation describing the interface of all public functions and configuration parameters.

\paragraph{Extended Task 3.} A comprehensive comparison of quantisation techniques is performed. The learner selects at least three open LLMs (e.g., Llama-3-8B, Mistral-7B, Phi-3-mini) and applies quantisation using four methods: FP16 (baseline), GPTQ (4-bit, 3-bit, 2-bit), AWQ (4-bit, 3-bit), and BitsAndBytes (4-bit NF4). For each model and quantisation method, the learner measures model size, inference latency, memory consumption, and task performance on all three task types from Task 1. The results are summarised in a table \texttt{quantisation\_comparison.csv}. Performance degradation curves are constructed showing accuracy as a function of bit width. The learner analyses the trade-offs between compression ratio, performance, and inference speed, identifying the optimal quantisation method for each model and task type.

\paragraph{Extended Task 4.} Post-Training Quantisation (PTQ) is implemented and evaluated. The learner selects two models (Llama-3-8B and Phi-3-mini) and applies GPTQ quantisation with varying calibration dataset sizes (128, 512, 2048 samples). The learner analyses the influence of calibration data on quantisation quality, measuring performance on the validation set and the distribution of quantisation errors. The results are visualised in a scatter plot showing performance as a function of calibration set size and bit width. The learner formulates conclusions about the minimum calibration data required for stable quantisation and the sensitivity of different models to calibration data quality.

\paragraph{Extended Task 5.} Quantisation-Aware Training (QAT) is implemented and compared with PTQ. The learner selects a small model (e.g., Phi-3-mini or a fine-tuned BERT) and applies QAT using the \texttt{torchao} or \texttt{brevitas} library. The model is fine-tuned on the training set for 5--10 epochs with simulated quantisation (8-bit, 4-bit). The learner compares QAT with PTQ (using the same bit width) and with the uncompressed baseline. The metrics include accuracy, training time, and quantisation error. Learning curves are constructed showing training accuracy and validation accuracy as a function of epoch for QAT and fine-tuning without quantisation. The learner analyses the trade-offs between QAT and PTQ, concluding when QAT is worth the additional training cost.

\paragraph{Extended Task 6.} Model distillation is implemented and evaluated. The learner selects a large teacher model (e.g., Llama-3-8B) and a smaller student model (e.g., Phi-3-mini or DistilBERT). Knowledge distillation is implemented using the \texttt{transformers} and \texttt{peft} libraries. The student is trained on the training set for 5--10 epochs, with the teacher providing soft labels. Three variants are compared: (1) standard distillation (matching teacher logits); (2) task-specific distillation (fine-tuning the student on the task, with teacher soft labels); and (3) on-device distillation (further optimised for the target hardware). The results are summarised in a table \texttt{distillation\_comparison.csv}. The learner analyses the quality-efficiency trade-off, calculating the compression ratio and performance retention relative to the teacher.

\paragraph{Extended Task 7.} Pruning techniques are implemented and compared. The learner selects a model (e.g., Mistral-7B or a fine-tuned BERT) and applies three pruning methods: (1) magnitude-based unstructured pruning (removing weights with the smallest magnitude) with sparsity rates of 30

\paragraph{Extended Task 8.} Efficient attention mechanisms are implemented and evaluated. The learner selects a model (e.g., Mistral-7B or a fine-tuned BERT) and replaces the standard multi-head attention with Flash Attention (using the \texttt{flash-attn} library), Multi-Query Attention (MQA), and Grouped-Query Attention (GQA). For each variant, the learner measures inference latency, memory consumption, and task performance for different sequence lengths (512, 1024, 2048, 4096). The results are visualised in a line chart showing latency as a function of sequence length for each attention variant. The learner analyses the impact of efficient attention on long-context tasks, comparing the quality and speed of different attention mechanisms.

\paragraph{Extended Task 9.} Inference optimisation techniques are implemented and benchmarked. The learner selects a model (Llama-3-8B or Mistral-7B) and implements three inference optimisation techniques: (1) speculative decoding (using a small draft model with 2--5 tokens of speculation); (2) continuous batching (processing multiple requests in dynamic batches); (3) tensor parallelism (splitting the model across two GPUs). The learner measures throughput (requests per second, tokens per second), latency (p50, p95, p99), and memory consumption for each technique. The results are summarised in a table \texttt{inference\_optimisation\_comparison.csv}. The learner analyses the effectiveness of each technique for different batch sizes and sequence lengths, identifying the optimal configuration for high-throughput and low-latency scenarios.

\paragraph{Extended Task 10.} A comprehensive compression pipeline is developed, combining multiple compression techniques. The learner selects a model (e.g., Mistral-7B) and applies a sequence of compression techniques: first pruning (structured, 50

\paragraph{Extended Task 11.} A benchmarking framework for compression techniques is developed and published. The framework must support: loading models from Hugging Face; applying compression techniques (quantisation, pruning, distillation); evaluating model size, latency, throughput, and memory consumption; generating reports in CSV, HTML, and PDF formats; visualising trade-offs between quality and efficiency; and comparing multiple models and compression techniques. The learner uses the framework to benchmark at least five models (Llama-3-8B, Mistral-7B, Phi-3-mini, Qwen1.5-7B, DeepSeek-7B) with three compression techniques each. The benchmarking results are published in a public repository with full documentation.

\paragraph{Extended Task 12.} An interactive web application for compression benchmarking is created using Streamlit or Gradio. The application must provide the following capabilities: selecting a model and compression technique; visualising model size, latency, and performance metrics; comparing multiple models and compression techniques side-by-side; generating downloadable reports (CSV, HTML); and exploring the trade-offs between quality and efficiency through interactive plots. The application is designed in such a way that it can be used by a practitioner without programming skills and is deployed locally with the possibility of subsequent deployment to Hugging Face Spaces.

\paragraph{Extended Task 13.} A suite of unit tests is developed based on the pytest framework, verifying the correctness and reproducibility of the key project components. The tests must check: the correctness of model size measurement (matches PyTorch's \texttt{model.parameters()}); the correctness of latency measurement (consistent across runs); the correctness of quantisation (weights are quantised correctly); the correctness of distillation loss (matches the implementation); the correctness of the compression pipeline; and the stability of benchmark results upon repeated runs. All dependencies are recorded in the \texttt{requirements.txt} file with an indication of the exact versions. A script is developed for the automatic execution of the full test suite and the generation of a pass/fail report.

\subsection{Further Research Directions (Optional)}
\textbf{First Additional Task} --- hardware-aware compression: optimisation of compression techniques for specific hardware (CPU, GPU, TPU, edge devices); implementation of hardware-specific quantisation (e.g., INT8 for GPU, FP16 for TPU).

\textbf{Second Additional Task} --- mixed-precision quantisation: investigation of quantisation with different bit widths for different layers (e.g., 4-bit for embeddings, 2-bit for attention, 8-bit for FFN); development of a mixed-precision quantisation algorithm.

\textbf{Third Additional Task} --- knowledge distillation with multiple teachers: aggregation of knowledge from multiple teacher models; analysis of teacher diversity and its impact on student performance.

\textbf{Fourth Additional Task} --- pruning with re-training: comparison of one-shot pruning and iterative pruning with re-training; analysis of the optimal pruning schedule.

\textbf{Fifth Additional Task} --- model merging: investigation of merging multiple fine-tuned models (task arithmetic, model merging) and its impact on model size and performance; comparison with compression techniques.

\textbf{Sixth Additional Task} --- energy efficiency benchmarking: measurement of energy consumption (watts, carbon footprint) for different compression techniques; development of energy efficiency metrics and recommendations.

\textbf{Seventh Additional Task} --- compressed RAG systems: application of compression to RAG systems (compressing the retriever, the generator, or both); analysis of the impact of compression on RAG quality and latency.

\textbf{Eighth Additional Task} --- compression for edge devices: deployment of compressed models on mobile phones (Android, iOS) or embedded devices (Raspberry Pi); evaluation of inference speed, memory consumption, and energy consumption.

\section{Conclusion}
This work is not simply ``making models smaller''. It is the systematic engineering of efficient AI systems that can be deployed in the real world, where computational resources are finite, latency matters, and every dollar counts. Model compression transforms research prototypes into production-ready systems, democratising access to state-of-the-art AI by enabling deployment on consumer hardware, edge devices, and cost-effective cloud infrastructure.

In the course of performing the work, the learner learns to ask the key questions that determine the maturity of production AI systems. Whether the model is small enough to fit in the available memory (a question about memory efficiency). Whether the model is fast enough to meet latency requirements (a question about speed). Whether the compressed model still performs the task with acceptable quality (a question about accuracy retention). Whether the compression technique is compatible with the target hardware (a question about hardware alignment). Whether the compression pipeline is reproducible and can be applied to new models (a question about engineering discipline).

The answers to these questions constitute the difference between a `research model` and a `production-ready model`. Upon completion of the work, the learner receives not simply a set of compressed models but a methodological framework for designing, evaluating, and deploying efficient AI systems --- from small-scale experiments to large-scale production deployments, capable of delivering state-of-the-art performance within real-world constraints.

\chapter{LLMOps: Deployment, Monitoring and Lifecycle Management}
\label{work:18}

\section{Aim and Learning Outcomes}
The objective is to connect a working NLP predictor to reproducible deployment practice. Students create testable endpoints, validate inputs, observe response behaviour and specify monitoring and rollback requirements. The work separates what a small local integration test can actually demonstrate from claims about security, service reliability or production performance.

The three tasks below constitute the required experimental core. Larger engineering studies and independent research questions are available as optional extensions.

\section{Theoretical Background}

\noindent\textit{Principal theoretical and methodological sources for this work:} \cite{ru_reference_73,huyen2022designing,sculley2015hidden,polyzotis2019ml}.\par

The lifecycle of a machine learning model does not end at the training stage. In real-world operation, a research prototype demonstrating high quality on static test data faces numerous challenges: changing input data distributions (drift), increasing load, the need to update the model without service interruption, and requirements for security and regulatory compliance. LLMOps (Large Language Model Operations) is a discipline that combines DevOps, MLOps, and data engineering practices to manage the full lifecycle of LLM systems: from development and deployment to monitoring, updating, and decommissioning. Mastering LLMOps transforms a researcher into an engineer capable not only of creating a model but also of making it accessible, reliable, and maintainable under real-world operational conditions. \cite{sculley2015hidden,huyen2022designing,polyzotis2019ml}

The LLMOps pipeline architecture includes several key components, each solving a specific task critical for industrial operation.

The first component --- the API server --- serves as the entry point for all external requests to the model. The modern standard is a REST API built on FastAPI or similar frameworks. The API server is responsible for receiving requests, validating input data, performing model inference, and returning structured responses. Key requirements for the API server include asynchronous processing (to support high loads), caching of frequently repeated requests (to reduce latency and cost), rate limiting (to protect against abuse), and automatic documentation generation (OpenAPI/Swagger). FastAPI provides built-in support for asynchrony, validation through Pydantic, and automatic documentation, making it the preferred choice.

The second component --- containerisation and orchestration --- ensures reproducibility of the runtime environment, isolation of dependencies, and scalability. Docker allows packaging the service with all its dependencies (Python libraries, system packages, model weights) into a single image that can be reproduced on any platform. Docker Compose allows orchestrating several related containers (API server, vector database, Prometheus, Grafana) into a single system, simplifying local deployment and testing. For industrial scale, Kubernetes is used, providing automatic scaling, self-healing, and configuration management.

The third component --- monitoring and logging --- is critically important for maintaining service reliability in production. Monitoring includes collecting performance metrics (response latency at percentiles --- p50, p95, p99; throughput --- requests per second; GPU usage --- load, memory; token consumption), quality metrics (accuracy, precision, recall, F1 on a sample validation set or through LLM-as-a-Judge), as well as system-level metrics (CPU usage, memory, disk space). Prometheus is the de facto standard for collecting metrics in microservice architectures, and Grafana provides powerful visualisation tools and dashboard construction. Logging includes recording all requests and responses (with anonymisation of personal data), errors, and warnings, enabling incident investigation and system behaviour analysis.

The fourth component --- experiment tracking and model version management --- ensures reproducibility and the ability to roll back to previous versions. MLflow (or Weights \& Biases) allows logging training parameters, metrics, artefacts (model weights, configurations), and linking them to code versions. The Model Registry provides centralised storage, versioning, and management of model lifecycle stages (Staging, Production, Archived). This makes it possible to track which model is currently in production and to quickly roll back to a previous version when problems are detected.

The fifth component --- drift detection --- enables the detection of changes in the distribution of input data or embeddings that may lead to model quality degradation. Data drift is detected by comparing feature distributions in the training set and the current request stream. Embedding drift is detected similarly, but at the level of vector representations, which is more sensitive to semantic shifts. Statistical tests (Kolmogorov--Smirnov, Wasserstein distance, MMD) with configurable thresholds are used for detection. When drift is detected, model retraining or engineer notification may be initiated.

The sixth component --- deployment strategies --- determines how new model versions are introduced into production without risk to users. Canary deployment involves gradually switching traffic to the new version: initially 5--10\% of requests are handled by the new version, while 90--95\% are handled by the old version. If the new version's metrics do not degrade (latency, errors, quality), the traffic share is gradually increased to 100\%. If metrics degrade, automatic rollback is performed. A/B testing in production allows comparing two model versions under real-world conditions, collecting statistically significant data on quality and user experience.

The seventh component --- security --- includes protection against prompt injection (separation of system and user prompts, escaping of special characters), filtering of personal data (PII) before logging, input validation (length limitation, format checking), authentication via API keys, and rate limiting to prevent abuse. The implementation of these mechanisms is critically important for compliance with regulatory requirements (GDPR, 152-FZ) and protection of the service's reputation.

\subsection{Experimental Interpretation}
Service health, robustness, factuality and safety are separate properties. A successful /health request does not establish model correctness, nor does a fixed seed make networked systems bitwise deterministic. Monitoring must specify denominators, sampling windows, alert thresholds and privacy protections appropriate to the deployment.

\section{Execution Procedure}

\subsection{Core: Three Required Tasks}
\begin{enumerate}[leftmargin=*,label=\textbf{\arabic*.}]

\item Package a documented NLP classifier behind a minimal prediction endpoint and a health endpoint with input validation.

Choose a fixed, documented text model and package it with an explicit input schema, output schema and one versioned configuration. Implement a prediction endpoint and a health endpoint, defining how empty or oversized input is rejected and what the service returns when the model is unavailable. Keep test credentials and source data non-sensitive. Record whether preprocessing occurs inside the service or in the client so that the inference boundary is reproducible.

\item Run functional and integration checks on the endpoints using test data, and record software versions, errors and response timing.

Run repeatable tests for valid, invalid and edge-case requests, including at least one batch or concurrency scenario if the implementation supports it. Check HTTP status codes, output types, model version metadata and error handling. Measure response time over multiple requests and report the machine, input sizes and concurrency settings. Distinguish a successful endpoint test from evidence of production-scale throughput, fault tolerance or monitoring readiness.

\item Create a basic monitoring plan covering request volume, latency, failures, data/model drift and rollback criteria. State explicitly which of these were measured in the demonstration and which are proposed controls.

Specify measurable operational signals, their collection points and a practical procedure for investigating changes in data or predictions. Define alert conditions and rollback decisions as proposed policies unless they are genuinely implemented and tested. Identify privacy constraints for request logging, including data minimisation and retention choices. Provide a concise deployment and incident-response plan connecting model version, test result, monitoring evidence and recovery actions without claiming that a classroom prototype has already met production service levels.

\end{enumerate}

\subsection{Reproducible Example and Its Limits}

The companion project contains executable code for this work. Run the real-data path with
\begin{center}\texttt{python practikum\_examples.py 18}\end{center}
Real-data baseline served in a FastAPI TestClient; functional endpoint checks only. For a technical-only check, add \texttt{-{}-smoke}; its synthetic inputs must not be used as scientific evaluation results. The program does not contain fabricated model scores. Record actual outputs, sample sizes, versions, timings and failures after running it.

\subsection{Extended Engineering Tasks (Optional)}
The following tasks provide a fuller engineering and research programme. They are optional rather than prerequisites for passing the core. The numerical scales and model choices below describe ambitious projects; they are not universal requirements. Documentation and reporting follow the shared protocol above.

\paragraph{Extended Task 1.} The RAG system built in Practical Work~\ref{work:12} or the agent system from Practical Work~\ref{work:15} (at the learner's choice) is used as the base service for deployment. The service must be fully functional, tested on the test data, and ready for wrapping in an API. If neither of the previous systems was implemented to a sufficient extent, the learner has the right, in agreement with the instructor, to use a simplified version (e.g., a simple classifier or a generative model without RAG) with the mandatory indication of this circumstance in the report and a discussion of its influence on the completeness of the LLMOps pipeline. All artefacts (model weights, configurations, index files) are saved in a structured directory ready for inclusion in the Docker image. References to the artefacts of the previous works are recorded in the methodological section of the report.

\paragraph{Extended Task 2.} A REST API for the service is developed using FastAPI. The API must provide at least the following endpoints: \texttt{/predict} (POST) --- the main endpoint for inference, accepting a request in JSON format with the field \texttt{query} (string, mandatory) and optional fields \texttt{top\_k} (number of relevant chunks, default 5) and \texttt{temperature} (generation parameter, default 0.1); \texttt{/health} (GET) --- an endpoint for checking the service's health, returning the status \texttt{ok}; \texttt{/metrics} (GET) --- an endpoint for exporting metrics in Prometheus format (via the \texttt{prometheus-client} library). All endpoints must be documented through automatic OpenAPI generation. Input validation is performed using Pydantic: it is checked that the query length does not exceed 4096 characters, that \texttt{top\_k} is in the range 1--20, and that \texttt{temperature} is in the range 0--1. Caching is implemented: if an identical query was executed within the last 24 hours, the response is returned from the cache without accessing the model. The \texttt{cachetools} library or Redis (optional) is used for caching. The API code is saved in the file \texttt{app.py}.

\paragraph{Extended Task 3.} A Docker image is created based on the FastAPI application. A Dockerfile is written that: uses the official Python 3.10-slim image as the base; installs system dependencies (gcc, libpq-dev, and others necessary for GPU operation); copies the \texttt{requirements.txt} file and installs Python dependencies; copies the application code and model artefacts (weights, configurations, FAISS index); exposes port 8000; defines the startup command \texttt{uvicorn app:app --host 0.0.0.0 --port 8000}. All dependencies with exact versions are recorded in \texttt{requirements.txt} (\texttt{fastapi}, \texttt{uvicorn}, \texttt{prometheus-client}, \texttt{cachetools}, \texttt{pydantic}, \texttt{transformers}, \texttt{torch}, \texttt{faiss-cpu}). The Docker image is built and tested locally. The image size should not exceed 5 GB (when using quantised models) or be justifiably larger with the reason indicated in the report.

\paragraph{Extended Task 4.} Orchestration is configured using docker-compose. A \texttt{docker-compose.yml} file is written, defining at least three services: \texttt{api} --- a container with the FastAPI application based on the built image; \texttt{prometheus} --- a container with Prometheus for collecting metrics; \texttt{grafana} --- a container with Grafana for dashboard visualisation. If necessary, services for the vector database (Qdrant or FAISS as a separate service with an index) and Redis for caching are added. Networks, volumes for data storage (logs, metrics, index), and environment variables are defined. The entire system must be launched with a single command \texttt{docker-compose up -d}. It is verified that all containers start successfully and interact with each other.

\paragraph{Extended Task 5.} Performance metric collection is configured using the \texttt{prometheus-client} library. The following metrics are defined: \texttt{http\_requests\_total} (counter of total requests with labels for method, endpoint, and response status), \texttt{http\_request\_duration\_seconds} (histogram of request duration with percentiles), \texttt{model\_inference\_duration\_seconds} (histogram of model inference time), \texttt{model\_tokens\_consumed} (counter of tokens consumed on input and output), \texttt{model\_memory\_usage\_bytes} (model memory usage), \texttt{cache\_hits\_total} and \texttt{cache\_misses\_total} (counters of cache hits and misses). All metrics are exported on the \texttt{/metrics} endpoint. Prometheus is configured to collect these metrics at 15-second intervals. It is verified that all metrics are correctly displayed in Prometheus.

\paragraph{Extended Task 6.} A dashboard for visualising the collected metrics is created in Grafana. The dashboard must include at least the following panels: total number of requests over the last hour (time-accumulated graph); response latency distribution (p50, p95, p99) with the ability to filter by time window; GPU/CPU and memory usage; token consumption (input and output) over time; proportion of successful/erroneous responses; number of cache hits. The dashboard must be exported in JSON format and saved in the repository. It is verified that the dashboard correctly displays metrics from Prometheus and can be used for operational monitoring of the service's status.

\paragraph{Extended Task 7.} The MLflow experiment tracking system is integrated. Logging of each request and response (with PII anonymisation) to an MLflow experiment is implemented in the API code. For each request, the following are logged: the input prompt (hashed or anonymised), the generated response, inference time, the model used (version identifier), as well as quality metrics (if a reference answer is available). An MLflow Tracking Server (locally or in the cloud) is configured for centralised storage of all experiments. The ability to switch between model versions through the MLflow Model Registry is implemented: the service administrator can change the active model version without restarting the API. It is verified that all requests are logged and that model switching works correctly.

\paragraph{Extended Task 8.} A data drift and embedding drift detection mechanism is implemented. For data drift detection, every 1000 requests, distribution statistics are computed: query length in characters, query length in tokens, word frequency, part-of-speech distribution (using the spaCy library), and compared with reference distributions from the training set. For embedding drift detection, the mean vector and covariance matrix for query embeddings are computed and compared with reference ones using Wasserstein distance and MMD. When the threshold is exceeded (configurable, by default the 95th percentile of the distance on the validation set), a warning is generated in the logs and, if necessary, an alert is sent (e.g., via a Telegram bot or email). The drift detection results are visualised on a separate panel in Grafana as a time series of the distance between distributions.

\paragraph{Extended Task 9.} Canary deployment is implemented using two API instances: one with the current (old) model version, the other with the new version. A load balancer is configured (a simple Python script or nginx can be used), which directs 90\% of traffic to the old version and 10\% to the new version. Metrics are collected for each version (latency, errors, quality on a sample validation set). If, within 10 minutes, the new version's metrics do not degrade (latency does not exceed 120\% of the old version, error rate does not exceed 105\% of the old version, validation quality does not drop by more than 2 percentage points), the traffic share to the new version is gradually increased (30\%, 50\%, 80\%, 100\%) with 5-minute intervals between stages. If at any stage metrics degrade, an automatic rollback to the old version is performed (100\% traffic to the old version). The deployment process is logged and visualised in Grafana. The implementation may be simplified (simulation of two versions within a single API) provided this is explicitly indicated in the report.

\paragraph{Extended Task 10.} Basic security mechanisms are implemented. Authentication via API keys is implemented: each request must contain the header \texttt{X-API-Key}; the service checks the key against a list of permitted keys (stored in an environment variable). Rate limiting is implemented: each API key is limited to 100 requests per minute (configurable parameter). If the limit is exceeded, HTTP 429 Too Many Requests is returned. Personal data (PII) filtering in logs is implemented: before logging, phone numbers, email addresses, passport data are removed from the text using regular expressions or the \texttt{pii-detection} library. Protection against prompt injection is implemented: the system prompt is separated from the user prompt, and special characters (e.g., ``System:'') are escaped. It is verified that all security mechanisms work correctly and do not interfere with the main functionality.

\paragraph{Extended Task 11.} An interactive web interface for monitoring the service status is developed using Gradio or Streamlit. The interface must provide the following capabilities: display of the current service status (online/offline, active model version, number of requests over the last hour); real-time metric viewing (latency, errors, GPU usage) with automatic update every 5 seconds; A/B test management --- manual switching of the traffic share between model versions; log viewing (last 100 entries with filtering by error level); trigger for manual model retraining on fresh data (simulation). The application must connect to Prometheus and MLflow to obtain data. The interface is designed in such a way that it can be used by an operations engineer without deep NLP knowledge.

\paragraph{Extended Task 12.} A suite of unit and integration tests is developed based on the pytest framework, verifying the correctness and reliability of all components. Unit tests verify: the correctness of input data validation (Pydantic); the correctness of cache operation (a repeated request returns the same response); the correctness of metric collection (Prometheus metrics are updated). Integration tests verify: the \texttt{/predict} endpoint returns a response of the correct format; the \texttt{/health} endpoint returns status 200; containers start successfully via docker-compose; Prometheus correctly collects metrics; MLflow correctly logs requests. The tests must run automatically in a CI/CD pipeline (e.g., via GitHub Actions) and generate a pass/fail report.

\subsection{Further Research Directions (Optional)}
\textbf{First Additional Task} --- cloud deployment (AWS, Yandex Cloud, Google Cloud): use of managed services (e.g., Yandex Managed Service for Kubernetes) for deploying a cluster with automatic scaling; comparison of cost and performance with local deployment.

\textbf{Second Additional Task} --- CI/CD integration: configuration of GitHub Actions or GitLab CI for automatic Docker image building, test execution, and deployment on every push to the main branch; implementation of automatic rollback on test failures.

\textbf{Third Additional Task} --- cache autocompletion configuration (Redis): use of Redis for caching embeddings and responses; evaluation of latency and cost reduction under various caching strategies (LRU, TTL).

\textbf{Fourth Additional Task} --- energy consumption assessment: integration of CodeCarbon for measuring the carbon footprint of inference in production; development of recommendations for energy-efficient operation.

\textbf{Fifth Additional Task} --- advanced security: implementation of JWT-based authentication, traffic encryption (TLS/SSL), integration with a secret management system (HashiCorp Vault).

\textbf{Sixth Additional Task} --- automatic retraining: configuration of a pipeline that, upon drift detection, automatically initiates model retraining on fresh data, performs validation, and, upon success, initiates canary deployment of the new version.

\textbf{Seventh Additional Task} --- multi-model deployment: deployment of several models (classification, NER, summarisation) in a single API with request routing based on task type; comparison of monolithic and microservice architecture.

\textbf{Eighth Additional Task} --- operational cost analysis: detailed calculation of inference cost in various scenarios (local GPU vs cloud instances vs API services) taking into account latency, throughput, and energy consumption; formulation of recommendations for cost optimisation.

\section{Conclusion}
This work is not simply ``running FastAPI and Docker''. It is the systematic design of an industrial LLM system, where every decision --- from the choice of metrics to the configuration of drift thresholds and the canary deployment strategy --- determines the reliability, observability, and security of the service in real-world operation.

In the course of performing the work, the learner learns to ask the key questions that determine the maturity of industrial AI systems. Whether it is possible to see how the model works in real time --- or whether it remains a ``black box'' (a question about observability and monitoring). How to know that the model has started to degrade --- before users complain (a question about drift detection). How to update the model without service interruption and risk to users (a question about deployment strategies). How to protect the system from attackers and data leaks (a question about security). Whether reproducibility of deployment is guaranteed --- from the Docker image to the Prometheus configuration (a question about engineering discipline).

The answers to these questions constitute the difference between a ``research prototype'' and a reliable, scalable, and maintainable industrial system. Upon completion of the work, the learner receives not simply a set of scripts but an architectural compass for designing, evaluating, and operating any LLM solutions --- from startups to corporate platforms, capable of working not only accurately but also reliably, observably, securely, and reproducibly under real-world load.

\chapter*{Final Integrative Project: A Domain-Specific NLP System}

\noindent\textit{Relevant methodological and safety foundations:} \cite{who2024lmm}.\par

\phantomsection\addcontentsline{toc}{chapter}{Final Integrative Project: A Domain-Specific NLP System}
\label{chap:capstone}
\begingroup
\setlength{\parskip}{2.5pt}
\setlist[enumerate]{topsep=3pt,itemsep=1pt,parsep=0pt}

\section*{Purpose and educational role}
The final project integrates competencies acquired throughout the practical works into one coherent, testable application. The student selects \emph{one} domain and specifies its users, inputs, outputs, permissible data sources and criteria for success. The project is part of the complete course assessment, not a nineteenth practical work: the eighteen topic-focused works retain their numbering and their own core tasks. A student does not have to implement all algorithms discussed in the book. Instead, the project must demonstrate justified choices across the main stages of an NLP workflow: data preparation, a baseline, modelling, evaluation, error analysis and responsible operation. Its complexity is adapted to available resources and the student's level of study.

The aim is not to assemble the greatest possible number of libraries or agents, but to determine which components make a measurable difference for a clearly defined task. In a project involving generation, a design such as \emph{Generator--Validator--Critic--Editor} coordinated by a \emph{Supervisor} is one possible extension. A simpler generator with an independently evaluated checking step is a valid core design. Likewise, retrieval-augmented generation (RAG) is required only when the application calls for an external evidence base, not for every poem or joke. Claims of superior performance must be supported by experiments, not by the presence of architectural components.

\section*{Choice of application area}
Each student or team chooses one of the following directions or proposes an equivalent task with an assessable output. Illustrative data sources must be checked for access permissions, licensing, coverage and quality before use.
\begin{enumerate}[leftmargin=*]
\item \textbf{Machine translation and terminology control.} Translate text between a documented language pair, possibly including a low-resource language. Compare an available translation baseline with a terminology-aware or retrieval-assisted version; assess semantic adequacy, terminology preservation and human-reviewed errors. Open parallel corpora such as those distributed through OPUS or Tatoeba may be considered subject to corpus-specific licensing and split integrity.
\item \textbf{Stories, fairy tales and fictional narratives.} Generate a story from a character list, setting, age-appropriate constraints and a plot outline. Evaluate constraint satisfaction, entity consistency, coherence and originality using a clearly defined human rating rubric. Do not reuse protected works or imitate identifiable living authors as if they endorsed the output.
\item \textbf{Poetry and constrained creative writing.} Generate poems under selected constraints such as theme, structure, rhyme or syllabic form. Evaluate formal compliance separately from reader preference; report annotation disagreements instead of claiming that a single automatic metric measures artistic quality.
\item \textbf{Humour and short-form writing.} Generate jokes, humorous dialogues or short sketches with a specified audience and style. Study comprehensibility, originality, repetition and potentially offensive outputs using consenting adult reviewers and a documented rubric. A successful output need not be maximally ``funny'' to every reader.
\item \textbf{Historical essays and evidence-based narratives.} Produce a short account of an event using a fixed, licensable corpus of historical sources. Evaluate dates, named entities, citation support and unsupported statements. Keep invented literary narration clearly distinguishable from claims presented as historical facts.
\item \textbf{Product descriptions from structured attributes.} Generate catalogue descriptions given a product's specifications, intended audience and prohibited claims. Measure how completely supplied attributes are covered and how frequently unsupported features, guarantees or specifications are invented. Structured input can be created by the student and labelled as illustrative.
\item \textbf{Document-grounded question answering and educational explanations.} Answer questions using an authorised document collection. Compare retrieval-free and retrieval-assisted approaches when meaningful; assess citation support, abstention on unanswerable questions and sensitivity to irrelevant retrieved text.
\item \textbf{Medical information explanation (restricted educational prototype).} Explain terminology or summarise vetted, non-personal health information for a hypothetical reader. Use synthetic cases or public educational material and document source provenance. Do \emph{not} generate individual diagnoses, treatment plans, medication choices or triage decisions. Outputs must not be presented as professional medical advice or deployed to patients; expert review is required before any clinical use.
\item \textbf{Psychoeducation and wellbeing resources (restricted educational prototype).} Provide general explanations of recognised wellbeing concepts or route fictional observations to appropriate public information resources. Do \emph{not} infer a mental disorder from personal observations, assign diagnoses, recommend psychotherapy or treatment for an individual, or impersonate a clinician. Avoid identifiable personal records, define escalation boundaries, and obtain competent review for any proposed real-world use.
\end{enumerate}
For projects involving high-stakes information, privacy, human oversight, reliability and limitations must be considered at the design stage, consistent with the WHO guidance on large multimodal models in health and the NIST generative-AI risk-management profile \cite{who2024lmm,nist2024genai}. A student may instead select a non-clinical, non-personal topic when expert oversight cannot be arranged.

\section*{Core: Three Required Project Stages}
\begin{enumerate}[leftmargin=*,label=\textbf{Stage \arabic*.},itemsep=7pt]
\item \textbf{Formulate the problem, prepare evidence and establish a baseline.} Describe the use case, intended users, permissible outputs, evaluation criteria and potential harms. Assemble or select an appropriately licensed dataset or a documented set of structured task instances; record the language, domains, source, permissions, relevant preprocessing and limitations. Establish a training/development/test protocol, using document-, author-, group- or time-aware separation where needed to prevent leakage. Build a transparent baseline appropriate to the chosen task: e.g., an existing translation system, retrieval-only responses, a deterministic description template, or an unassisted pretrained generator. The baseline must be evaluated on the same held-out material used for later comparison. For small creative datasets, a predeclared test set with an independent human rubric is acceptable; inventing numerical quality results is not.
\item \textbf{Implement a complete and inspectable NLP pipeline.} Construct a working path from input validation and normalisation to model or rule-based processing and a documented output. Connect at least two meaningful components from different parts of the course, such as retrieval plus grounded generation, extraction plus generation, classification plus constrained rewriting, or generation plus deterministic quality checks. Record model identifiers and revisions, prompts or templates, preprocessing settings, decoding parameters, dependency versions and hardware. Add a simple independent check appropriate to the task (e.g., required-attribute verification, source citation support, placeholder preservation or format constraints). When useful, implement a Generator--Critic--Editor or Supervisor workflow, a hybrid BM25/vector retriever or a human-review queue as an \emph{optional} extension; none is universally necessary. The pipeline must stop predictably on invalid input or component failure, rather than silently returning invented evidence.
\item \textbf{Evaluate, compare and defend the result.} Compare the baseline with the completed pipeline on the same held-out set, recording test-set sizes, actual observations and any human-evaluation protocol. Use task-matched measures rather than applying one general score to every application: translation may use chrF/BLEU together with adequacy review; grounded answers require supported-claim and abstention checks; product descriptions require attribute coverage and hallucinated-claim counts; creative writing requires explicit human criteria and inter-rater disagreement. Inspect both successful and failed examples, report cost or latency where measurable and perform at least one focused component comparison (for example, with and without retrieval, deterministic checking or a revision step). Report observed differences without treating a small sample or an LLM judge as independent proof. Discuss errors, legal provenance, safety constraints, and conditions under which the system should decline or defer a request.
\end{enumerate}

\section*{Indicative architecture and comparisons}
A possible flow is \emph{input and specification $\rightarrow$ cleaning and validation $\rightarrow$ retrieval (if required) $\rightarrow$ generator or task model $\rightarrow$ deterministic validator $\rightarrow$ optional critic and editor $\rightarrow$ final output and logged evaluation}. The \emph{Supervisor} in a multi-agent variant manages routing, an iteration budget, termination, failures and optional rollback. The Critic's score is an \emph{estimate}, not an authoritative label: corrections and regressions should be measured independently whenever possible. A fixed random seed is useful to document but does not guarantee identical generation across hardware and inference backends.

Ablations may compare \texttt{baseline}, \texttt{+retrieval}, \texttt{+validation}, \texttt{+revision} and \texttt{full} only when those configurations are meaningful for the selected task. Distinguish an agent role from a separate model or service: several roles can be executed by one model, and several processes do not automatically imply better reasoning. Evaluate incremental gains against additional inference calls, latency, memory and maintenance burden.

\section*{Depth by level of study}
\begin{description}[leftmargin=*,style=nextline]
\item[Senior undergraduate] Deliver all three required stages using modest resources, one reproducible baseline, one complete pipeline and a meaningful held-out comparison. A notebook or command-line demonstration is sufficient; a web application and cloud deployment are not obligatory.
\item[Master's student] Add a clearly motivated engineering or research question, a stronger or contrasting baseline, a controlled ablation and a more detailed treatment of reliability or human assessment. Explain whether improvements exceed the measurement uncertainty supported by the data.
\item[Doctoral researcher] Formulate a novel or under-tested hypothesis; investigate transfer, robustness, multilingual or low-resource behaviour, uncertainty, independent human assessment or reproducible efficiency trade-offs. The expectation is stronger evidence and scientific argumentation, not merely a larger software stack.
\end{description}
Students may work individually or in teams if approved by the instructor. A team submission must describe each member's independent contribution. No minimum GPU size, fixed number of tokens, paid API subscription or requirement to publish proprietary data is imposed for all domains.

\section*{Optional engineering and research extensions}
\begin{enumerate}[leftmargin=*]
\item \textbf{Collaborative reasoning:} compare a single model, Generator--Critic and Generator--Critic--Editor with a Supervisor, bounded revision cycles, error-aware rollback and an external reference-based evaluation.
\item \textbf{Hybrid RAG:} combine BM25 and vector retrieval, compare fusion methods and document relevance judgements, source attribution and contamination checks.
\item \textbf{Human-in-the-Loop:} design an explicit review queue for ambiguous or unsafe outputs; study how reviewer corrections change quality without contaminating the test set.
\item \textbf{Adaptation and efficiency:} compare suitable prompting, parameter-efficient fine-tuning or quantisation when licences, compute and task data allow.
\item \textbf{Engineering and LLMOps:} build an API, a user interface, containerisation, automated tests, access controls and monitoring when they address a real delivery need; do not require all these services for an offline poetry experiment.
\item \textbf{Research extensions:} investigate cross-language transfer, creative constraints, factual grounding, judge bias, safety failures, dataset shift or energy/latency trade-offs using a defensible evaluation design.
\end{enumerate}
These extensions are optional; the number of technologies used is not itself evidence of competence.

\section*{Deliverables and assessment}
Submit a working demonstration with code; a brief project specification; legally sourced or shareable sample inputs; an executable configuration or environment description; a baseline-versus-system results table with real observations; an error and safety analysis; and a concise report following the shared requirements at the beginning of the book. Where material cannot be distributed, provide lawful access arrangements or a minimal synthetic reproduction and explain what could not be reproduced. A short live defence should demonstrate a normal input, a boundary or failure case, and the evidence supporting the main conclusion.

A suggested rubric for the final project is: research question and data provenance (20\%), implementation and component justification (25\%), experimental evaluation and credible baselines (25\%), error analysis and responsible use (15\%), reproducibility and tests (10\%), and presentation and defence (5\%). A strong small project with a fair comparison may outperform an elaborate system without evidence. Project grades should reflect the student's level and the objectives agreed with the instructor; optional experiments may enrich the evaluation without becoming hidden compulsory requirements.

\medskip\noindent\textbf{Expected educational outcome.} The completed project demonstrates an ability to integrate selected NLP methods into a reliable, testable solution, distinguish measured results from assumptions and articulate where an apparently attractive architecture does not improve the task. The outcome may support a portfolio, a master's thesis or an initial doctoral research question, but scientific publication is neither necessary nor guaranteed.
\endgroup

\backmatter
\renewcommand{\bibname}{References}
\phantomsection
\addcontentsline{toc}{chapter}{References}
\bibliographystyle{unsrt}
\bibliography{references}
\end{document}